\documentclass{article}

\usepackage{microtype}
\usepackage{graphicx}
\usepackage{subcaption}
\usepackage{booktabs} 

\usepackage{hyperref}

\newcommand{\tightbox}[2]{\begingroup\setlength{\fboxsep}{0pt}\colorbox{#1}{#2}\endgroup}

\usepackage[preprint]{icml2026}

\usepackage{amsmath}
\usepackage{amssymb}
\usepackage{mathtools}
\usepackage{amsthm}

\usepackage[capitalize,noabbrev]{cleveref}

\theoremstyle{plain}

\theoremstyle{definition}

\theoremstyle{remark}

\usepackage[textsize=tiny]{todonotes}

\usepackage[nolist,nohyperlinks]{acronym}
\usepackage[inline]{enumitem}
\usepackage{xcolor}
\usepackage{soul}
\usepackage{multirow}
\usepackage[table]{xcolor}
\usepackage{makecell}
\usepackage{cancel}
\usepackage{dblfloatfix}
\usepackage{svg}
\usepackage{placeins}

\definecolor{bestrow}{HTML}{DFF0D8}
\definecolor{baselinerow}{HTML}{FFF4CC}

\newcommand{\ourmethod}{SIMSHIFT}
\newcommand{\textbfp}[1]{\vspace{0.5em}\noindent\textbf{#1.}}

\sethlcolor{yellow!30}
\soulregister\cite7
\soulregister\citet7
\soulregister\citep7
\soulregister\citeauthor7
\soulregister\citeyear7
\soulregister\ref7
\soulregister\cref7
\soulregister\crefrange7
\soulregister\Cref7
\soulregister\ac7
\soulregister\acs7
\soulregister\acl7
\soulregister\acf7
\soulregister\acp7

\begin{acronym}[LONGEST]
  \acro{CV}{Computer Vision}
\end{acronym}
\begin{acronym}[LONGEST]
  \acro{NLP}{Natural Language Processing}
\end{acronym}
\begin{acronym}[LONGEST]
  \acro{PDE}{Partial Differential Equation}
\end{acronym}
\begin{acronym}[LONGEST]
  \acro{CFD}{Computational Fluid Dynamics}
\end{acronym}
\begin{acronym}[LONGEST]
  \acro{DEM}{Discrete Element Method}
\end{acronym}
\begin{acronym}[LONGEST]
  \acro{SPH}{Smoothed Particle Hydrodynamics}
\end{acronym}
\begin{acronym}[LONGEST]
  \acro{UDA}{Unsupervised Domain Adaptation}
\end{acronym}
\begin{acronym}[LONGEST]
  \acro{DA}{Domain Adaptation}
\end{acronym}
\begin{acronym}[LONGEST]
  \acro{FDM}{Finite Difference Method}
\end{acronym}
\begin{acronym}[LONGEST]
  \acro{FEM}{Finite Element Method}
\end{acronym}
\begin{acronym}[LONGEST]
  \acro{FE}{Finite Element}
\end{acronym}
\begin{acronym}[LONGEST]
  \acro{FVM}{Finite Volume Method}
\end{acronym}
\begin{acronym}[LONGEST]
  \acro{UPT}{Universal Physics Transformer}
\end{acronym}
\begin{acronym}[LONGEST]
  \acro{MSE}{Mean Squared Error}
\end{acronym}
\begin{acronym}[LONGEST]
  \acro{RMSE}{Root Mean Squared Error}
\end{acronym}
\begin{acronym}[LONGEST]
  \acro{nRMSE}{normalized RMSE}
\end{acronym}
\begin{acronym}[LONGEST]
  \acro{IWV}{Importance Weighted Validation}
\end{acronym}
\begin{acronym}[LONGEST]
  \acro{DEV}{Deep Embedded Validation}
\end{acronym}
\begin{acronym}[LONGEST]
  \acro{GNN}{Graph Neural Network}
\end{acronym}
\begin{acronym}[LONGEST]
  \acro{PEEQ}{Equivalent Plastic Strain}
\end{acronym}
\begin{acronym}[LONGEST]
  \acro{RANS}{Reynolds-Averaged Navier-Stokes}
\end{acronym}
\begin{acronym}[LONGEST]
  \acro{nRMSE}{normalized Root Mean Squared Error}
\end{acronym}
\begin{acronym}[LONGEST]
  \acro{PAD}{Proxy $\mathcal{A}$-Distance}
\end{acronym}
\begin{acronym}[LONGEST]
  \acro{GINO}{Geometry-Informed Neural Operator}
\end{acronym}
\begin{acronym}[LONGEST]
  \acro{LHS}{Latin Hypercube Sampling}
\end{acronym}

\icmltitlerunning{\ourmethod: Adapting Neural Surrogates to Distribution Shifts}

\begin{document}

\twocolumn[
  \icmltitle{SIMSHIFT: A Benchmark for Adapting \\
    Neural Surrogates to Distribution Shifts}



  \icmlsetsymbol{equal}{*}

  \begin{icmlauthorlist}
    \icmlauthor{Paul Setinek}{jku}
    \icmlauthor{Gianluca Galletti}{jku}
    \icmlauthor{Thomas Gross}{lcm}
    \icmlauthor{Dominik Schn\"urer}{lcm} \\
    \icmlauthor{Johannes Brandstetter}{jku,emmi}
    \icmlauthor{Werner Zellinger}{jku}
  \end{icmlauthorlist}

  \icmlaffiliation{jku}{LIT AI Lab and Institute for Machine Learning, JKU Linz, Austria}
  \icmlaffiliation{lcm}{Linz Center of Mechatronics GmbH, Linz, Austria}
  \icmlaffiliation{emmi}{Emmi AI GmbH, Linz, Austria}

  \icmlcorrespondingauthor{Paul Setinek}{\href{mailto:setinek@ml.jku.at}{\color{black}{setinek@ml.jku.at}}}
    
  \icmlkeywords{Machine Learning, ICML, Neural Surrogates, Numerical Simulation, Dataset, Benchmark, Unsupervised Domain Adaptation}

  \vskip 0.3in
]



\printAffiliationsAndNotice{}  

\begin{abstract}
    Neural surrogates for \acp{PDE} often suffer significant performance degradation when evaluated on problem configurations outside their training distribution, such as new initial conditions or structural dimensions.
    While \ac{UDA} techniques have been widely used in vision and language to generalize across domains without additional labeled data, their application to complex engineering simulations remains largely unexplored.
    In this work, we address this gap through two focused contributions.
    First, we introduce \ourmethod, a novel benchmark dataset and evaluation suite composed of four industrial simulation tasks spanning diverse processes and physics: \textit{hot rolling}, \textit{sheet metal forming}, \textit{electric motor design} and \textit{heatsink design}.
    Second, we extend established \ac{UDA} methods to state-of-the-art neural surrogates and systematically evaluate them.
    Extensive experiments on \ourmethod~highlight the challenges of out-of-distribution neural surrogate modeling, demonstrate the potential of \ac{UDA} in simulation, and reveal open problems in achieving robust neural surrogates under distribution shifts in industrially relevant scenarios.
\end{abstract}

\section{Introduction}
\label{intro}

\ac{PDE} simulations are essential tools for understanding and predicting physical phenomena in engineering and science \citep{evans2010pde}.
Over recent years, machine learning has emerged as a promising modeling option for complex systems \citep{brunton2020machine}, significantly accelerating and augmenting simulation workflows across diverse applications, including weather and climate forecasting \citep{pathak2022fourcastnet, bodnar2024aurora}, material design \citep{merchant2023scalingdlformaterials, zeni2025generativematerials} and protein folding \citep{abramson2024accurate} to name a few.

In theory, machine learning models assume i.i.d. samples.
In practice, however, models are often deployed outside of their training distribution. This \emph{distribution shift} \citep{quionero-candela2009datasetshift,wang2023scientific} often leads to a significant performance degradation \citep{bonnet2022airfrans, herde2024poseidon}.
A prominent example is clinical microscopy: models trained with data collected at a few hospitals often fail when deployed at others because microscopes, staining protocols, and lighting conditions differ \citep{tellez2019computational_pathology, koh2020wilds}.
For neural surrogates an analogous ``instrument shift'' arises from new initial conditions, such as material parameters or mesh geometries not encountered during training.
Robustness to such distribution shifts is crucial for adoption and deployment of models also because it is becoming a compliance requirement, as stated by Article 15 of the EU AI Act~\citep{EUAIAct2024}.

While methods for increasing out-of-distribution performance have been at the center of research for a long time \citep{ben2006analysis, shimodaira2000improving, sugiyama2007direct}, to the best of our knowledge, no benchmark systematically investigates such methods on physics based simulation tasks.
Addressing this gap is particularly relevant in scientific and industrial settings, where generating ground truth simulation data is costly, limiting the diversity of training configurations.
In contrast, parametric descriptions serving as model inputs, such as material types or structural dimensions, are often readily available or easy to generate.
This problem is known as \emph{\acf{UDA}} \citep{ben-david2010learningfromdifferentdomains}, where parametric (input) descriptions and full simulation outputs are available for each \emph{source} configuration, while only input descriptions are provided for \emph{target} configurations, without corresponding outputs.

\begin{figure*}[ht]
    \centering
    \includegraphics[width=0.75\textwidth]{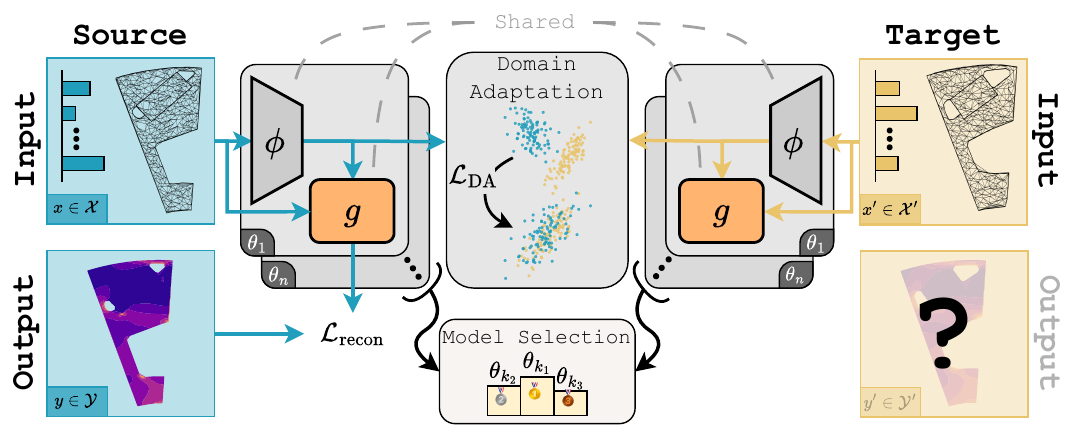}
    \caption{Schematic overview of the \ourmethod~framework.
    In training, the model has access to inputs (e.g., parameters and meshes) with the corresponding outputs $(x, y)$ from the source domain (left, blue), while only inputs $x'$ from the target domain (right, yellow) are available.
    The neural operator $g$ and the conditioning network $\phi$ are shared across domains and jointly optimized.
    Two loss terms are used: $\mathcal{L}_{\text{recon}}$, computed on source labels, and $\mathcal{L}_{\text{DA}}$, which aligns source and target feature representations produced by $\phi$.
    After training, unsupervised model selection strategies choose $\theta_{k1}$, which is expected to perform best on the target domain.}
    \label{fig:figure_1}
\end{figure*}

To investigate the potential of \ac{UDA} for neural surrogate modeling, we provide simulation data across a range of realistic tasks from industrial engineering design.
We introduce a comprehensive benchmark that evaluates established UDA methods and neural surrogates. An overview of the framework is shown in \cref{fig:figure_1}. Our contributions can be summarized as follows:

\begin{itemize}
    \item We propose four practical datasets with flexible distribution shifts in \textit{hot rolling}, \textit{sheet metal forming}, \textit{electric motor}, and \textit{heatsink} design, based on realistic simulation setups.
    \item We present, to the best of our knowledge, the first joint study of established neural surrogates and \ac{UDA} on engineering simulations with unstructured meshes.
    \item We introduce \emph{\ourmethod}, a modular benchmarking suite that complements our datasets with baseline models and algorithms. It allows for easy integration of new simulations, machine learning methods, domain adaptation techniques, and model selection strategies.
\end{itemize}

\section{Related Work}
\label{related}
\textbfp{Unsupervised Domain Adaptation}
\ac{UDA} research covers a wide spectrum of results from theoretical foundations \citep{ben-david2010learningfromdifferentdomains,zellinger2021generalization} to modern deep learning methods~\citep{Liu2021kernel, cmd, Zhu2021subdomain, Long2018conditional}.
A prominent class of methods, dubbed as \emph{representation learning}, aims to map the data to a feature space, where source and target representations appear similar, while maintaining enough information for accurate prediction. To enforce feature similarity between domains, algorithms often employ statistical \citep{coral, mmd, zhang2019bridging, shalit2017estimating} or adversarial \citep{ganin2015dann, tzeng2017adversarial} discrepancy measures.
One crucial yet frequently overlooked factor in the success of \ac{UDA} methods is model selection.
Multiple studies underline the critical impact of hyperparameter choices on \ac{UDA} algorithm performance, often overshadowing the adaptation method itself~\citep{musgrave2021realitycheck,zellinger2021balancing,dinu2023iwa,yang2024can}.
Even more, since labeled data is unavailable in the target domain, standard validation approaches become infeasible.
Thus, it is essential to jointly evaluate adaptation algorithms alongside their associated unsupervised model selection strategies.
In this work, we focus on importance weighting strategies~\citep{sugiyama2007iwv,you2019dev}, which stand out by their general applicability, theoretical guarantees and high empirical performance.

\textbfp{Benchmarks for Unsupervised Domain Adaptation}
Numerous benchmark datasets and evaluation protocols have been established for \ac{UDA} methods across various machine learning domains, including computer vision \citep{venkateswara2017officehome, peng2018syn2real, arjovsky2019irm}, natural language processing \citep{blitzer2007daforsentimentanalysis}, timeseries data \citep{ragab2022adatime} and tabular data \citep{gardner2023tableshift}.
However, to the best of our knowledge, systematic UDA benchmarking for neural surrogates remains unexplored.

\textbfp{Benchmarks for Neural Surrogates}
Recent years have seen a surge of surrogates belonging to the group of neural operators (see \cref{app:neural_operators}), and benchmarks have grown alongside them.
However, designing a robust and fair benchmark in the realm of \acp{PDE} is difficult and the current literature is not without shortcomings \citep{brandstetter2025envisioning}.
Many focus on solving \acp{PDE} on structured, regular grids \citep{gupta2022pdearena, takamoto2022pdebench, ohana2024thewell}, which serve as valuable platforms for developing and testing new algorithms.
However, these overlook the irregular meshes commonly used in large scale industrial simulations.
In that direction, other benchmarks extend to \ac{CFD} on irregular static meshes for airfoil simulations \citep{bonnet2022airfrans}, aerodynamics for automotive \citep{elrefaie2024drivaernet, elrefaie2024drivaernetpp}, more academic fluid problems \citep{luo2023cfdbench}, and even particle based Smoothed Particle Hydrodynamics simulations \citep{toshev2023lagrangebench, toshev2024jaxsph}. 
Finally, and most closely related to our work, recent efforts have explored the application of Active Learning techniques \citep{cohn1996active, ren2021survey} to neural surrogates, introducing a benchmark specifically designed for scenarios where data is scarce \citep{al4pde}.
Despite these contributions, all current benchmarks often fall short when addressing a critical issue: the performance drop models exhibit under distribution shifts, i.e., when encountering simulation configurations beyond their training setting \citep{quionero-candela2009datasetshift}.

\section{Dataset Presentation}
\label{sec:datasets}
Our datasets follow three design principles.
\begin{enumerate*}[label=(\roman*)]
    \item \textbf{Industry relevance:} They reflect practical, real-world simulation use-cases.
    The benchmark covers a diverse set of problems, including 2D as well as 3D cases.
    \item \textbf{Parametrized conditions:} The behavior of all simulations depends on the set of initial parameters only.
    \item \textbf{Steady-state scenarios:} We constrain them to time independent problems, being the standard use case in industry. Take for example design optimization tasks: most rely on either steady-state or time-averaged solutions rather than detailed transient dynamics. This is not just a modeling convenience, but reflects how simulation is integrated into design pipelines: numerical simulations are used to assess candidates by computing scalar objective values. This practice is well established various application areas, including thermal systems \citep{simulation_of_thermal_systems}, aerodynamic shape optimization for aircrafts \citep{martins2022aerodynamic_design_optimization}, wind turbine design \citep{martins2022aerodynamic_design_optimization}, and car aerodynamics \citep{Dumas2007Chapter1C}.
    Additionally with this constraint we avoid additional complexities such as autoregressive error accumulation in neural surrogates \citep{lippe2023pderefiner}.
\end{enumerate*}

The datasets were generated using the commercial \ac{FEM} software \textit{Abaqus}, the open-source simulation software \textit{HOTINT} and the open-source CFD package \textit{OpenFoam 9}, mirroring the diverse landscape of solvers used across various industry domains.
\footnote{\href{https://www.3ds.com/products/simulia/abaqus}{Abaqus}; \href{https://hotint.lcm.at/}{HOTINT}; \href{https://www.openfoam.com/}{OpenFoam 9}.}
An overview of each dataset together with its most important parameters and a custom metric, motivated by engineering practice, is presented in \crefrange{sec:rolling}{sec:heatsink}.
Additionally, we provide detailed descriptions of the respective numerical simulations in \cref{app:dataset_generation}.
\cref{table:datasets} summarizes key characteristics of each dataset, including physical dimensionality, mesh resolution and total dataset size.
All datasets are publicly hosted on Hugging Face\footnote{\href{https://huggingface.co/datasets/simshift/SIMSHIFT_data}{https://huggingface.co/datasets/simshift/SIMSHIFT\_data}}.

Since the behavior of each simulation task is entirely determined by its input parameters, we predefine source and target domains by partitioning the parameter space into distinct, non-overlapping regions.
A detailed explanation of the domain splitting strategy is provided in \cref{sec:domain_splits}.

\begin{table}[t]
\centering
\def\arraystretch{1.0}
\caption{Overview of the \ourmethod~datasets.}
\label{table:datasets}
  \def\arraystretch{1.2}
  \resizebox{\columnwidth}{!}{
  \begin{tabular}{lccccc}
    \toprule
    \textbf{Dataset} & \textbf{Samples} & \textbf{\# Avg nodes} & \textbf{Dim} & \textbf{Size} (GB) \\
    \midrule
    Rolling   & 5,000 & 508       & 2D & 3.1  \\
    Forming   & 4,000 & 9,080     & 2D & 19   \\
    Motor     & 3,195 & 4,846     & 2D & 15   \\
    Heatsink  & 512   & 4,443,114 & 3D & 520  \\
    \bottomrule
  \end{tabular}
  }
\end{table}

\subsection{Hot Rolling}
\label{sec:rolling}
\begin{figure*}[htb]
    \centering
    \begin{subfigure}{0.4\textwidth}
        \centering
        \includegraphics[width=0.82\textwidth]{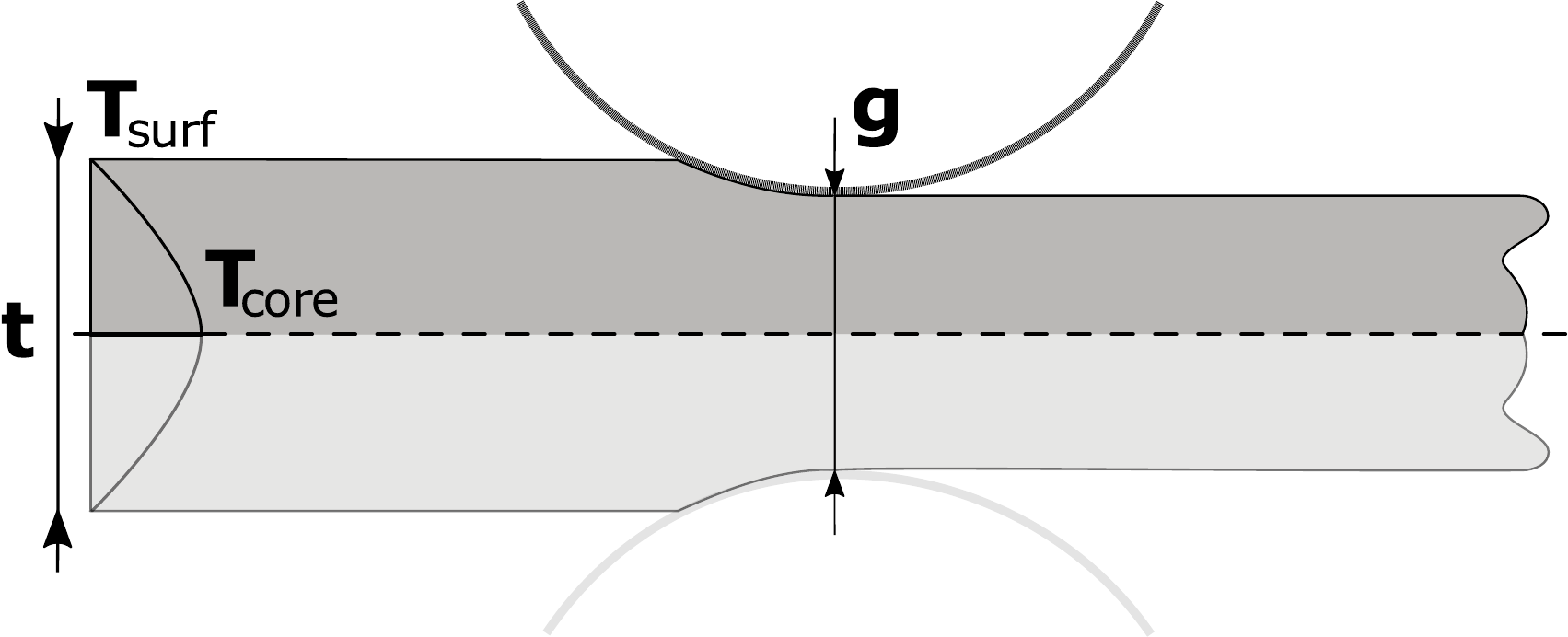}
        \caption{Illustration of the simulation setup. The parameters correspond to those in \cref{tab:rolling_params}. We use symmetry constraints and only simulate half of the slab.}
        \label{fig:rolling_sketch}
    \end{subfigure}
    \hspace{0.09\textwidth}
    \begin{subfigure}{0.4\textwidth}
        \centering
        \includegraphics[width=0.82\textwidth]{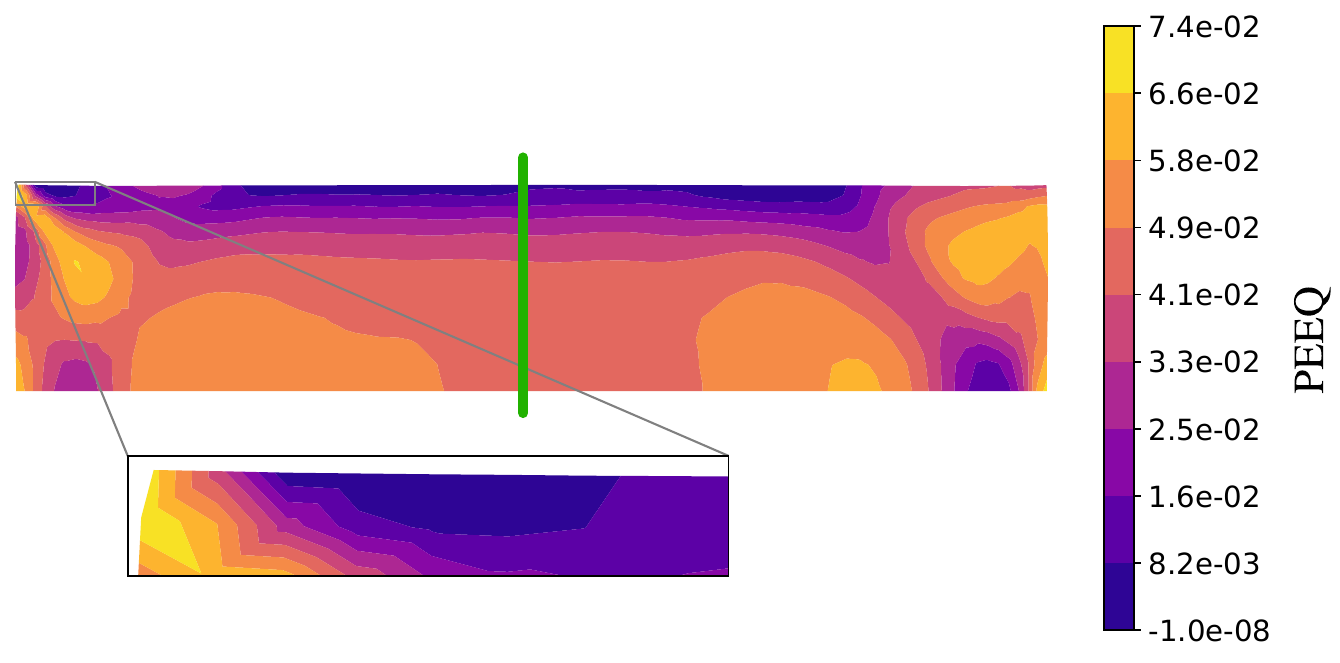}
        \caption{Metal slab after the process, showing \ac{PEEQ} as a contour plot. The green line indicates the center cord, along which we measure the custom metric.}
        \label{fig:rolling_fields}
    \end{subfigure}
    \vspace{0.6em}
    \begin{subfigure}{0.4\textwidth}
        \centering
        \includegraphics[width=0.82\textwidth]{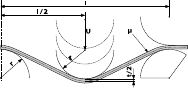}
        \caption{Illustration of the simulation setup. The parameters correspond to those listed in \cref{tab:forming_params}.}
        \label{fig:forming_sketch}
    \end{subfigure}
    \hspace{0.09\textwidth}
    \begin{subfigure}{0.4\textwidth}
        \centering
        \includegraphics[width=0.82\textwidth]{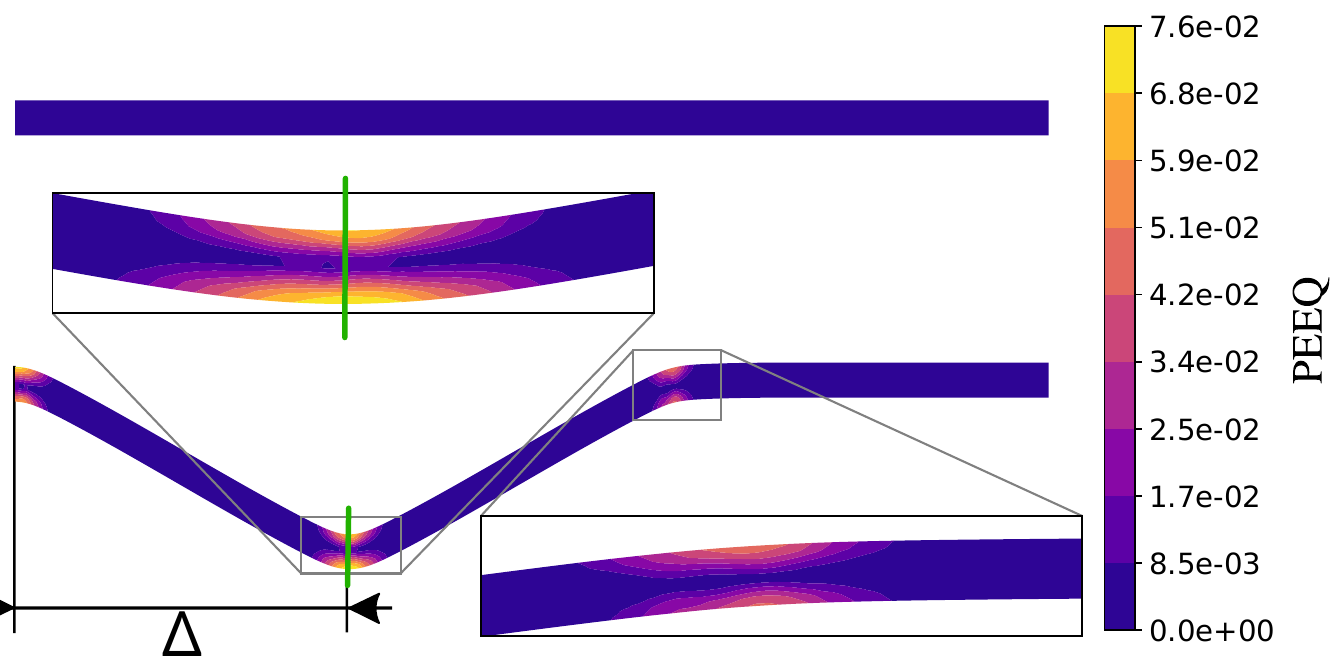}
        \caption{Material before (top) and after (bottom) the process, shown as \ac{PEEQ} contours. $\Delta=l/2$}
        \label{fig:forming_fields}
    \end{subfigure}
    \caption{Overview of the \emph{hot rolling} (top) and \emph{sheet metal forming} (bottom) simulation scenarios.}
    \label{fig:forming_overview}
\end{figure*}

\textbfp{Problem Description}
The \emph{hot rolling} process plastically deforms a metal slab into a sheet metal product, as visualized in \cref{fig:rolling_sketch,fig:rolling_fields}.
This complex thermo-mechanical operation involves coupled elasto-plastic deformation and heat transfer phenomena \citep{gupta2021steel, galantucci1999, jo2023}.
The \ac{FE} simulation models the progressive thickness reduction and thermal evolution of the material as it passes through a rolling gap, incorporating temperature dependent material properties and contact between the slab and the rolls.
Among the output fields, the key quantity is \ac{PEEQ}, representing the material's plastic deformation, visualized in \cref{fig:rolling_fields}.
The custom metric measures the relative error of the \ac{PEEQ} profile along the slab's vertical center cord (green line in \cref{fig:rolling_fields}).

\textbf{Input parameters} are the initial slab thickness $t$, temperature characteristics $T_{\text{core}}$ and $T_{\text{surf}}$ of the slab, as well as the geometry of the roll gap.
To vary the slab deformation we define the thickness reduction as a percentage of the initial thickness: $\text{reduction} = \frac{t-g}{t}$, where $g$ is the rolling gap distance.
\cref{tab:rolling_params} in \cref{app:rolling_detailed} shows a detailed overview of the parameter values used to generate the dataset.

\subsection{Sheet Metal Forming}
\label{sec:forming}
\textbfp{Problem Description}
The \emph{sheet metal forming} process is a critical manufacturing operation widely used across industries such as automotive and aerospace. 
\acs{FEM} simulations are commonly employed to estimate critical quantities such as thinning, local plastic deformation and residual stress distribution \citep{tekkaya2000, ablat2017, Folle2024}.
The simulation setup consists of a symmetrical workpiece supported at the ends and center, a holder and a punch that deforms the sheet by applying a displacement ($U$ in \cref{fig:forming_sketch}).
The 2D simulation predicts the sheet’s elasto-plastic deformation, providing quantities such as stress or plastic strain distributions (shown in \cref{fig:forming_fields}).
An essential engineering metric used in practice is the transverse stress (xx-component) distribution along the vertical center cord (green line in \cref{fig:forming_fields}).

\textbf{Input parameters} include the deformed sheet length $l$, the sheet thickness $t$, friction coefficient $\mu$ and the radii of the holder, punch, and supports $r$.
\cref{tab:forming_params} in \cref{app:forming_detailed} provides the sampling ranges for data generation.

\subsection{Electric Motor Design}
\label{sec:motor}
\textbfp{Problem Description}
The \textit{electric motor design} dataset encompasses a structural \acs{FEM} simulation of a rotor in electric machinery, subjected to mechanical loading at burst speed.
It is motivated by the conflicting design objectives in rotor development \citep{gerlach2021mechanical, dorninger2021}.
The 2D simulation predicts stress and deformation responses due to assembly pressing forces and centrifugal loads, accounting for the rotor's topology, material properties, and rotation speed.
The custom metric measures the relative error in Mises stress along the cord shown in green \cref{fig:motor_overview}.

\textbf{Input Parameters} together with their variations and detailed technical drawings are omitted from the main body, as this case is more complex.
They are provided in \cref{fig:motor_technical_drawing} and \cref{tab:motor_params}, both in \cref{app:motor_detailed}.


\subsection{Heatsink Design}
\label{sec:heatsink}

\begin{figure*}[htbp]
    \centering
    \begin{subfigure}[t]{0.52\textwidth}
        \centering
        \includegraphics[width=0.95\linewidth]{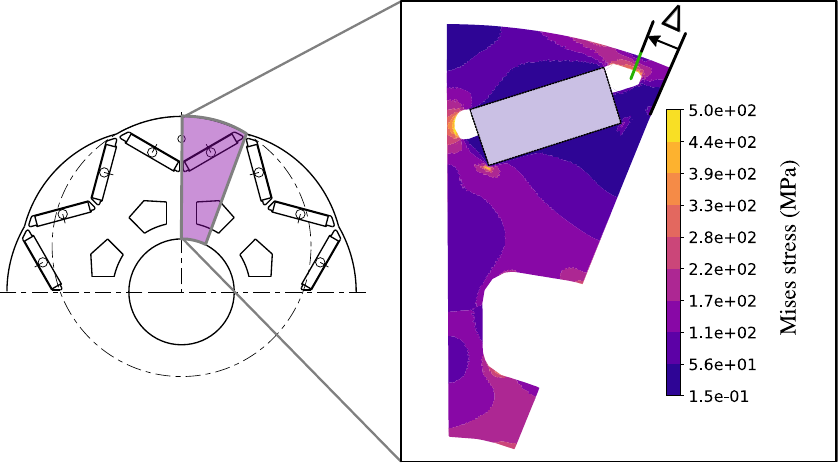}
        \caption{Electric motor design scenario (geometry and Mises stress field).}
        \label{fig:motor_overview}
    \end{subfigure}
    \hspace{2em}
    \begin{subfigure}[t]{0.35\textwidth}
        \centering
        \includegraphics[width=0.48\linewidth]{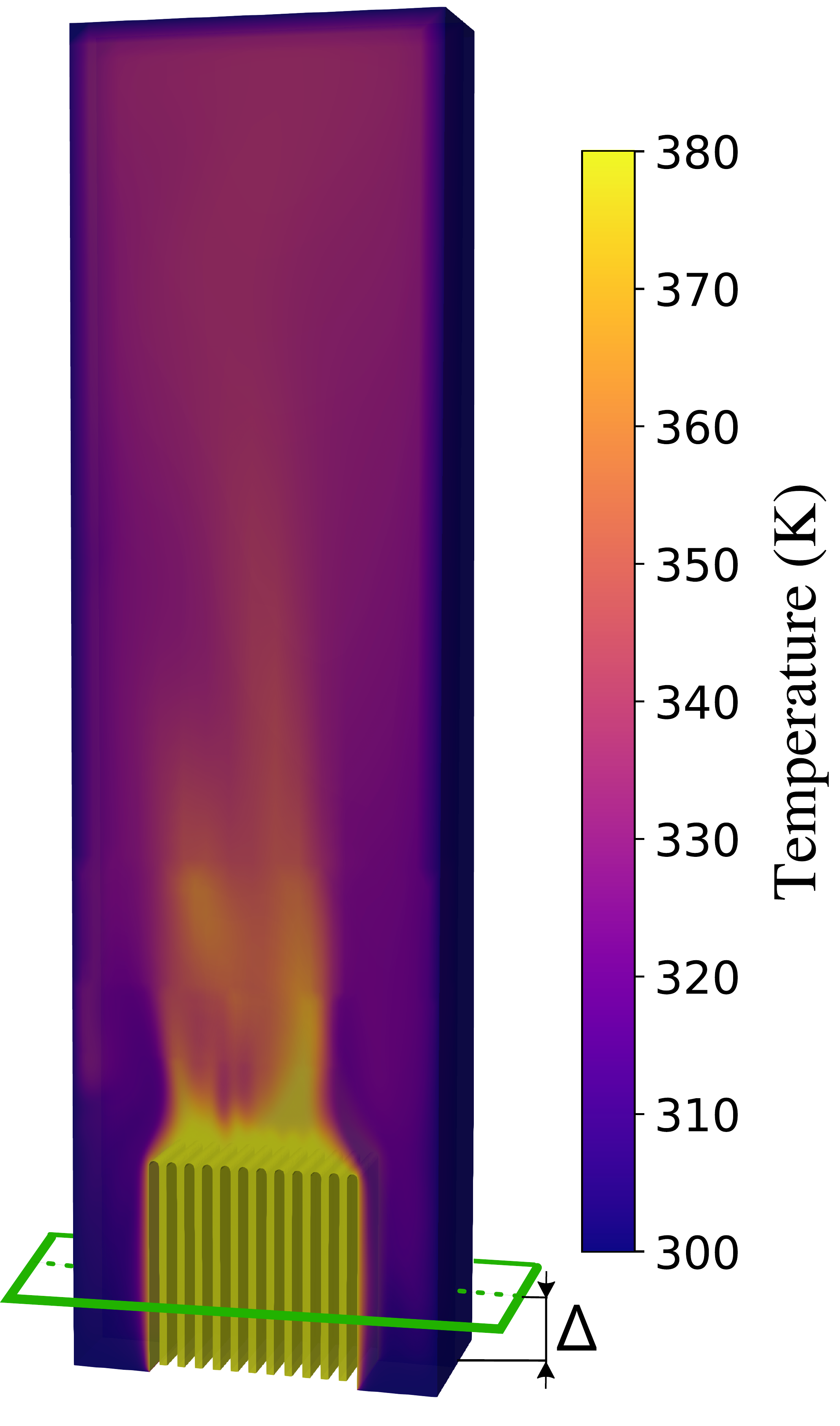}
        \caption{Heatsink: 3D temperature slice.}
        \label{fig:heatsink_overview}
    \end{subfigure}

    \caption{Overview of \emph{electric motor design} (left) and \emph{heatsink design} (right) simulation scenarios.}
    \label{fig:datasets_overview}
\end{figure*}

\textbf{Problem Description.}
The \textit{heatsink design} dataset represents a \ac{CFD} simulation focused on the thermal performance of heat sinks, commonly used in electronic cooling applications \citep{Arularasan2010, Rahman2024}.
It models the convective heat transfer from a heated base through an array of fins to the surrounding air.
The simulation captures how geometric fin characteristics along with the temperature of the heatsink affect the overall heat dissipation.
Outputs include steady state temperature, velocity and pressure fields, enabling the assessment of design efficiency and thermal resistance under varying configurations.
The main engineering metric measures the relative error in the temperature distribution along the dashed green line in \cref{fig:heatsink_overview}.

\textbf{Input Parameters} and their variations as well as an overview of the setup are provided in \cref{app:heatsink_detailed}.

\subsection{Distribution Shifts}
\label{sec:domain_splits}
\begin{figure}[!b]
    \centering
    \includegraphics[width=\columnwidth]{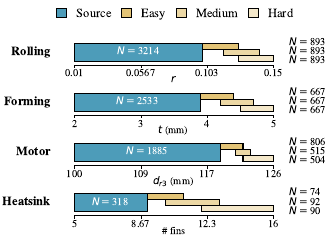}
    \caption{\ourmethod's predefined distribution shifts. $N$ denotes the number of samples in the respective domain.}
    \label{fig:domain_splits}
\end{figure}

\ourmethod's functionality allows for generating arbitrary n-dimensional parametric shifts for each problem, ensuring flexibility and extensibility.
For benchmarking, each dataset includes three predefined distribution shifts: \textit{easy}, \textit{medium} and \textit{hard}, which reflect increasing distributional distance in the respective input and output spaces.
The source and target domains are constructed by shifting along the dominant input parameter of each simulation scenario, as suggested by domain experts.
A schematic overview of the shifts is given in \cref{fig:domain_splits}.

To validate the design of our domain shifts we perform the following experiments:
\begin{enumerate*}[label=(\roman*)]
  \item \textbf{Latent space inspection:}
  We train models across the full input parameter ranges and perform a cluster analysis of their latent representations as the input conditions are varied.
  The resulting clusters consistently align with the parameters proposed by the experts, indicating that the chosen parameters dominate latent space variation (see visualizations in \crefrange{fig:tsne_rolling}{fig:tsne_heatsink}, \cref{app:tsne_plots}).
  \item \textbf{Transfer difficulty validation:}
  Differences in scalar input parameters  alone can be misleading regarding the actual shift difficulty experienced by models.
  We therefore complement them with two additional analyses.
  First, we report the \ac{PAD} computed directly in the space of ground truth simulation output fields.
  This way we aim to quantify the distributional divergence between source and target outputs, serving as a stronger proxy for the expected transfer gap.
  \cref{tab:pad_values} displays the \ac{PAD} values for our datasets across the different shift difficulties.
  Details on the estimation of these distances are provided in \cref{app:PAD}.
  Additionally, we provide the scaling behavior of model error across difficulties on one of our presented datasets in \cref{fig:error_scaling} and discuss it in \cref{sec:results}.
\end{enumerate*}

\begin{table}[ht]
\renewcommand{\arraystretch}{1.0}
\caption{PAD values across difficulty levels.}
\label{tab:pad_values}
\centering
\resizebox{0.32\textwidth}{!}{
\begin{tabular}{lccc}
\toprule
\textbf{Dataset} & \textbf{Easy} & \textbf{Medium} & \textbf{Hard} \\
\midrule
Rolling & 1.470 & 1.765 & 1.787 \\
Forming & 0.622 & 0.929 & 1.123 \\
Electric Motor & 0.211 & 0.314 & 0.421 \\
Heatsink & 0.765 & 1.767 & 1.995 \\
\bottomrule
\end{tabular}
}
\end{table}

Beyond the predefined one-dimensional splits, we explore higher-dimensional distribution shifts. 
In \cref{app:2d_shifts}, we demonstrate that models, adaptation algorithms and model selection strategies exhibit consistent behavior under a two-dimensional shift.

\section{Benchmark Setup}
\label{sec:benchmarking_setup}
This section outlines the learning problem (\cref{sec:learning_problem}), the \ac{UDA} algorithms considered (\cref{sec:uda_algorithms}), the unsupervised model selection strategies (\cref{sec:model_selection}), and the baseline models used (\cref{sec:models}). Finally, we describe the experimental setup and evaluation metrics in \cref{sec:experimental_setup}.

\subsection{Learning Problem}
\label{sec:learning_problem}

Let $\mathcal{X}$ be an input space containing geometries and conditioning parameters (e.g., thickness and temperatures in \cref{fig:rolling_sketch}) and $\mathcal{Y}$ be an output space containing ground truth solution fields, obtained from a numerical solver (e.g., \ac{PEEQ} field in \cref{fig:rolling_fields}).
Following~\citep{ben-david2010learningfromdifferentdomains}, a \textit{domain} is represented by a probability density function $p$ on $\mathcal{X}\times\mathcal{Y}$ (e.g., describing the probability of observing an input-output pair corresponding to the parameter range $r\in [0.01,0.125)$ in \cref{fig:domain_splits}).
\ac{UDA} has been formulated as follows:
Given a source dataset $(x_1, y_1),..., (x_n, y_n)$ drawn from a source domain $p_S$ together with an \textit{unlabeled} target dataset $x_{1}',..., x_{m}'$ drawn from the ($\mathcal{X}$-marginal) of a target domain $p_T$, the problem is to find a model $f: \mathcal{X} \to \mathcal{Y}$ that has small expected risk on the target domain:
\begin{equation}
\mathbb{E}_{(x, y) \sim p_T}\! \left[ \ell(f(x),y) \right]
\end{equation}
with $\ell:\mathcal{Y}\times\mathcal{Y}\to\mathbb{R}$ being some loss function.
In our setup $f(x) = g(x,\phi(x))$ is composed of a conditioning network $\phi$ and a surrogate $g$ (see \cref{fig:figure_1}).

\subsection{Unsupervised Domain Adaptation Algorithms}
\label{sec:uda_algorithms}
Our UDA baseline algorithms are from the class of \textit{domain-invariant representation learning} methods. These methods are strong baselines, in the sense that their performance typically lies within the standard deviation of the winning algorithms in large scale empirical evaluations (i.e., no significant outperformance is observed), see CMD, Deep CORAL and DANN in~\citep[Tables~12--14]{dinu2023iwa}, ~M3SDA in~\citep{peng2019moment}, MMDA and HoMM in~\citep{ragab2022adatime}.

Following~\citet{johansson2019support} and \citet{zellinger2021balancing}, we express the objective of domain-invariant learning using two learning models: a \textit{representation} mapping $\phi\in\Phi\subset\left\{\phi:\mathcal{X}\to\mathcal{R}\right\}$, which in our case corresponds to the conditioning network that maps simulation parameters into some representation space $\mathcal{R}\subset \mathbb{R}^k$ and a \textit{regressor} $g\in\mathcal{G}\subset\left\{g:\mathcal{X}\times\mathcal{R}\to\mathcal{Y}\right\}$, which is realized by a neural surrogate.
The goal is to find a mapping $\phi$ under which the source representations $\phi(\mathbf{x}):=(\phi(x_1),\ldots,\phi(x_n))$ and the target representations $\phi(\mathbf{x}'):=(\phi(x_1'),\ldots,\phi(x_m'))$ appear similar, and, at the same time, enough information is preserved for prediction by $g$, see~\citep{quionero-candela2009datasetshift}. This is realized by estimating objectives of the form
%
%
\begin{equation}
\label{eq:principled_algorithm_problem_statement}
\begin{aligned}
\min_{g\in\mathcal{G},\, \phi\in\Phi}\;
&\underbrace{
  \mathbb{E}_{(x, y) \sim p_S}
  \left[ \ell\big(g(x,\phi(x)),y\big) \right]
}_{\mathcal{L}_{\text{recon}}} \\
&\quad +\;
\lambda\cdot
\underbrace{
  d\big(\phi(\mathbf{x}),\phi(\mathbf{x}')\big)
}_{\mathcal{L}_{\text{DA}}} \, .
\end{aligned}
\end{equation}
The training objective therefore consists of minimizing both terms: the reconstruction loss $\mathcal{L}_{\text{recon}}$ and the domain adaptation loss $\mathcal{L}_{\text{DA}}$, as shown in \cref{fig:figure_1}.
A variety of \ac{UDA} algorithms correspond to different implementations of the distance $d$. 
Good choices have been found to be the Wasserstein distance~\citep{courty2017optimal}, the Maximum Mean Discrepancy~\citep{baktashmotlagh2013unsupervised}, moment distances~\citep{coral,cmd}, adversarially learned distances~\citep{ganin2015dann}
and other divergence measures~\citep{johansson2019support}.

\subsection{Unsupervised Model Selection Strategies}
\label{sec:model_selection}

Appropriately choosing the regularization parameter $\lambda$ is crucial \citep{musgrave2021realitycheck,dinu2023iwa,yang2024can}, with sub-optimal choices potentially leading to \textit{negative transfer}~\citep{pan2010transferlearningsurvey}.
However, classical approaches (e.g., validation set, cross-validation, information criterion) cannot be used due to missing labels and distribution shifts.
It is therefore a natural benchmark requirement to jointly evaluate \ac{UDA} algorithms and model selection.

In this work, we rely on \ac{IWV} \citep{sugiyama2007iwv} and \ac{DEV} \citep{you2019dev} to overcome the two challenges:
\begin{enumerate*}[label=(\roman*)]
    \item distribution shift and
    \item missing target labels.
\end{enumerate*}
These methods rely on the Radon-Nikod\'ym derivative and the covariate shift assumption $p_S(y|x)=p_T(y|x)$ to obtain
\begin{equation}
\label{eq:improtance_weighting}
    \mathbb{E}_{(x, y) \sim p_T}\! \left[ \ell(f(x),y) \right]
    = \mathbb{E}_{(x, y) \sim p_S}\! \left[ \beta(x)\ell(f(x),y) \right].
\end{equation}





\cref{eq:improtance_weighting} motivates to estimate the target error by a two step procedure: First, approaching challenge 
\begin{enumerate*}[label=(\roman*)]
    \item by estimating the density ratio $\beta(x)=\frac{p_T(x)}{p_S(x)}$ from the input data only, and, approaching challenge
    \item by estimating target error by the weighted source error using \textit{labeled} source data.
\end{enumerate*}

\subsection{Baseline Models}
\label{sec:models}
We provide a range of machine learning methods, adapted to our conditioned simulation task, organized by their capacity to model interactions across different scales:

\emph{Global context models} such as PointNet \citep{qi2017pointnet} incorporate global information into local Multi-Layer Perceptrons (MLPs) by summarizing features of all input points by aggregation into a global representation, which is then shared among nodes.
Recognizing the necessity of \emph{local information} when dealing with complex meshes and structures, we include GraphSAGE \citep{graphsage}, a proven \ac{GNN} architecture \citep{scarselli2009graph, battaglia2018relational}.
However, large scale applications of \acp{GNN} are challenging due to computational expense \citep{alkin2024upt} and issues like oversmoothing \citep{oversmoothing}. 
Finally, to overcome these limitations, we employ \emph{attention based models} \citep{attention}.
These models typically scale better with the number of points, and integrate both global and local information enabling stronger long-range interactions and greater expressivity.
We include Transolver \citep{wu2024Transolver}, a modern neural operator Transformer.


As an alternative categorization, baselines can also be classified by input-output pairings, into \textit{point-to-point} and \textit{latent} approaches.
The former explicitly encodes nodes, while the latter represents the underlying fields in a latent space, and optionally requires queries to retrieve nodes.
While all previously mentioned models are \textit{point-to-point}, we also include \ac{UPT} \citep{alkin2024upt, furst2025upt} and \ac{GINO} \cite{li2023gino}, as examples of latent field methods.
Both methods are designed for large problems, offering favorable scaling on big meshes through latent field modeling. Therefore, we benchmark them on the \textit{heatsink design} dataset.
The main difference is that \ac{GINO}'s latent space is constrained to a regular grid, where it operates in the \textit{frequency} domain. \ac{UPT}, in contrast, learns in an unconstrained latent space.
We provide implementational details of all architectures in \cref{app:model_architectures}.


Our framework explicitly conditions neural operators on configuration parameters.
We first embed them using a sin-cos encoding and a shallow MLP $\phi$ to produce a latent representation and then condition the neural operator $g$ by using either concatenation of the latent conditioning vector, or modulation through FiLM \citep{perez2018film} or DiT \citep{peebles2023scalable}.
As an ablation, we also evaluate replacing $\phi$ with a mesh encoder that derives the representation directly from the input geometry.
On the \textit{electric motor design} dataset, this variant performs worse (see \cref{app:geometric_pointnet}), supporting our design choice.

\subsection{Experiments and Evaluation}
\label{sec:experimental_setup}
\textbfp{Experimental Setup}
We benchmark four prominent \ac{UDA} algorithms (Deep Coral \citep{coral}, CMD \citep{cmd}, DANN \citep{ganin2015dann} and DARE-GRAM \citep{nejjar2023dare_gram}) in combination with the following four unsupervised model selection strategies: \ac{IWV} \citep{sugiyama2007iwv}, ~\mbox{\ac{DEV}~\citep{you2019dev}}, Source Best (SB) (selecting models based on source domain validation performance) and Target Best (TB) (selecting models based on target labels, which are not available in \ac{UDA} but serves as a lower bound for model selection).

Concerning neural surrogates, we evaluate PointNet, GraphSAGE, and Transolver on the \textit{hot rolling}, \textit{sheet metal forming}, and \textit{electric motor design} datasets.
Due to memory and runtime constraints on the large scale \textit{heatsink design} dataset, we omit GraphSAGE and instead benchmark UPT and GINO alongside PointNet and Transolver.

\textbf{Experimental Scale.}
We perform an extensive sweep over the critical \ac{UDA} parameter $\lambda$ and average across four seeds, resulting in a total of $\mathbf{1,\!508}$ training runs (see \cref{tab:lambda_choices}).
Details on architectures, hyperparameters, training setup and normalization, as well as a breakdown of training times are included in \cref{app:experiments,app:model_architectures}.


\textbfp{Evaluation Metrics}
For each dataset, we report the \ac{nRMSE} averaged over all output fields, as well as the per field \ac{RMSE} values (on denormalized data), the Euclidean error for deformation predictions and the custom error metrics described in \crefrange{sec:rolling}{sec:heatsink}.
Additionally we provide physics-based evaluation metrics for all datasets.
These metrics are tailored to the underlying PDEs.
Detailed metric definitions are provided in \cref{app:evaluation_metrics}.

\section{Benchmarking Results}
\label{sec:results}

\begin{table*}[t]
  \caption{Best pair of \ac{UDA} algorithm and unsupervised model selection method for each architecture on all datasets (at \emph{medium} difficulty).
  We also report an Oracle with Target Best (TB) selection, serving as a lower bound for unsupervised model selection.
  Metrics shown are \ac{nRMSE} across all fields, the custom metrics for each dataset, as described in \crefrange{sec:rolling}{sec:heatsink} and
  physics-based metrics (Rolling: von mises consistency, Forming: plastic law residual, Motor: Hooke's law consistency and Heatsink: boundary condition consistency in the velocity field).
  Improvements over the unregularized baseline are shown as negative values in parentheses.
  \tightbox{bestrow}{Green} highlights the best configuration per dataset (chosen by lowest target \ac{nRMSE} across all fields) and the best entry per metric for each dataset is \textbf{bold}.}
  \label{tab:results}
  \centering
  \begin{subtable}{\textwidth}
  \centering
  \resizebox{\textwidth}{!}{
  \begin{tabular}{lllcccc}
    \toprule
    \multirow{2}{*}{\textbf{Dataset}} & \multirow{2}{*}{\textbf{Model}} & \textbf{Best UDA Method} & \multicolumn{2}{c}{\textbf{All Fields Normalized Avg}} & \textbf{Custom Metric} &\textbf{Physics Metric} \\
    & & \textbf{+ Model Selection} & \textbf{Source} & \textbf{Target} & \textbf{Target}& \textbf{Target}\\
    \midrule
    \multirow{4}{*}{Rolling} & \cellcolor{bestrow}\textbf{GraphSAGE} & \cellcolor{bestrow}\textbf{Deep Coral + SB} & \cellcolor{bestrow}\textbf{0.020 (-0.004)} & \cellcolor{bestrow}\textbf{0.199 (-0.062)} & \cellcolor{bestrow}\textbf{0.101 (-0.016)} & \cellcolor{bestrow}\textbf{0.047 (-0.004)} \\
     & PointNet & DANN + IWV & 0.029 (-0.006) & 0.237 (-0.068) & 0.135 (-0.004) & 0.058 (--------) \\
     & Transolver & Deep Coral + IWV & 0.023 (-0.005) & 0.679 (-0.260) & 0.374 (-0.059) & 0.114 (-0.002) \\
     & Oracle (GraphSAGE) & DANN + TB & 0.020 (-0.005) & 0.195 (-0.066) & 0.094 (-0.022) & 0.046 (-0.004) \\
    \midrule
    \multirow{4}{*}{Forming} & GraphSAGE & Deep Coral + DEV & 0.062 (-0.017) & 0.149 (-0.032) & 0.705 (+0.530) & \textbf{0.491 (+0.008)} \\
     & PointNet & Deep Coral + SB & 0.081 (-0.023) & 0.112 (-0.040) & 0.379 (+0.210) & 0.512 (-0.020) \\
     & \cellcolor{bestrow}\textbf{Transolver} & \cellcolor{bestrow}\textbf{Deep Coral + DEV} & \cellcolor{bestrow}\textbf{0.053 (-0.014)} & \cellcolor{bestrow}\textbf{0.055 (-0.017)} & \cellcolor{bestrow}\textbf{0.132 (+0.100)} & \cellcolor{bestrow}0.500 (-0.056) \\
     & Oracle (Transolver) & Deep Coral + TB & 0.053 (-0.015) & 0.053 (-0.019) & 0.138 (+0.106) & 0.515 (-0.041) \\
    \midrule
    \multirow{4}{*}{Motor} & GraphSAGE & Deep Coral + SB & 0.186 (-0.048) & 0.237 (-0.071) & 0.279 (-0.027) & 0.004 (--------) \\
     & PointNet & Deep Coral + SB & 0.146 (-0.042) & 0.208 (-0.058) & 0.136 (-0.040) & 0.003 (-0.001) \\
     & \cellcolor{bestrow}\textbf{Transolver} & \cellcolor{bestrow}\textbf{Deep Coral + SB} & \cellcolor{bestrow}\textbf{0.044 (-0.015)} & \cellcolor{bestrow}\textbf{0.043 (-0.014)} & \cellcolor{bestrow}\textbf{0.040 (-0.004)} & \cellcolor{bestrow}\textbf{0.003 (--------)} \\
     & Oracle (Transolver) & Deep Coral + TB & 0.042 (-0.016) & 0.043 (-0.015) & 0.039 (-0.005) & 0.003 (--------) \\
    \midrule
    \multirow{5}{*}{Heatsink} & \cellcolor{bestrow}\textbf{PointNet} & \cellcolor{bestrow}\textbf{DANN + SB} & \cellcolor{bestrow}0.315 (-0.113) & \cellcolor{bestrow}\textbf{0.365 (-0.125)} & \cellcolor{bestrow}0.030 (+0.007) & \cellcolor{bestrow}0.078 (-0.004) \\
     & Transolver & Dare Gram + IWV & \textbf{0.181 (-0.044)} & 0.400 (-0.160) & \textbf{0.019 (+0.004)} & 0.080 (-0.002) \\
     & UPT & Dare Gram + IWV & 0.184 (-0.062) & 0.406 (-0.150) & 0.023 (--------) & \textbf{0.070 (-0.007)} \\
     & ConditionedGINO & Deep Coral Accumulating + DEV & 0.237 (-0.081) & 0.422 (-0.142) & 0.022 (-0.001) & 0.091 (+0.004) \\
     & Oracle (PointNet) & DANN + TB & 0.321 (-0.107) & 0.348 (-0.142) & 0.033 (+0.010) & 0.073 (-0.008) \\
    \bottomrule
  \end{tabular}
  }
  \end{subtable}
\end{table*}

\cref{tab:results} summarizes our benchmarking results, showing the best \ac{UDA} algorithm and model selection method combination per model for each dataset.
Inspecting the \ac{nRMSE} across all fields, all adapted models exhibit lower target errors than their unregularized counterparts across all datasets.
Notably, the impact of adaptation varies by dataset: the source-target gap remains significant for the \emph{hot rolling} and \emph{heatsink} datasets, whereas for the \emph{sheet metal forming} and \emph{electric motor design} datasets, the best methods achieve target errors comparable to their source performance.
This finding aligns with the \ac{PAD} magnitudes for the medium difficulties reported in \cref{tab:pad_values}, where higher values correspond to larger source-target performance gaps of models.

However, gains are not uniform across all metrics.
While the global \ac{nRMSE} generally improves with adaptation, we observe cases where performance on our custom or physics-based metrics degrades.
For example, in the \emph{sheet metal forming} task, the best performing model (Transolver) achieves a reduction in \ac{nRMSE} but exhibits increased errors in the custom metric. 
Despite the clear benefits of \ac{UDA}, no single algorithm or model selection strategy consistently outperforms the others across datasets or architectures.
Furthermore, the gaps between the best combinations and the TB Oracle (lower bound on target error) highlight the room for improvement in unsupervised model selection.
In addition to this summary, we report detailed metrics and visualizations across architectures, algorithms, and selection strategies in \cref{app:detailed_results}.

\begin{figure}[!b]
    \centering
    \includegraphics[width=0.85\linewidth]{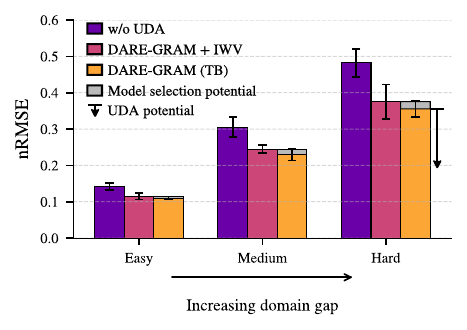}
    \caption{Target error scaling with increasing domain gap. We show the averaged nRMSE across all fields for the \emph{easy}, \emph{medium}, and \emph{hard} shifts on the \emph{hot rolling} task.
    We compare PointNet models without \ac{UDA} to DARE-GRAM combined with IWV and TB selection (lower bound).
    Error bars indicate the standard deviation across four seeds.
    Furthermore, we highlight potentials of \ac{UDA} algorithm and model selection improvements.}
    \label{fig:error_scaling}
\end{figure}

Finally, since the main results focus on the \emph{medium} difficulty setting, we additionally visualize error scaling across all shift difficulties for the \emph{hot rolling} dataset in \cref{fig:error_scaling}.
It illustrates the increase in target error of PointNet as the domain gap widens and highlights the consistent improvements achieved by applying \ac{UDA} algorithms with unsupervised model selection strategies on the \emph{easy}, \emph{medium} and \emph{hard} settings.
This analysis again aligns with the increasing \ac{PAD} values with increasing shift difficulty in \cref{tab:pad_values}.
Overall, \cref{fig:error_scaling} highlights the two promising directions for further research:
\begin{enumerate*}[label=(\roman*)]
  \item enhancement of neural surrogate architectures and \ac{UDA} algorithms, and
  \item especially, improvement of unsupervised model selection strategies.
\end{enumerate*}

\section{Discussion}
\label{sec:discussion}

We presented \ourmethod, a collection of industry relevant datasets paired with a benchmarking library for comparing \ac{UDA} algorithms, model selection strategies and neural surrogates in real world scenarios.
We adapted available techniques, applied them on physical simulation data and performed extensive experiments to evaluate their performance on the presented datasets.
Our findings suggest that standard \ac{UDA} training methods can improve performance of models in unseen parameter ranges in physical simulations, with improvement margins in line with those seen in \ac{UDA} literature \citep{dinu2023iwa, ragab2022adatime}.
Additionally, we find unsupervised model selection to be extremely important in downstream target performance, with it arguably having as much impact as the \ac{UDA} training itself, which is also in agreement with other \ac{DA} works \citep{musgrave2021realitycheck}.

\textbf{Limitations.} We acknowledge that our datasets are limited under two main aspects: 
\begin{enumerate*}[label=(\roman*)]
  \item They only cover \emph{steady-state} problems, which represent a large portion of industrial simulation tasks. However, an extension with \emph{time-dependent} datasets could be valuable for certain application areas.
  \item They cover a wide range of mesh sizes, ranging from roughly \(\mathcal{O}(10^2)\) up to \(\mathcal{O}(10^6)\) nodes. Nevertheless, many industrial scenarios require substantially larger meshes.
\end{enumerate*}
These limitations reflect design choices aimed at benchmarking clarity and computational feasibility and leave room for future extensions.

\textbf{Future Directions.}
Motivated by our results, we identify several promising research directions:
\begin{enumerate*}[label=(\roman*)]
  \item Although we include a diverse and competitive set of \ac{UDA} algorithms and unsupervised model selection techniques, a wide range of methods remain unexplored in the context of scientific ML.
  Examples include ensembling based adaptation \citep{cha2021swad}, adversarial information bottleneck approaches \citep{luo2019bottleneck,song2020bottleneck} or diffusion based methods \citep{peng2024udadiff,liao2025udadenoising}.
  In addition, test-time adaptation methods \citep{wang2021tent,adachi2025ssa} could be designed and tested using our benchmark.
  \item \ourmethod~currently evaluates standard \ac{UDA} algorithms and does not integrate physics constraints \citep{Karniadakis2021} into training.
  Our framework and datasets allows for physics constraints, and we find the direction of a specific physics-inspired \ac{UDA} method an interesting and potentially fruitful gap in the current research.
\end{enumerate*}

\section*{Impact Statement}
This paper presents work whose goal is to advance the field of Machine Learning, in particular applied to surrogate models of numerical physics simulations.
There are some potential business and engineering consequences, none of which we recognize could impact society in a way we feel must be specifically highlighted here.

\section*{Acknowledgements}
The authors thank Benedikt Alkin for his support on our UPT implementation as well as
Wei Lin for the discussions and feedback on the work.
Furthermore, we thank Judith Resch, Simon Weitzhofer, Jagoba Lekue, and Barbara Hartl for implementing the numerical models for the benchmark dataset.
This work has been supported by the COMET-K2 Center of the Linz Center of Mechatronics (LCM).

The ELLIS Unit Linz, the LIT AI Lab, the Institute for Machine Learning, are supported by
the Federal State Upper Austria.
We thank the projects FWF AIRI FG 9-N (10.55776/FG9), AI4GreenHeatingGrids (FFG- 899943), Stars4Waters (HORIZON-CL6-2021-CLIMATE-01-01), FWF Bilateral Artificial Intelligence (10.55776/COE12).
We thank NXAI GmbH, Silicon Austria Labs (SAL), Merck Healthcare KGaA, GLS (Univ. Waterloo), TÜV Holding GmbH, Software Competence Center Hagenberg GmbH, dSPACE GmbH, TRUMPF SE + Co. KG.
We acknowledge EuroHPC Joint Undertaking for access to Leonardo at CINECA, Italy, and MareNostrum5 at BSC, Spain.

\bibliography{references.bib}
\bibliographystyle{icml2026}

\newpage
\appendix
\onecolumn

\section*{LLM Usage Disclosure}
In general, LLM tools were used to refine writing in parts of the paper.
Gemini 3 and GPT-5.2 were additionally used to make visualizations prettier, speed up the development of plotting functions, and dump experimental results neatly into latex tables.
AI assistants were strictly editors and decorators, i.e. they were not involved in ideation, reordering ideas, or at any higher or lower conceptual level.

\section{On neural operators}
\label{app:neural_operators}
One prominent approach in neural surrogate modeling for \ac{PDE}s is operator learning~\citep{kovachki2021neuraloperator, li2020fno, lu2021deeponet, alkin2024upt, li2020gkn}. In this setting, an operator maps input functions, such as boundary or initial conditions, to the corresponding solution of the \ac{PDE}.
During training, neural operators typically learn from input-output pairs of discretized functions \citep{kovachki2021neuraloperator, li2020fno, lu2021deeponet, alkin2024upt}.
While some methods expect regular, grid based inputs \citep{li2020fno}, others can be applied to any kind of data structure \citep{alkin2024upt, li2020gkn, li2023gino}.
One notable property is \emph{discretization invariance}, which, along with the ability to handle irregular data, enables generalization across different resolutions and mesh geometries.
This is a highly desirable property for industrial simulations \citep{meshgraphnet, alkin2024upt, furst2025upt, li2023geometryinformed, Franco2022MeshInformedNN}, where non-uniform meshes are the standard due to the computational and modeling advantages.
In this work, we focus on domain adaptation rather than benchmarking discretization invariance, and include neural surrogates that may not satisfy this property, such as~\citep{graphsage}.
Such models have been leveraged in several large scale industrial contexts, including \ac{CFD} for automotive \citep{bleeker2025neuralcfd} or \ac{DEM} simulations for industrial processes \citep{alkin2024neuraldem}.

\newpage
\section{Detailed results}
\label{app:detailed_results}
Complementing the summary in \cref{tab:results} of the main paper, the following subsections present detailed results for each dataset.
For every dataset, we present a complete empirical evaluation of our benchmark that compares the performance for all combinations of models, \ac{UDA} algorithms and model selection strategies across all output fields and metrics.

While these quantitative metrics offer a high level summary of model performance, industry practitioners often need a more fine grained picture to assess the neural surrogate's capabilities under distribution shifts.
To address this, we include additional analyses and visualizations alongside the quantitative results.
First, we provide error distribution histograms to better illustrate the difficulty of the domain shift occurring in each dataset.
Additionally, we present fringe plots comparing model predictions with the respective ground truth numerical solutions.

\subsection{Hot Rolling}

\cref{tab:rolling_results} presents the complete benchmarking results for the \emph{hot rolling} dataset.

\begin{table}[h]
  \centering
    \caption{Performance metrics on the source and target domains (mean $\pm$ std over 4 seeds) for the \emph{hot rolling} task at \textit{medium} difficulty. RMSE is reported for all global metrics (All Fields Normalized Avg through Equivalent Plastic Strain), while further columns report custom engineering and physics-based metrics. \textbf{Bold} indicates the overall best combination of architecture, \ac{UDA} algorithm, and model selection. Within each architecture group, the unregularized baseline is shaded \tightbox{baselinerow}{beige}, and the best \ac{UDA} configuration is \underline{underlined} and shaded \tightbox{bestrow}{green}.}  \label{tab:rolling_results}
  \resizebox{\textwidth}{!}{%
  \definecolor{bestrow}{HTML}{DFF0D8}
  \definecolor{baselinerow}{HTML}{FFF4CC}
\begin{tabular}{lllcccccccccccccccccc}
    \toprule
    \multirow{2}{*}{\textbf{Model}} & \multirow{2}{*}{\makecell{\textbf{DA}\\ \textbf{Algorithm}}} & \multirow{2}{*}{\makecell{\textbf{Model}\\ \textbf{Selection}}} & \multicolumn{2}{c}{\textbf{All Fields Normalized Avg (-)}} & \multicolumn{2}{c}{\textbf{Deformation (mm)}} & \multicolumn{2}{c}{\textbf{Cauchy Stress (MPa)}} & \multicolumn{2}{c}{\textbf{Mises Stress (MPa)}} & \multicolumn{2}{c}{\textbf{Logarithmic Strain ($\mathbf{\times 10^{-2}}$)}} & \multicolumn{2}{c}{\textbf{Equivalent Plastic Strain ($\mathbf{\times 10^{-2}}$)}} & \multicolumn{2}{c}{\textbf{Temperature (K)}} & \multicolumn{2}{c}{\textbf{Relative Custom Error (-)}} & \multicolumn{2}{c}{\textbf{VM Consistency (-)}} \\
\cmidrule(lr){4-5} \cmidrule(lr){6-7} \cmidrule(lr){8-9} \cmidrule(lr){10-11} \cmidrule(lr){12-13} \cmidrule(lr){14-15} \cmidrule(lr){16-17} \cmidrule(lr){18-19} \cmidrule(lr){20-21}
      &   &  & \textbf{SRC} & \textbf{TGT} & \textbf{SRC} & \textbf{TGT} & \textbf{SRC} & \textbf{TGT} & \textbf{SRC} & \textbf{TGT} & \textbf{SRC} & \textbf{TGT} & \textbf{SRC} & \textbf{TGT} & \textbf{SRC} & \textbf{TGT} & \textbf{SRC} & \textbf{TGT} & \textbf{SRC} & \textbf{TGT} \\
\bottomrule
  \end{tabular}
  }
\end{table}

\begin{figure*}[t]
    \centering
    \begin{minipage}[t]{0.48\linewidth} 
        \vspace{0pt}
        \centering
        \includegraphics[width=\linewidth]{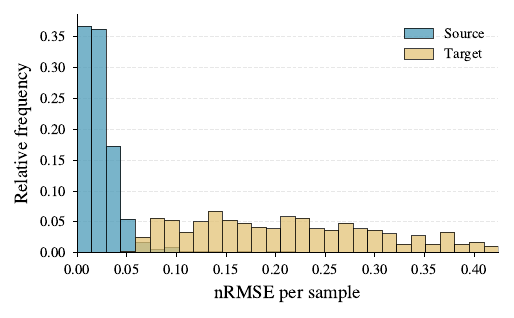}
        \caption{Distribution of \ac{nRMSE} (averaged across all fields) for the test sets of the source (blue) and target (yellow) domains.}
        \label{fig:error_dist_rolling}
    \end{minipage}
    \hfill
    \begin{minipage}[t]{0.48\linewidth}
        \vspace{0pt}
        \centering
        \captionof{table}{Absolute error of PEEQ predictions for representative samples from the source and target domain of the \emph{hot rolling} dataset. Lowest value per metric is bold.}
        \label{tab:rolling_error_table}
        \def\arraystretch{1.1}
        \setlength{\tabcolsep}{8pt}
          \begin{tabular}{lcc}
            \toprule
            \textbf{Metric} & \textbf{Source} & \textbf{Target} \\
            \midrule
            Mean            & \textbf{2.89e-04} & 8.08e-03 \\
            Std             & \textbf{2.37e-04} & 2.72e-03 \\
            Median          & \textbf{2.28e-04} & 8.57e-03 \\
            Q\textsubscript{01} & \textbf{5.42e-06} & 9.56e-04 \\
            Q\textsubscript{25} & \textbf{1.17e-04} & 6.79e-03 \\
            Q\textsubscript{75} & \textbf{4.00e-04} & 9.59e-03 \\
            Q\textsubscript{99} & \textbf{1.09e-03} & 1.52e-02 \\
            \bottomrule
          \end{tabular}
    \end{minipage}
\end{figure*}

To gain more insights, we conduct additional analyses on the best performing model, selected based on having the lowest average normalized target domain error across all fields.
\cref{fig:error_dist_rolling} shows the error distribution of this model and clearly highlights the substantial distribution shift between the source and target domain of the \emph{hot rolling} dataset.
Errors in the target domain are noticeably larger, almost up to an order of magnitude higher than those observed in the source domain.

To further illustrate the model's performance, we analyze two representative samples, one from the source and one from the target domain.
Since the most critical field for downstream applications is \ac{PEEQ}, we restrict the following analysis on this scalar field only.

\begin{figure}[htbp]
    \centering
    \includegraphics[width=\textwidth]{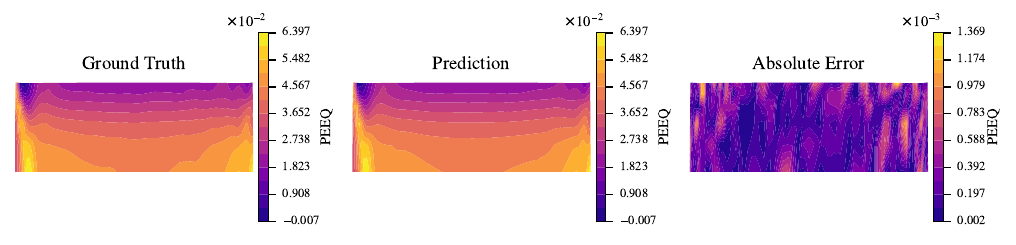}
    \caption{Fringe plot of the \emph{hot rolling} dataset (representative source sample). Shown is the ground truth (left) and predicted (middle) PEEQ, as well as the absolute error (right).}
    \label{fig:rolling_representative_source}
\end{figure}

\begin{figure}[htbp]
    \centering
    \includegraphics[width=\textwidth]{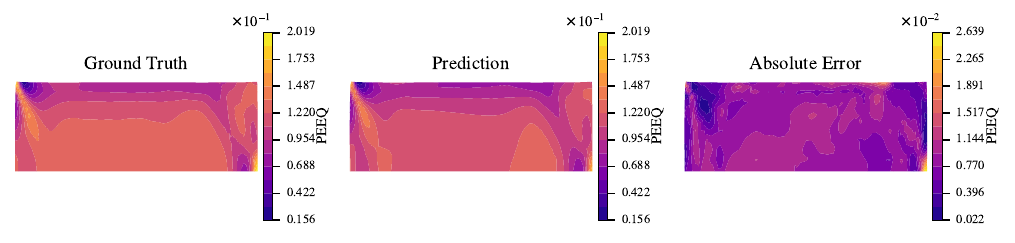}
    \caption{Fringe plot of the \emph{hot rolling} dataset (representative target sample). Shown is the ground truth (left) and predicted (middle) PEEQ, as well as the absolute error (right).}
    \label{fig:rolling_representative_target}
\end{figure}

\cref{tab:rolling_error_table} presents a summary of the absolute PEEQ prediction errors for the selected source and target samples, while \cref{fig:rolling_representative_source} and \cref{fig:rolling_representative_target} provide a visualization of the ground truth, predictions, and absolute errors for these samples using fringe plots.

\FloatBarrier
\newpage
\subsection{Sheet Metal Forming}

In contrast to the substantial shift observed in the hot rolling dataset, the distribution shift in the \emph{sheet metal forming} dataset is moderate.
\cref{tab:forming_results} presents the detailed performance across all models, algorithms, and selections for this dataset.

\begin{table}[h]
  \centering
    \caption{Performance metrics on the source and target domains (mean $\pm$ std over 4 seeds) for the \emph{sheet metal forming} task at \textit{medium} difficulty. RMSE is reported for all global metrics (All Fields Normalized Avg through Equivalent Plastic Strain), while further columns report custom engineering and physics-based metrics. \textbf{Bold} indicates the overall best combination of architecture, \ac{UDA} algorithm, and model selection. Within each architecture group, the unregularized baseline is shaded \tightbox{baselinerow}{beige}, and the best \ac{UDA} configuration is \underline{underlined} and shaded \tightbox{bestrow}{green}.}
  \label{tab:forming_results}
  \resizebox{\textwidth}{!}{%
  \definecolor{bestrow}{HTML}{DFF0D8}
  \definecolor{baselinerow}{HTML}{FFF4CC}
\begin{tabular}{lllcccccccccccccccccccc}
    \toprule
    \multirow{2}{*}{\textbf{Model}} & \multirow{2}{*}{\makecell{\textbf{DA}\\ \textbf{Algorithm}}} & \multirow{2}{*}{\makecell{\textbf{Model}\\ \textbf{Selection}}} & \multicolumn{2}{c}{\textbf{All Fields Normalized Avg (-)}} & \multicolumn{2}{c}{\textbf{Deformation (mm)}} & \multicolumn{2}{c}{\textbf{Stress (MPa)}} & \multicolumn{2}{c}{\textbf{Mises Stress (MPa)}} & \multicolumn{2}{c}{\textbf{Logarithmic Strain ($\mathbf{\times 10^{-2}}$)}} & \multicolumn{2}{c}{\textbf{Equivalent Plastic Strain ($\mathbf{\times 10^{-2}}$)}} & \multicolumn{2}{c}{\textbf{Rel Custom Error (-)}} & \multicolumn{2}{c}{\textbf{VM Consistency (-)}} & \multicolumn{2}{c}{\textbf{Violation Elastic ($\mathbf{\times 10^{-2}}$)}} & \multicolumn{2}{c}{\textbf{Residual Plastic (-)}} \\
\cmidrule(lr){4-5} \cmidrule(lr){6-7} \cmidrule(lr){8-9} \cmidrule(lr){10-11} \cmidrule(lr){12-13} \cmidrule(lr){14-15} \cmidrule(lr){16-17} \cmidrule(lr){18-19} \cmidrule(lr){20-21} \cmidrule(lr){22-23}
      &   &  & \textbf{SRC} & \textbf{TGT} & \textbf{SRC} & \textbf{TGT} & \textbf{SRC} & \textbf{TGT} & \textbf{SRC} & \textbf{TGT} & \textbf{SRC} & \textbf{TGT} & \textbf{SRC} & \textbf{TGT} & \textbf{SRC} & \textbf{TGT} & \textbf{SRC} & \textbf{TGT} & \textbf{SRC} & \textbf{TGT} & \textbf{SRC} & \textbf{TGT} \\
\bottomrule
  \end{tabular}
  }
\end{table}

\begin{figure*}[b]
    \centering
    \begin{minipage}[t]{0.48\linewidth} 
        \vspace{0pt}
        \centering
        \includegraphics[width=\linewidth]{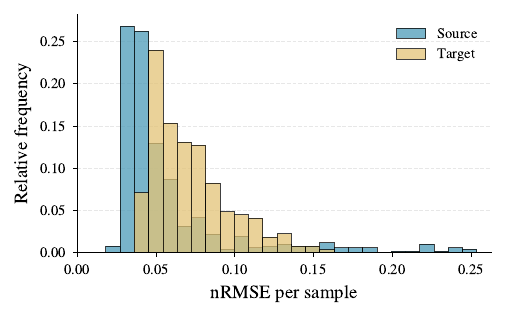}
        \caption{Distribution of \ac{nRMSE} (averaged across all fields) for the test sets of the source (blue) and target (yellow) domains.}
        \label{fig:error_dist_forming}
    \end{minipage}
    \hfill
    \begin{minipage}[t]{0.48\linewidth}
        \vspace{0pt}
        \centering
        \captionof{table}{Absolute error of PEEQ predictions for representative samples from the source and target domain of the \emph{sheet metal forming} dataset. Lowest value per metric is bold.}
        \label{tab:forming_best_worst}
        \def\arraystretch{1.1}
        \setlength{\tabcolsep}{8pt}
          \begin{tabular}{lcc}
            \toprule
            \textbf{Metric} & \textbf{Source} & \textbf{Target} \\
            \midrule
            Mean            & \textbf{2.09e-04} & 2.21e-04 \\
            Std             & \textbf{5.04e-04} & 5.35e-04 \\
            Median          & 5.64e-05 & \textbf{5.03e-05} \\
            Q\textsubscript{01} & 1.32e-06 & \textbf{9.38e-07} \\
            Q\textsubscript{25} & 2.88e-05 & \textbf{2.74e-05} \\
            Q\textsubscript{75} & 1.22e-04 & \textbf{1.10e-04} \\
            Q\textsubscript{99} & \textbf{2.65e-03} & 2.84e-03 \\
            \bottomrule
          \end{tabular}
    \end{minipage}
\end{figure*}

To further illustrate model behavior under distribution shift, we examine the best performing model, selected by lowest normalized average target domain error.
The error distribution (\cref{fig:error_dist_forming}) shows a moderate distribution shift between the source and target domain with some outliers in the source domain.

\begin{figure}[htbp]
    \centering
    \includegraphics[width=0.9\textwidth]{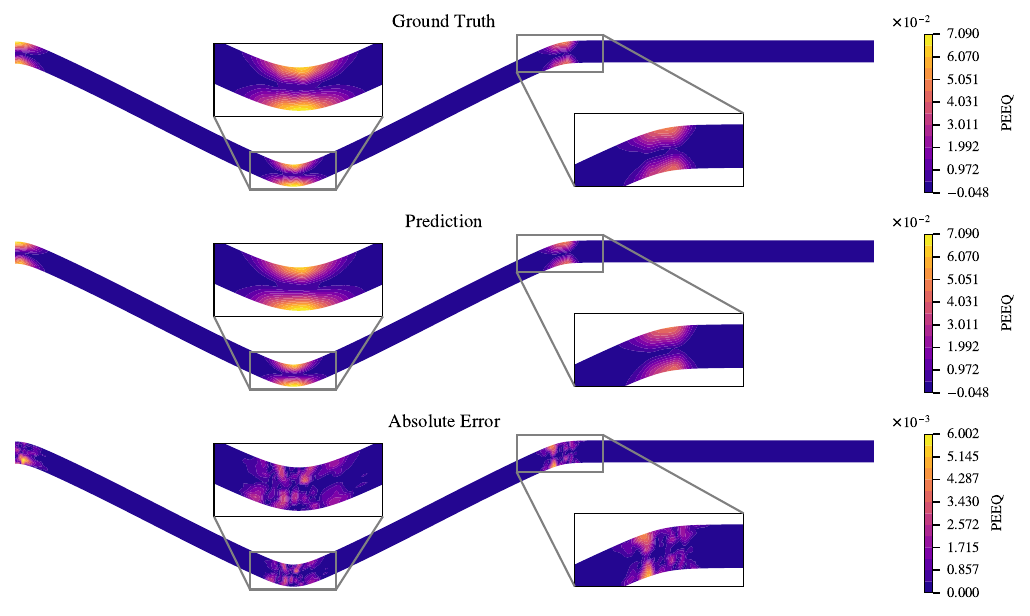}
    \caption{Fringe plot of the \emph{sheet metal forming} dataset (representative source sample). Shown is the ground truth (top) and predicted (middle) PEEQ, aswell as the absolute error (bottom).}
    \label{fig:forming_representative_source}
\end{figure}

\begin{figure}[htbp]
    \centering
    \includegraphics[width=0.9\textwidth]{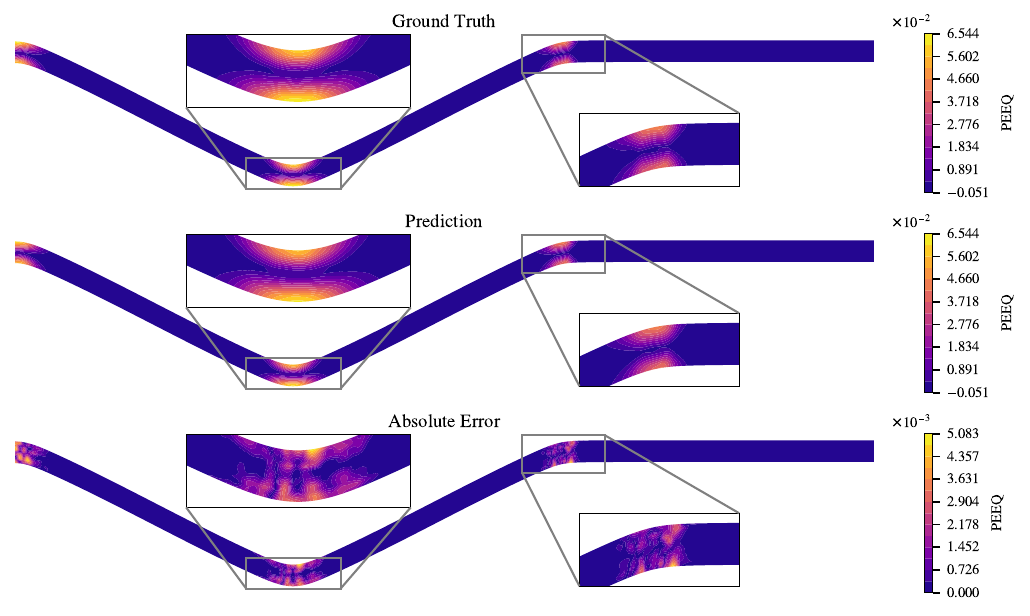}
    \caption{Fringe plot of the \emph{sheet metal forming} dataset (representative target sample). Shown is the ground truth (top) and predicted (middle) PEEQ, as well as the absolute error (bottom).}
    \label{fig:forming_representative_target}
\end{figure}

To better understand the model's predictive behavior in this setting, we analyze representative sample in each domain, again focusing on the critical \ac{PEEQ} field.
\cref{tab:forming_best_worst} provides a statistical summary of the absolute PEEQ prediction errors across the selected cases.
Fringe plots in \cref{fig:forming_representative_source,fig:forming_representative_target} provide a visual understanding of model accuracy.

\subsection{Electric Motor Design}

\cref{tab:motor_results} presents the complete benchmarking results for the \emph{electric motor design} dataset. For this dataset the relative degradation in model performance in the target domain is in general smaller than in the previous two presented above.

\begin{table}[h]
  \centering
    \caption{Performance metrics on the source and target domains (mean $\pm$ std over 4 seeds) for the \emph{electric motor design} task at \textit{medium} difficulty. RMSE is reported for all global metrics (All Fields Normalized Avg through Logarithmic Strain), while further columns report custom engineering and physics-based metrics. \textbf{Bold} indicates the overall best combination of architecture, \ac{UDA} algorithm, and model selection. Within each architecture group, the unregularized baseline is shaded \tightbox{baselinerow}{beige}, and the best \ac{UDA} configuration is \underline{underlined} and shaded \tightbox{bestrow}{green}.}
  \label{tab:motor_results}
  \resizebox{\textwidth}{!}{%
  \definecolor{bestrow}{HTML}{DFF0D8}
  \definecolor{baselinerow}{HTML}{FFF4CC}
  \begin{tabular}{lllcccccccccccc}
    \toprule
    \multirow{2}{*}{\textbf{Model}} & \multirow{2}{*}{\makecell{\textbf{DA}\\ \textbf{Algorithm}}} & \multirow{2}{*}{\makecell{\textbf{Model}\\ \textbf{Selection}}} & \multicolumn{2}{c}{\textbf{All Fields Normalized Avg (-)}} & \multicolumn{2}{c}{\textbf{Deformation (m)}} & \multicolumn{2}{c}{\textbf{Cauchy Stress (MPa)}} & \multicolumn{2}{c}{\textbf{Logarithmic Strain ($\mathbf{\times 10^{-2}}$)}} & \multicolumn{2}{c}{\textbf{Rel Custom Error (-)}} & \multicolumn{2}{c}{\textbf{Constitutive Error ($\mathbf{\times 10^{-2}}$)}} \\
\cmidrule(lr){4-5} \cmidrule(lr){6-7} \cmidrule(lr){8-9} \cmidrule(lr){10-11} \cmidrule(lr){12-13} \cmidrule(lr){14-15}
      &   &  & \textbf{SRC} & \textbf{TGT} & \textbf{SRC} & \textbf{TGT} & \textbf{SRC} & \textbf{TGT} & \textbf{SRC} & \textbf{TGT} & \textbf{SRC} & \textbf{TGT} & \textbf{SRC} & \textbf{TGT} \\
    \midrule
    \multirow{17}{*}{GraphSAGE} & \cellcolor{baselinerow}- & \cellcolor{baselinerow}- & \cellcolor{baselinerow}$0.234(\pm0.004)$ & \cellcolor{baselinerow}$0.308(\pm0.011)$ & \cellcolor{baselinerow}$0.001(\pm0.001)$ & \cellcolor{baselinerow}$0.001(\pm0.001)$ & \cellcolor{baselinerow}$10.307(\pm0.174)$ & \cellcolor{baselinerow}$13.864(\pm0.565)$ & \cellcolor{baselinerow}$0.590(\pm0.011)$ & \cellcolor{baselinerow}$0.817(\pm0.034)$ & \cellcolor{baselinerow}$0.305(\pm0.015)$ & \cellcolor{baselinerow}$0.306(\pm0.006)$ & \cellcolor{baselinerow}$0.389(\pm0.008)$ & \cellcolor{baselinerow}$0.386(\pm0.016)$ \\
    \cmidrule(lr){2-15}
    & DANN & DEV & $0.193(\pm0.005)$ & $0.320(\pm0.050)$ & $0.001(\pm0.000)$ & $0.001(\pm0.000)$ & $10.651(\pm0.318)$ & $18.657(\pm3.344)$ & $0.611(\pm0.020)$ & $1.098(\pm0.199)$ & $0.326(\pm0.026)$ & $0.293(\pm0.020)$ & $0.399(\pm0.010)$ & $0.391(\pm0.021)$ \\
    & DANN & IWV & $0.192(\pm0.004)$ & $0.357(\pm0.015)$ & $0.002(\pm0.001)$ & $0.002(\pm0.001)$ & $10.546(\pm0.249)$ & $21.051(\pm0.921)$ & $0.605(\pm0.016)$ & $1.240(\pm0.053)$ & $0.318(\pm0.023)$ & $0.309(\pm0.030)$ & $0.397(\pm0.018)$ & $0.390(\pm0.025)$ \\
    & DANN & SB & $0.189(\pm0.006)$ & $0.269(\pm0.052)$ & $0.001(\pm0.000)$ & $0.001(\pm0.000)$ & $10.395(\pm0.317)$ & $15.291(\pm3.489)$ & $0.595(\pm0.019)$ & $0.901(\pm0.207)$ & $0.305(\pm0.030)$ & $0.282(\pm0.020)$ & $0.391(\pm0.020)$ & $0.381(\pm0.028)$ \\
    & DANN & TB & $0.185(\pm0.003)$ & $0.237(\pm0.004)$ & $0.001(\pm0.001)$ & $0.002(\pm0.001)$ & $10.192(\pm0.162)$ & $13.212(\pm0.223)$ & $0.583(\pm0.010)$ & $0.776(\pm0.016)$ & $0.294(\pm0.020)$ & $0.267(\pm0.011)$ & $0.384(\pm0.025)$ & $0.386(\pm0.022)$ \\
    \cmidrule(lr){2-15}
    & CMD & DEV & $0.189(\pm0.006)$ & $0.331(\pm0.043)$ & $0.001(\pm0.000)$ & $0.001(\pm0.000)$ & $10.427(\pm0.309)$ & $19.380(\pm2.662)$ & $0.597(\pm0.019)$ & $1.150(\pm0.145)$ & $0.296(\pm0.019)$ & $0.295(\pm0.038)$ & $0.377(\pm0.017)$ & $0.343(\pm0.028)$ \\
    & CMD & IWV & $0.189(\pm0.004)$ & $0.311(\pm0.071)$ & $0.001(\pm0.000)$ & $0.001(\pm0.000)$ & $10.399(\pm0.193)$ & $17.933(\pm4.575)$ & $0.595(\pm0.011)$ & $1.057(\pm0.267)$ & $0.284(\pm0.019)$ & $0.317(\pm0.071)$ & $0.383(\pm0.019)$ & $0.356(\pm0.062)$ \\
    & CMD & SB & $0.185(\pm0.003)$ & $0.289(\pm0.073)$ & $0.002(\pm0.001)$ & $0.002(\pm0.001)$ & $10.176(\pm0.122)$ & $16.522(\pm4.645)$ & $0.582(\pm0.008)$ & $0.972(\pm0.270)$ & $0.284(\pm0.012)$ & $0.313(\pm0.074)$ & $0.383(\pm0.014)$ & $0.365(\pm0.052)$ \\
    & CMD & TB & $0.189(\pm0.007)$ & $0.241(\pm0.004)$ & $0.002(\pm0.000)$ & $0.002(\pm0.000)$ & $10.436(\pm0.390)$ & $13.488(\pm0.250)$ & $0.598(\pm0.024)$ & $0.793(\pm0.016)$ & $0.301(\pm0.035)$ & $0.289(\pm0.031)$ & $0.400(\pm0.030)$ & $0.402(\pm0.027)$ \\
    \cmidrule(lr){2-15}
    & DARE-GRAM & DEV & $0.189(\pm0.006)$ & $0.250(\pm0.004)$ & $0.001(\pm0.001)$ & $0.001(\pm0.001)$ & $10.419(\pm0.367)$ & $14.014(\pm0.229)$ & $0.596(\pm0.022)$ & $0.825(\pm0.013)$ & $0.305(\pm0.030)$ & $0.286(\pm0.028)$ & $0.393(\pm0.016)$ & $0.408(\pm0.015)$ \\
    & DARE-GRAM & IWV & $0.188(\pm0.007)$ & $0.250(\pm0.005)$ & $0.002(\pm0.001)$ & $0.002(\pm0.001)$ & $10.347(\pm0.408)$ & $14.008(\pm0.272)$ & $0.592(\pm0.025)$ & $0.826(\pm0.017)$ & $0.302(\pm0.022)$ & $0.274(\pm0.006)$ & $0.379(\pm0.023)$ & $0.391(\pm0.033)$ \\
    & DARE-GRAM & SB & $0.187(\pm0.004)$ & $0.247(\pm0.006)$ & $0.001(\pm0.001)$ & $0.001(\pm0.001)$ & $10.295(\pm0.250)$ & $13.878(\pm0.411)$ & $0.589(\pm0.015)$ & $0.818(\pm0.027)$ & $0.303(\pm0.018)$ & $0.278(\pm0.018)$ & $0.395(\pm0.033)$ & $0.409(\pm0.038)$ \\
    & DARE-GRAM & TB & $0.182(\pm0.005)$ & $0.237(\pm0.001)$ & $0.001(\pm0.000)$ & $0.001(\pm0.000)$ & $10.018(\pm0.280)$ & $13.206(\pm0.089)$ & $0.573(\pm0.017)$ & $0.776(\pm0.005)$ & $0.287(\pm0.018)$ & $0.272(\pm0.019)$ & $0.377(\pm0.012)$ & $0.374(\pm0.009)$ \\
    \cmidrule(lr){2-15}
    & Deep Coral & DEV & $0.190(\pm0.007)$ & $0.255(\pm0.007)$ & $0.002(\pm0.001)$ & $0.002(\pm0.001)$ & $10.490(\pm0.380)$ & $14.383(\pm0.462)$ & $0.601(\pm0.023)$ & $0.851(\pm0.025)$ & $0.309(\pm0.022)$ & $0.281(\pm0.020)$ & $0.410(\pm0.027)$ & $0.420(\pm0.021)$ \\
    & Deep Coral & IWV & $0.190(\pm0.007)$ & $0.249(\pm0.014)$ & $0.001(\pm0.001)$ & $0.001(\pm0.001)$ & $10.467(\pm0.387)$ & $14.020(\pm0.909)$ & $0.599(\pm0.024)$ & $0.826(\pm0.058)$ & $0.304(\pm0.019)$ & $0.279(\pm0.010)$ & $0.395(\pm0.016)$ & $0.403(\pm0.019)$ \\
    & \cellcolor{bestrow}\underline{Deep Coral} & \cellcolor{bestrow}\underline{SB} & \cellcolor{bestrow}$0.186(\pm0.003)$ & \cellcolor{bestrow}$\underline{0.237(\pm0.008)}$ & \cellcolor{bestrow}$0.001(\pm0.000)$ & \cellcolor{bestrow}$0.001(\pm0.000)$ & \cellcolor{bestrow}$10.246(\pm0.166)$ & \cellcolor{bestrow}$13.270(\pm0.473)$ & \cellcolor{bestrow}$0.586(\pm0.011)$ & \cellcolor{bestrow}$0.778(\pm0.029)$ & \cellcolor{bestrow}$0.292(\pm0.015)$ & \cellcolor{bestrow}$0.279(\pm0.002)$ & \cellcolor{bestrow}$0.389(\pm0.012)$ & \cellcolor{bestrow}$0.386(\pm0.007)$ \\
    & Deep Coral & TB & $0.184(\pm0.003)$ & $0.235(\pm0.006)$ & $0.001(\pm0.000)$ & $0.001(\pm0.001)$ & $10.181(\pm0.180)$ & $13.162(\pm0.361)$ & $0.582(\pm0.012)$ & $0.772(\pm0.024)$ & $0.283(\pm0.017)$ & $0.275(\pm0.007)$ & $0.376(\pm0.013)$ & $0.375(\pm0.011)$ \\
    \midrule
    \multirow{17}{*}{PointNet} & \cellcolor{baselinerow}- & \cellcolor{baselinerow}- & \cellcolor{baselinerow}$0.188(\pm0.012)$ & \cellcolor{baselinerow}$0.267(\pm0.005)$ & \cellcolor{baselinerow}$0.003(\pm0.001)$ & \cellcolor{baselinerow}$0.003(\pm0.001)$ & \cellcolor{baselinerow}$8.245(\pm0.453)$ & \cellcolor{baselinerow}$12.108(\pm0.284)$ & \cellcolor{baselinerow}$0.481(\pm0.025)$ & \cellcolor{baselinerow}$0.732(\pm0.019)$ & \cellcolor{baselinerow}$0.159(\pm0.018)$ & \cellcolor{baselinerow}$0.175(\pm0.018)$ & \cellcolor{baselinerow}$0.376(\pm0.091)$ & \cellcolor{baselinerow}$0.389(\pm0.091)$ \\
    \cmidrule(lr){2-15}
    & DANN & DEV & $0.151(\pm0.007)$ & $0.368(\pm0.018)$ & $0.002(\pm0.001)$ & $0.002(\pm0.001)$ & $8.314(\pm0.357)$ & $21.768(\pm1.027)$ & $0.486(\pm0.017)$ & $1.291(\pm0.052)$ & $0.154(\pm0.018)$ & $0.271(\pm0.051)$ & $0.299(\pm0.018)$ & $0.328(\pm0.019)$ \\
    & DANN & IWV & $0.145(\pm0.006)$ & $0.368(\pm0.012)$ & $0.002(\pm0.001)$ & $0.002(\pm0.001)$ & $8.063(\pm0.357)$ & $21.842(\pm0.745)$ & $0.472(\pm0.021)$ & $1.296(\pm0.041)$ & $0.150(\pm0.018)$ & $0.262(\pm0.007)$ & $0.293(\pm0.014)$ & $0.331(\pm0.015)$ \\
    & DANN & SB & $0.145(\pm0.006)$ & $0.333(\pm0.079)$ & $0.002(\pm0.001)$ & $0.002(\pm0.001)$ & $8.046(\pm0.347)$ & $19.720(\pm4.936)$ & $0.471(\pm0.020)$ & $1.173(\pm0.283)$ & $0.151(\pm0.018)$ & $0.232(\pm0.066)$ & $0.299(\pm0.017)$ & $0.334(\pm0.011)$ \\
    & DANN & TB & $0.152(\pm0.003)$ & $0.212(\pm0.001)$ & $0.002(\pm0.001)$ & $0.002(\pm0.001)$ & $8.368(\pm0.129)$ & $12.036(\pm0.057)$ & $0.489(\pm0.007)$ & $0.727(\pm0.008)$ & $0.154(\pm0.007)$ & $0.145(\pm0.016)$ & $0.325(\pm0.005)$ & $0.334(\pm0.011)$ \\
    \cmidrule(lr){2-15}
    & CMD & DEV & $0.188(\pm0.046)$ & $0.314(\pm0.012)$ & $0.002(\pm0.001)$ & $0.001(\pm0.001)$ & $10.617(\pm2.806)$ & $18.637(\pm0.876)$ & $0.627(\pm0.170)$ & $1.114(\pm0.045)$ & $0.188(\pm0.058)$ & $0.197(\pm0.031)$ & $0.348(\pm0.027)$ & $0.345(\pm0.023)$ \\
    & CMD & IWV & $0.149(\pm0.011)$ & $0.277(\pm0.069)$ & $0.002(\pm0.001)$ & $0.002(\pm0.001)$ & $8.211(\pm0.593)$ & $16.338(\pm4.393)$ & $0.477(\pm0.038)$ & $0.976(\pm0.280)$ & $0.162(\pm0.029)$ & $0.184(\pm0.026)$ & $0.358(\pm0.075)$ & $0.353(\pm0.055)$ \\
    & CMD & SB & $0.144(\pm0.007)$ & $0.224(\pm0.047)$ & $0.002(\pm0.001)$ & $0.002(\pm0.001)$ & $7.951(\pm0.373)$ & $12.901(\pm3.066)$ & $0.464(\pm0.025)$ & $0.778(\pm0.198)$ & $0.159(\pm0.029)$ & $0.165(\pm0.041)$ & $0.346(\pm0.061)$ & $0.370(\pm0.055)$ \\
    & CMD & TB & $0.145(\pm0.010)$ & $0.202(\pm0.018)$ & $0.002(\pm0.001)$ & $0.002(\pm0.001)$ & $8.023(\pm0.513)$ & $11.468(\pm1.154)$ & $0.467(\pm0.031)$ & $0.689(\pm0.089)$ & $0.163(\pm0.028)$ & $0.161(\pm0.041)$ & $0.327(\pm0.036)$ & $0.339(\pm0.017)$ \\
    \cmidrule(lr){2-15}
    & DARE-GRAM & DEV & $0.154(\pm0.010)$ & $0.228(\pm0.008)$ & $0.002(\pm0.001)$ & $0.002(\pm0.001)$ & $8.486(\pm0.533)$ & $12.973(\pm0.526)$ & $0.495(\pm0.029)$ & $0.791(\pm0.041)$ & $0.152(\pm0.017)$ & $0.147(\pm0.012)$ & $0.345(\pm0.050)$ & $0.359(\pm0.047)$ \\
    & DARE-GRAM & IWV & $0.151(\pm0.012)$ & $0.218(\pm0.010)$ & $0.002(\pm0.001)$ & $0.002(\pm0.001)$ & $8.336(\pm0.614)$ & $12.391(\pm0.698)$ & $0.487(\pm0.035)$ & $0.755(\pm0.044)$ & $0.155(\pm0.016)$ & $0.141(\pm0.014)$ & $0.359(\pm0.039)$ & $0.370(\pm0.036)$ \\
    & DARE-GRAM & SB & $0.146(\pm0.004)$ & $0.216(\pm0.013)$ & $0.002(\pm0.001)$ & $0.003(\pm0.001)$ & $8.071(\pm0.224)$ & $12.366(\pm0.775)$ & $0.473(\pm0.012)$ & $0.756(\pm0.054)$ & $0.146(\pm0.012)$ & $0.135(\pm0.013)$ & $0.330(\pm0.030)$ & $0.340(\pm0.024)$ \\
    & DARE-GRAM & TB & $0.144(\pm0.005)$ & $0.206(\pm0.008)$ & $0.002(\pm0.001)$ & $0.002(\pm0.001)$ & $7.977(\pm0.268)$ & $11.710(\pm0.434)$ & $0.467(\pm0.013)$ & $0.714(\pm0.025)$ & $0.144(\pm0.006)$ & $0.138(\pm0.015)$ & $0.332(\pm0.030)$ & $0.340(\pm0.025)$ \\
    \cmidrule(lr){2-15}
    & Deep Coral & DEV & $0.153(\pm0.010)$ & $0.215(\pm0.006)$ & $0.001(\pm0.001)$ & $0.001(\pm0.001)$ & $8.415(\pm0.484)$ & $12.195(\pm0.315)$ & $0.492(\pm0.025)$ & $0.740(\pm0.012)$ & $0.158(\pm0.018)$ & $0.143(\pm0.019)$ & $0.366(\pm0.085)$ & $0.374(\pm0.084)$ \\
    & Deep Coral & IWV & $0.157(\pm0.008)$ & $0.220(\pm0.010)$ & $0.002(\pm0.001)$ & $0.002(\pm0.001)$ & $8.616(\pm0.402)$ & $12.409(\pm0.618)$ & $0.502(\pm0.021)$ & $0.749(\pm0.035)$ & $0.158(\pm0.015)$ & $0.146(\pm0.019)$ & $0.407(\pm0.127)$ & $0.416(\pm0.127)$ \\
    & \cellcolor{bestrow}\underline{Deep Coral} & \cellcolor{bestrow}\underline{SB} & \cellcolor{bestrow}$0.146(\pm0.007)$ & \cellcolor{bestrow}$\underline{0.208(\pm0.009)}$ & \cellcolor{bestrow}$0.002(\pm0.001)$ & \cellcolor{bestrow}$0.002(\pm0.001)$ & \cellcolor{bestrow}$8.056(\pm0.390)$ & \cellcolor{bestrow}$11.861(\pm0.519)$ & \cellcolor{bestrow}$0.472(\pm0.022)$ & \cellcolor{bestrow}$0.723(\pm0.031)$ & \cellcolor{bestrow}$0.148(\pm0.012)$ & \cellcolor{bestrow}$0.136(\pm0.010)$ & \cellcolor{bestrow}$0.313(\pm0.014)$ & \cellcolor{bestrow}$0.325(\pm0.015)$ \\
    & Deep Coral & TB & $0.147(\pm0.009)$ & $0.204(\pm0.004)$ & $0.001(\pm0.001)$ & $0.001(\pm0.001)$ & $8.143(\pm0.508)$ & $11.628(\pm0.163)$ & $0.477(\pm0.028)$ & $0.707(\pm0.008)$ & $0.151(\pm0.012)$ & $0.139(\pm0.007)$ & $0.362(\pm0.109)$ & $0.370(\pm0.097)$ \\
    \midrule
    \multirow{17}{*}{\underline{\textbf{Transolver}}} & \cellcolor{baselinerow}- & \cellcolor{baselinerow}- & \cellcolor{baselinerow}$0.059(\pm0.002)$ & \cellcolor{baselinerow}$0.057(\pm0.003)$ & \cellcolor{baselinerow}$0.001(\pm0.000)$ & \cellcolor{baselinerow}$0.001(\pm0.000)$ & \cellcolor{baselinerow}$2.426(\pm0.094)$ & \cellcolor{baselinerow}$2.501(\pm0.141)$ & \cellcolor{baselinerow}$0.133(\pm0.006)$ & \cellcolor{baselinerow}$0.143(\pm0.008)$ & \cellcolor{baselinerow}$0.045(\pm0.003)$ & \cellcolor{baselinerow}$0.044(\pm0.004)$ & \cellcolor{baselinerow}$0.274(\pm0.006)$ & \cellcolor{baselinerow}$0.294(\pm0.003)$ \\
    \cmidrule(lr){2-15}
    & DANN & DEV & $0.046(\pm0.002)$ & $0.079(\pm0.017)$ & $0.001(\pm0.000)$ & $0.001(\pm0.000)$ & $2.387(\pm0.118)$ & $4.437(\pm0.994)$ & $0.132(\pm0.008)$ & $0.260(\pm0.060)$ & $0.044(\pm0.006)$ & $0.054(\pm0.008)$ & $0.278(\pm0.007)$ & $0.310(\pm0.006)$ \\
    & DANN & IWV & $0.047(\pm0.003)$ & $0.061(\pm0.010)$ & $0.001(\pm0.000)$ & $0.001(\pm0.000)$ & $2.371(\pm0.156)$ & $3.384(\pm0.582)$ & $0.130(\pm0.009)$ & $0.198(\pm0.037)$ & $0.042(\pm0.004)$ & $0.045(\pm0.003)$ & $0.284(\pm0.002)$ & $0.306(\pm0.011)$ \\
    & DANN & SB & $0.046(\pm0.001)$ & $0.068(\pm0.010)$ & $0.001(\pm0.000)$ & $0.001(\pm0.000)$ & $2.349(\pm0.113)$ & $3.771(\pm0.539)$ & $0.129(\pm0.008)$ & $0.221(\pm0.033)$ & $0.044(\pm0.006)$ & $0.050(\pm0.006)$ & $0.282(\pm0.008)$ & $0.312(\pm0.005)$ \\
    & DANN & TB & $0.049(\pm0.003)$ & $0.050(\pm0.002)$ & $0.001(\pm0.000)$ & $0.001(\pm0.000)$ & $2.523(\pm0.121)$ & $2.751(\pm0.111)$ & $0.140(\pm0.006)$ & $0.158(\pm0.006)$ & $0.050(\pm0.004)$ & $0.044(\pm0.003)$ & $0.292(\pm0.011)$ & $0.303(\pm0.017)$ \\
    \cmidrule(lr){2-15}
    & CMD & DEV & $0.046(\pm0.003)$ & $0.056(\pm0.002)$ & $0.001(\pm0.000)$ & $0.001(\pm0.000)$ & $2.381(\pm0.148)$ & $3.112(\pm0.100)$ & $0.131(\pm0.006)$ & $0.181(\pm0.007)$ & $0.046(\pm0.004)$ & $0.046(\pm0.004)$ & $0.280(\pm0.006)$ & $0.299(\pm0.008)$ \\
    & CMD & IWV & $0.045(\pm0.002)$ & $0.050(\pm0.006)$ & $0.001(\pm0.000)$ & $0.001(\pm0.000)$ & $2.342(\pm0.089)$ & $2.757(\pm0.337)$ & $0.130(\pm0.005)$ & $0.159(\pm0.020)$ & $0.048(\pm0.005)$ & $0.045(\pm0.005)$ & $0.282(\pm0.005)$ & $0.300(\pm0.008)$ \\
    & CMD & SB & $0.046(\pm0.002)$ & $0.053(\pm0.004)$ & $0.001(\pm0.000)$ & $0.001(\pm0.000)$ & $2.370(\pm0.086)$ & $2.942(\pm0.246)$ & $0.131(\pm0.005)$ & $0.171(\pm0.014)$ & $0.048(\pm0.005)$ & $0.047(\pm0.004)$ & $0.279(\pm0.002)$ & $0.298(\pm0.008)$ \\
    & CMD & TB & $0.046(\pm0.002)$ & $0.046(\pm0.002)$ & $0.001(\pm0.000)$ & $0.001(\pm0.000)$ & $2.394(\pm0.108)$ & $2.510(\pm0.137)$ & $0.133(\pm0.005)$ & $0.144(\pm0.008)$ & $0.044(\pm0.001)$ & $0.041(\pm0.002)$ & $0.279(\pm0.009)$ & $0.296(\pm0.006)$ \\
    \cmidrule(lr){2-15}
    & DARE-GRAM & DEV & $0.046(\pm0.002)$ & $0.046(\pm0.002)$ & $0.001(\pm0.000)$ & $0.001(\pm0.000)$ & $2.400(\pm0.080)$ & $2.513(\pm0.093)$ & $0.132(\pm0.004)$ & $0.144(\pm0.005)$ & $0.045(\pm0.003)$ & $0.041(\pm0.002)$ & $0.276(\pm0.004)$ & $0.293(\pm0.004)$ \\
    & DARE-GRAM & IWV & $0.045(\pm0.003)$ & $0.048(\pm0.003)$ & $0.001(\pm0.000)$ & $0.001(\pm0.000)$ & $2.315(\pm0.142)$ & $2.591(\pm0.193)$ & $0.128(\pm0.007)$ & $0.148(\pm0.012)$ & $0.042(\pm0.001)$ & $0.040(\pm0.003)$ & $0.276(\pm0.005)$ & $0.296(\pm0.003)$ \\
    & DARE-GRAM & SB & $0.045(\pm0.003)$ & $0.047(\pm0.003)$ & $0.001(\pm0.000)$ & $0.001(\pm0.000)$ & $2.326(\pm0.133)$ & $2.577(\pm0.191)$ & $0.129(\pm0.006)$ & $0.148(\pm0.012)$ & $0.044(\pm0.003)$ & $0.042(\pm0.003)$ & $0.277(\pm0.005)$ & $0.293(\pm0.006)$ \\
    & DARE-GRAM & TB & $0.046(\pm0.001)$ & $0.045(\pm0.002)$ & $0.001(\pm0.000)$ & $0.001(\pm0.000)$ & $2.387(\pm0.072)$ & $2.443(\pm0.121)$ & $0.131(\pm0.004)$ & $0.140(\pm0.007)$ & $0.046(\pm0.003)$ & $0.042(\pm0.003)$ & $0.280(\pm0.002)$ & $0.291(\pm0.004)$ \\
    \cmidrule(lr){2-15}
    & Deep Coral & DEV & $0.044(\pm0.001)$ & $0.044(\pm0.001)$ & $0.001(\pm0.000)$ & $0.001(\pm0.000)$ & $2.273(\pm0.024)$ & $2.383(\pm0.056)$ & $0.126(\pm0.001)$ & $0.137(\pm0.003)$ & $0.042(\pm0.002)$ & $0.039(\pm0.001)$ & $0.277(\pm0.006)$ & $0.291(\pm0.004)$ \\
    & Deep Coral & IWV & $0.044(\pm0.001)$ & $0.044(\pm0.001)$ & $0.001(\pm0.000)$ & $0.001(\pm0.000)$ & $2.266(\pm0.037)$ & $2.394(\pm0.041)$ & $0.126(\pm0.002)$ & $0.137(\pm0.002)$ & $0.042(\pm0.003)$ & $0.040(\pm0.002)$ & $0.276(\pm0.006)$ & $0.291(\pm0.004)$ \\
    & \cellcolor{bestrow}\underline{\textbf{Deep Coral}} & \cellcolor{bestrow}\underline{\textbf{SB}} & \cellcolor{bestrow}$0.044(\pm0.002)$ & \cellcolor{bestrow}$\underline{\mathbf{0.043(\pm0.000)}}$ & \cellcolor{bestrow}$0.001(\pm0.000)$ & \cellcolor{bestrow}$0.001(\pm0.000)$ & \cellcolor{bestrow}$2.288(\pm0.072)$ & \cellcolor{bestrow}$2.370(\pm0.024)$ & \cellcolor{bestrow}$0.127(\pm0.003)$ & \cellcolor{bestrow}$0.136(\pm0.002)$ & \cellcolor{bestrow}$0.043(\pm0.003)$ & \cellcolor{bestrow}$0.040(\pm0.002)$ & \cellcolor{bestrow}$0.274(\pm0.002)$ & \cellcolor{bestrow}$0.291(\pm0.005)$ \\
    & Deep Coral & TB & $0.042(\pm0.001)$ & $0.043(\pm0.000)$ & $0.001(\pm0.000)$ & $0.001(\pm0.000)$ & $2.211(\pm0.038)$ & $2.320(\pm0.019)$ & $0.123(\pm0.002)$ & $0.133(\pm0.001)$ & $0.041(\pm0.003)$ & $0.039(\pm0.001)$ & $0.277(\pm0.008)$ & $0.295(\pm0.007)$ \\
    \bottomrule
  \end{tabular}
  }
\end{table}

\begin{figure*}[!b]
    \centering
    \begin{minipage}[t]{0.48\linewidth} 
        \vspace{0pt}
        \centering
        \includegraphics[width=\linewidth]{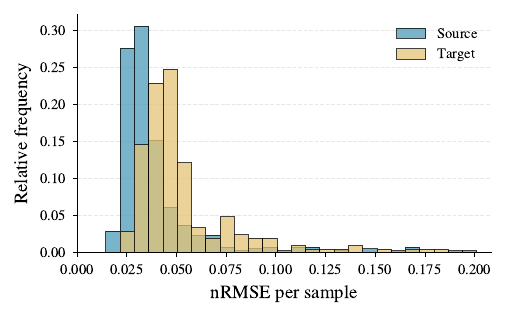}
        \caption{Distribution of \ac{nRMSE} (averaged across all fields) for the test sets of the source (blue) and target (yellow) domains.}
        \label{fig:error_dist_motor}
    \end{minipage}
    \hfill
    \begin{minipage}[t]{0.48\linewidth}
        \vspace{0pt}
        \centering
        \captionof{table}{Absolute error of Cauchy stress (xx-component) predictions for representative samples from the source and target domain of the \emph{electric motor design} dataset.
        Lowest value per metric is bold.}
        \label{tab:motor_best_worst}
        \def\arraystretch{1.1}
        \setlength{\tabcolsep}{8pt}
          \begin{tabular}{lcc}
            \toprule
            \textbf{Metric} & \textbf{Source} & \textbf{Target} \\
            \midrule
            Mean            & 2.56e+06 & \textbf{2.37e+06} \\
            Std             & \textbf{4.40e+06} & 4.46e+06 \\
            Median          & \textbf{1.30e+06} & 1.48e+06 \\
            Q\textsubscript{01} & \textbf{1.81e+04} & 2.35e+04 \\
            Q\textsubscript{25} & \textbf{5.19e+05} & 6.90e+05 \\
            Q\textsubscript{75} & 2.75e+06 & \textbf{2.55e+06} \\
            Q\textsubscript{99} & 2.03e+07 & \textbf{1.67e+07} \\
            \bottomrule
          \end{tabular}
    \end{minipage}
\end{figure*}

To assess the effect of the domain shift on prediction accuracy in the \emph{electric motor design} dataset further, \cref{fig:error_dist_motor} shows the distribution of \ac{nRMSE} for the best performing model, selected by lowest average error in the target domain, in the source and target domain.

In this task, the Cauchy stress (xx-component) is particularly interesting for downstream analysis and optimization.
We therefore focus our closer inspection on this field.

\begin{figure}[htbp]
    \centering
    \includegraphics[width=\textwidth]{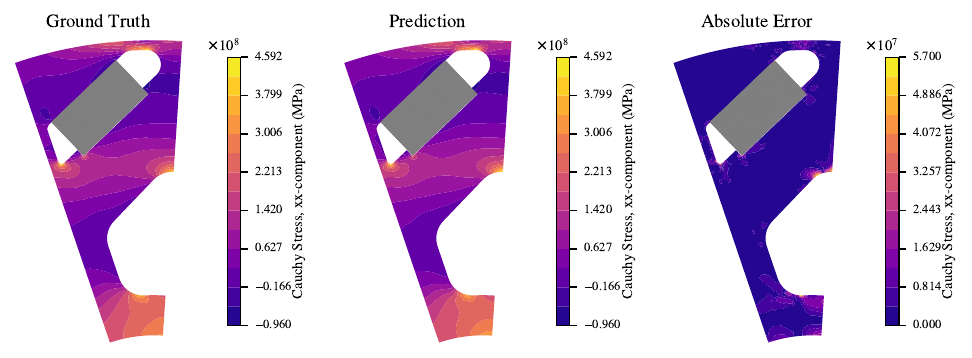}
    \caption{Fringe plot of the \emph{electric motor design} dataset (representative source sample). Shown is the ground truth (left) and predicted (middle) Cauchy stress (xx-component), as well as the absolute error (right).}
    \label{fig:motor_representative_source}
\end{figure}

\begin{figure}[htbp]
    \centering
    \includegraphics[width=\textwidth]{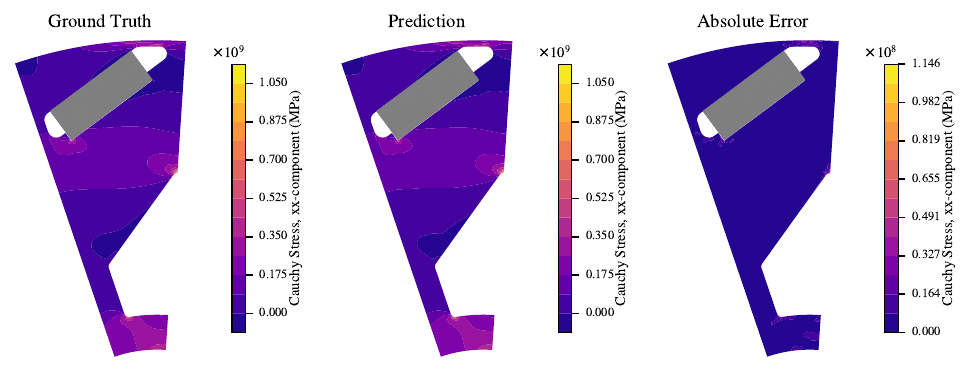}
    \caption{Fringe plot of the \emph{electric motor design} dataset (representative target sample). Shown is the ground truth (left) and predicted (middle) Cauchy stress (xx-component), as well as the absolute error (right).}
    \label{fig:motor_representative_target}
\end{figure}

\cref{tab:motor_best_worst} presents a comparison of absolute Cauchy stress (xx-component) errors for the representative samples both from the source and target test sets.
The corresponding fringe plots are shown in \cref{fig:motor_representative_source,fig:motor_representative_target}, comparing the ground truth and predicted fields alongside their absolute errors.

\FloatBarrier
\newpage
\subsection{Heatsink Design}

\cref{tab:heatsink_results} presents the complete benchmarking results for the \emph{heatsink design} dataset.

\begin{table}[h]
  \centering
    \caption{Performance metrics on the source and target domains (mean $\pm$ std over 4 seeds) for the \emph{heatsink design} task at \textit{medium} difficulty. RMSE is reported for all global metrics (All Fields Normalized Avg through Dynamic Pressure), while further columns report custom engineering and physics-based metrics. \textbf{Bold} indicates the overall best combination of architecture, \ac{UDA} algorithm, and model selection. Within each architecture group, the unregularized baseline is shaded \tightbox{baselinerow}{beige}, and the best \ac{UDA} configuration is \underline{underlined} and shaded \tightbox{bestrow}{green}.}
  \label{tab:heatsink_results}
  \resizebox{\textwidth}{!}{%
  \definecolor{bestrow}{HTML}{DFF0D8}
  \definecolor{baselinerow}{HTML}{FFF4CC}
\begin{tabular}{lllcccccccccccccc}
    \toprule
    \multirow{2}{*}{\textbf{Model}} & \multirow{2}{*}{\makecell{\textbf{DA}\\ \textbf{Algorithm}}} & \multirow{2}{*}{\makecell{\textbf{Model}\\ \textbf{Selection}}} & \multicolumn{2}{c}{\textbf{All Fields Normalized Avg (-)}} & \multicolumn{2}{c}{\textbf{Velocity (m/s)}} & \multicolumn{2}{c}{\textbf{Temperature (K)}} & \multicolumn{2}{c}{\textbf{Dynamic Pressure (Pa)}} & \multicolumn{2}{c}{\textbf{Rel Custom Error (-)}} & \multicolumn{2}{c}{\textbf{BC Violation Temperature (-)}} & \multicolumn{2}{c}{\textbf{BC Violation Velocity (m/s)}} \\
\cmidrule(lr){4-5} \cmidrule(lr){6-7} \cmidrule(lr){8-9} \cmidrule(lr){10-11} \cmidrule(lr){12-13} \cmidrule(lr){14-15} \cmidrule(lr){16-17}
      &   &  & \textbf{SRC} & \textbf{TGT} & \textbf{SRC} & \textbf{TGT} & \textbf{SRC} & \textbf{TGT} & \textbf{SRC} & \textbf{TGT} & \textbf{SRC} & \textbf{TGT} & \textbf{SRC} & \textbf{TGT} & \textbf{SRC} & \textbf{TGT} \\
    \midrule
    \multirow{17}{*}{ConditionedGINO} & \cellcolor{baselinerow}- & \cellcolor{baselinerow}- & \cellcolor{baselinerow}$0.318(\pm0.004)$ & \cellcolor{baselinerow}$0.564(\pm0.017)$ & \cellcolor{baselinerow}$0.030(\pm0.000)$ & \cellcolor{baselinerow}$0.049(\pm0.002)$ & \cellcolor{baselinerow}$7.392(\pm0.179)$ & \cellcolor{baselinerow}$14.728(\pm0.823)$ & \cellcolor{baselinerow}$270.213(\pm11.947)$ & \cellcolor{baselinerow}$1219.779(\pm179.521)$ & \cellcolor{baselinerow}$0.013(\pm0.001)$ & \cellcolor{baselinerow}$0.023(\pm0.002)$ & \cellcolor{baselinerow}$0.106(\pm0.006)$ & \cellcolor{baselinerow}$0.195(\pm0.016)$ & \cellcolor{baselinerow}$0.046(\pm0.003)$ & \cellcolor{baselinerow}$0.087(\pm0.004)$ \\
    \cmidrule(lr){2-17}
    & DANN & DEV & $0.238(\pm0.003)$ & $0.430(\pm0.039)$ & $0.030(\pm0.000)$ & $0.049(\pm0.003)$ & $7.423(\pm0.080)$ & $16.226(\pm4.127)$ & $266.337(\pm15.950)$ & $1104.462(\pm91.802)$ & $0.012(\pm0.000)$ & $0.029(\pm0.015)$ & $0.106(\pm0.002)$ & $0.282(\pm0.083)$ & $0.046(\pm0.001)$ & $0.091(\pm0.017)$ \\
\bottomrule
  \end{tabular}
  }
\end{table}

\begin{figure*}[htbp]
    \centering
    \begin{minipage}[t]{0.48\linewidth} 
        \vspace{0pt}
        \centering
        \includegraphics[width=\linewidth]{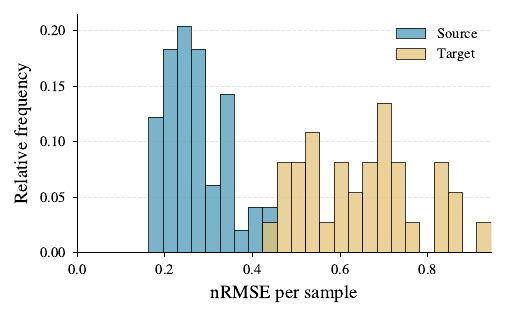}
        \caption{Distribution of \ac{nRMSE} (averaged across all fields) for the test sets of the source (blue) and target (yellow) domains.}
        \label{fig:error_dist_heatsink}
    \end{minipage}
    \hfill
    \begin{minipage}[t]{0.48\linewidth}
        \vspace{0pt}
        \centering
        \captionof{table}{Absolute error (in K) of temperature predictions for representative samples from the source and target domain of the \emph{heatsink design} dataset. Lowest value per metric is bold.}
        \label{tab:heatsink_best_worst}
        \def\arraystretch{1.1}
        \setlength{\tabcolsep}{8pt}
          \begin{tabular}{lcc}
            \toprule
            \textbf{Metric} & \textbf{Source} & \textbf{Target} \\
            \midrule
            Mean            & \textbf{2.87e+00} & 1.09e+01 \\
            Std             & \textbf{3.02e+00} & 1.17e+01 \\
            Median          & \textbf{1.88e+00} & 6.55e+00 \\
            Q\textsubscript{01} & \textbf{3.20e-02} & 7.18e-02 \\
            Q\textsubscript{25} & \textbf{8.40e-01} & 2.18e+00 \\
            Q\textsubscript{75} & \textbf{3.89e+00} & 1.60e+01 \\
            Q\textsubscript{99} & \textbf{1.44e+01} & 5.19e+01 \\
            \bottomrule
          \end{tabular}
    \end{minipage}
\end{figure*}

We further investigate model performance under distribution shift by examining predictions from the best performing Transolver model, selected by lowest average error in the target domain.
\cref{fig:error_dist_heatsink} presents the respective distribution of prediction errors in the source and target domain, clearly indicating the negative effects of the distribution shift on model performance.

In this task, the temperature field is the most critical for downstream analysis and optimization, which is why we focus our detailed analysis on it.

\begin{figure}[htbp]
    \centering
    \includegraphics[width=0.9\textwidth]{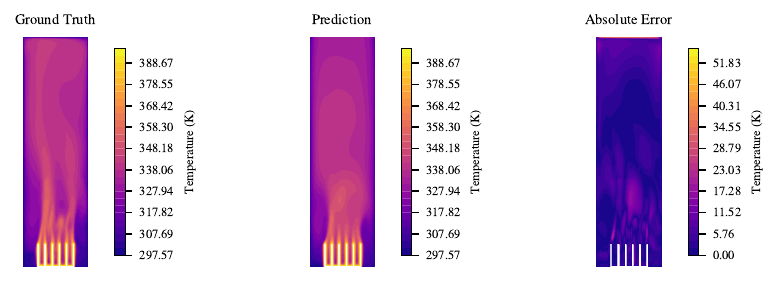}
    \caption{Sliced fringe plot of the \emph{heatsink design} dataset (representative source sample). Shown is the ground truth (left) and predicted (middle) temperature field, as well as the absolute error (right).}
    \label{fig:heatsink_source_representative}
\end{figure}

\begin{figure}[htbp]
    \centering
    \includegraphics[width=0.9\textwidth]{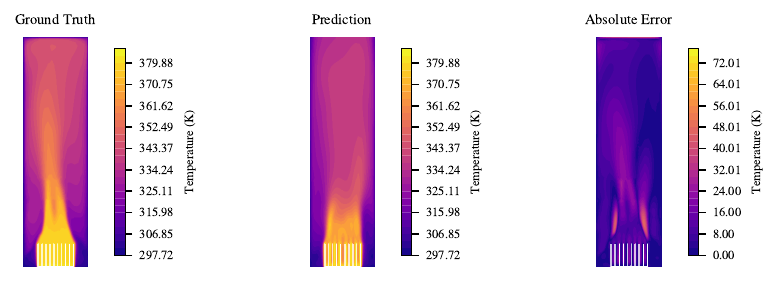}
    \caption{Sliced fringe plot of the \emph{heatsink design} dataset (representative target sample). Shown is the ground truth (left) and predicted (middle) temperature field, as well as the absolute error (right).}
    \label{fig:heatsink_target_representative}
\end{figure}

\cref{tab:heatsink_best_worst} compares the absolute temperature prediction errors for the best and worst samples from both the source and target test sets.
The corresponding fringe plots are shown in \cref{fig:heatsink_source_representative,fig:heatsink_target_representative}, comparing the ground truth and predicted temperature fields, alongside their absolute errors.


\newpage
\section{Distribution Shifts}
\label{app:distribution_shifts}
\subsection{Latent space visualizations}
\label{app:tsne_plots}

\begin{figure}[htbp]
    \centering
    \includegraphics[width=0.58\textwidth]{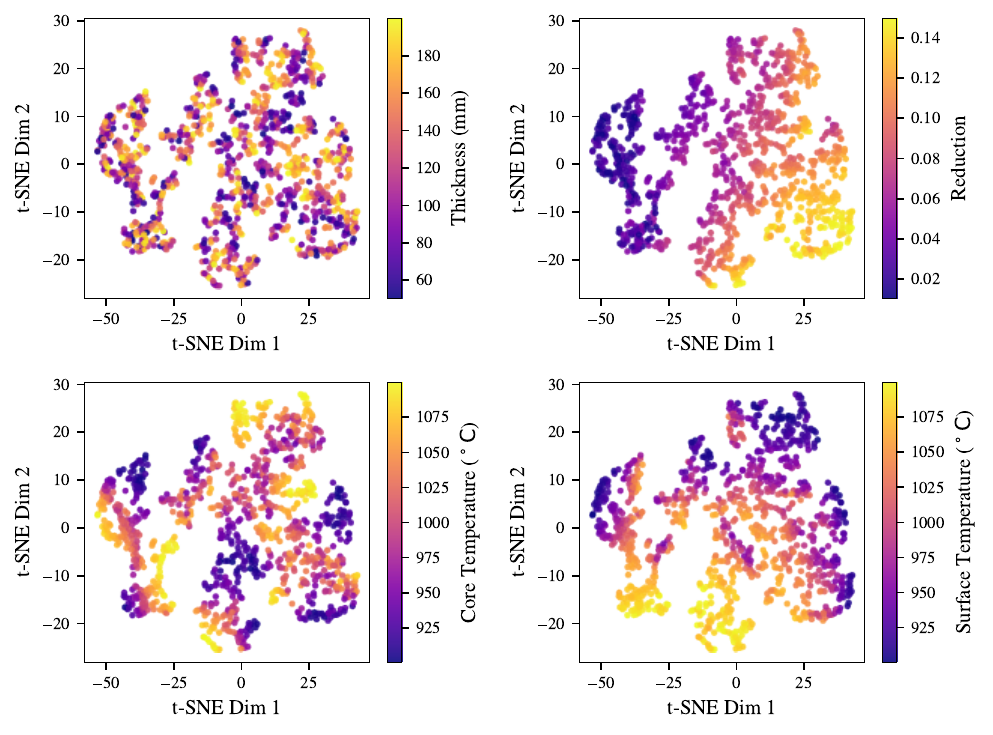}
    \caption{T-SNE visualization of the conditioning vectors for the \emph{hot rolling} dataset.
    Point color indicates the magnitude of the respective parameter.
    Slab thickness (top left) is roughly uniformly distributed, while the remaining three exhibit distinct clustering patterns.
    Taking into account domain knowledge from industry experts, we base our distribution shifts on the reduction parameter $r$ (top right).}
    \label{fig:tsne_rolling}
\end{figure}

\begin{figure}[htbp]
    \centering
    \includegraphics[width=0.58\textwidth]{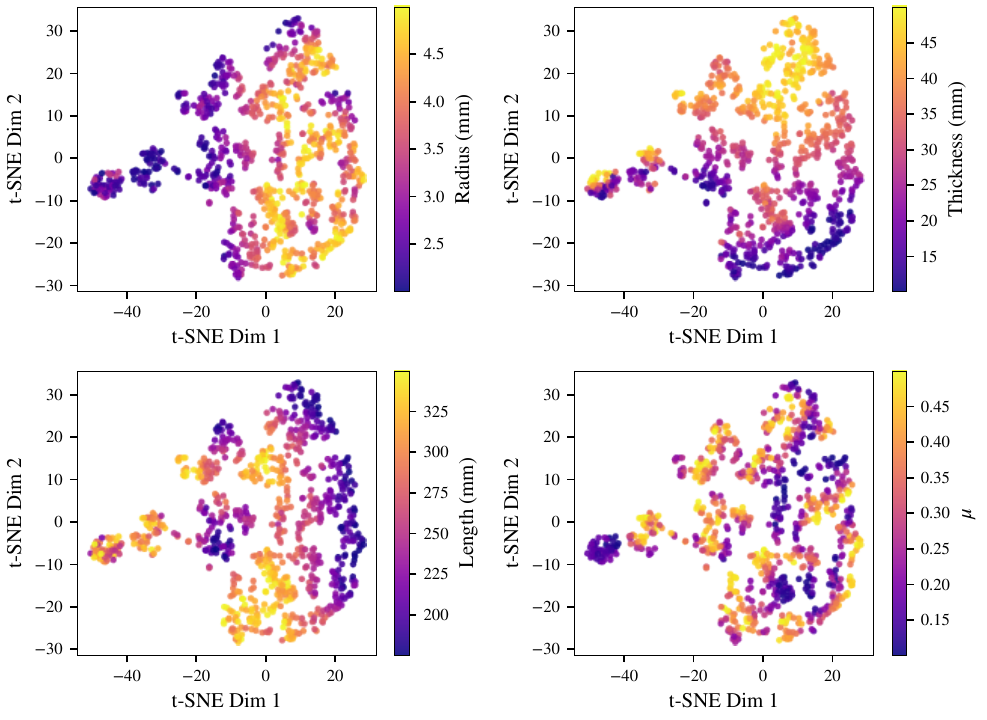}
    \caption{
    T-SNE visualization of the conditioning vectors for the \emph{sheet metal forming} dataset.
    Point color indicates the magnitude of the respective parameter.
    Showing the strongest clustering behavior, we choose sheet thickness $t$ (top right) as the domain defining parameter.
    }
    \label{fig:tsne_forming}
\end{figure}

\begin{figure}[htbp]
    \centering
    \includegraphics[width=0.65\textwidth]{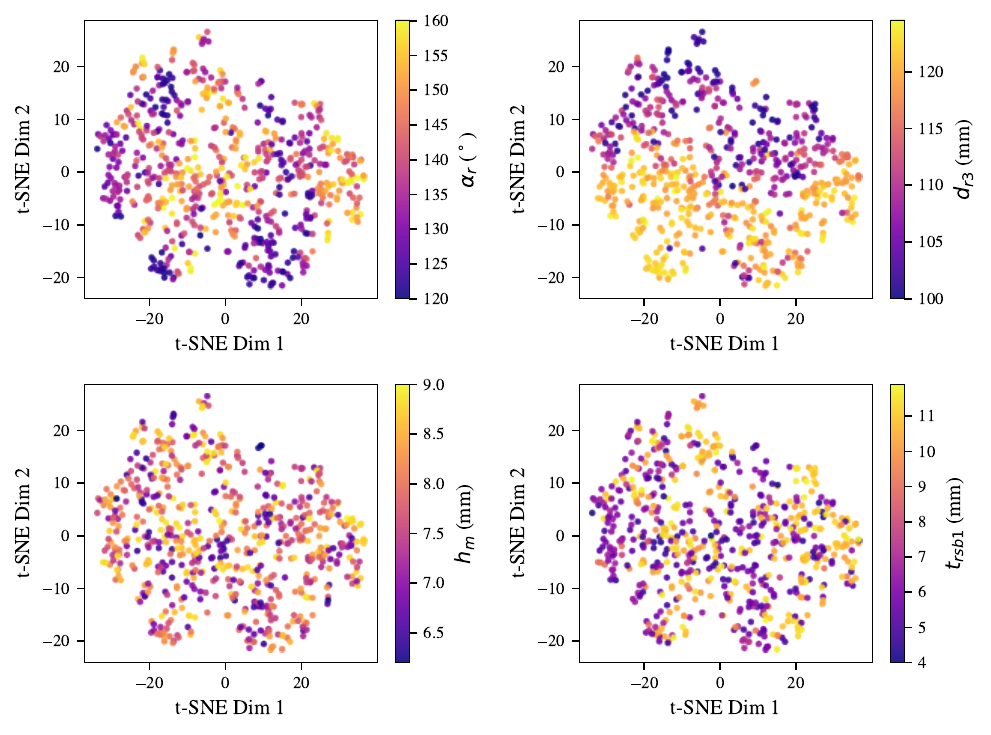}
    \caption{
    T-SNE visualization of the conditioning vectors for the \emph{electric motor design} dataset. Point color indicates the magnitude of the respective parameter.
    For clarity, we only show selected parameters.
    The parameter exhibiting the most structure in the latent space is $d_{r3}$.
    We therefore choose this to be our domain defining parameter in accordance with domain experts.
    }
    \label{fig:tsne_motor}
\end{figure}

\begin{figure}[htbp]
    \centering
    \includegraphics[width=0.65\textwidth]{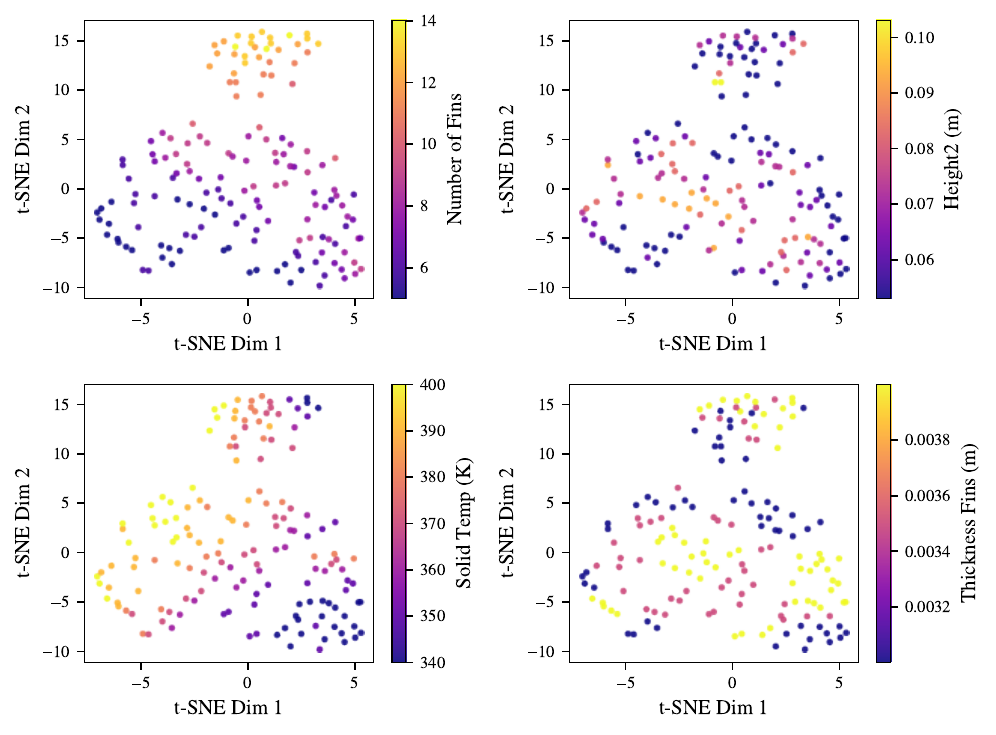}
    \caption{
    T-SNE visualization of the conditioning vectors for the \emph{heatsink design} dataset.
    Point color indicates the magnitude of the respective parameter.
    Height 2 is distributed equally across the representation, but the other parameters show concrete grouping behavior.
    In accordance with domain experts, we choose the number of fins as the domain defining parameter.
    }
    \label{fig:tsne_heatsink}
\end{figure}

To gain more insights into the parameter importance besides the domain experts' opinion, we visualize the latent space of the conditioning network for all presented datasets in \crefrange{fig:tsne_rolling}{fig:tsne_heatsink}.


\newpage
\newpage
\subsection{PAD Analysis}
\label{app:PAD}
To evaluate domain discrepancy on the output spaces, we investigated transfer error of models.
The \ac{PAD} can serve as a bound to the $\mathcal{H}$-divergence, which in turn is an upper bound the maximum transfer error itself (for details see \cite{bouvier2020robust_da}, \cite{johansson2019support} and \cite{zellinger2021balancing}).
We estimate \ac{PAD} by training a binary PointNet based domain classifier on mesh level simulation outputs and converting its test error into a divergence estimate.
The resulting \ac{PAD} values for each difficulty are reported in \cref{tab:pad_values}.
The \ac{PAD} values indicate a clear output space distribution shift across all datasets.
Additionally, the distances increase with increasing shift difficulty in all presented problems.


\section{Distance Measures}
\label{app:distance_measures}
\paragraph{Deep CORAL.}

This distance measures the difference in second-order statistics (covariances) of source and target latent features and can be calculated as follows:

\[
d_{\text{deep\_coral}}(\phi(\mathbf{x}),\phi(\mathbf{x'})) = \frac{1}{4k^2} \left\| \mathbf{C} - \mathbf{C'} \right\|_{F}^2,
\]

where $\phi(\mathbf{x}), \phi(\mathbf{x'}) \in \mathbb{R}^{n \times k}$ denote latent source and target features for a batch size $n$ and a feature dimension $k$, $\mathbf{C}$ and $\mathbf{C'}$ are the source and target feature covariances and $\|\cdot\|_{F}^2$ is the squared Frobenius norm.

\paragraph{CMD.}
CMD measures not only the difference in first and second moments of source and target latent features, but also in higher-order central moments. Let $\phi(\mathbf{x}), \phi(\mathbf{x}') \in \mathbb{R}^{n \times k}$ denote the latent activations for a batch size $n$ and feature dimension $k$. The CMD distance up to order $P$ is defined as

\[
d_{\text{cmd}}(\phi(\mathbf{x}), \phi(\mathbf{x}'))
= \frac{1}{|b-a|}
  \left\lVert \boldsymbol{\mu} - \boldsymbol{\mu}' \right\rVert_2
  + \sum_{p=2}^{P}
     \frac{1}{|b-a|^p}
     \left\lVert \mathbf{c}_p(\phi(\mathbf{x})) - \mathbf{c}_p(\phi(\mathbf{x}')) \right\rVert_2,
\]

where $\boldsymbol{\mu}, \boldsymbol{\mu'} \in \mathbb{R}^{k}$ are the source and target empirical mean feature vectors, $|b-a|^p$ can be seen as a hyperparameter of the method which we set to 2 to reflect the original implementation, and $\mathbf{c}_p(\phi(\mathbf{x})), \mathbf{c'}_p(\phi(\mathbf{x'})) \in \mathbb{R}^{k}$ are the respective $p$-th central moments which are calculated as:

\[
\mathbf{c}_p(\phi(\mathbf{x}))
= \frac{1}{n}\sum_{i=1}^{n}
  (\phi(\mathbf{x})_i - \boldsymbol{\mu})^{\odot p},
\qquad
\mathbf{c}_p(\phi(\mathbf{x}'))
= \frac{1}{n}\sum_{i=1}^{n}
  (\phi(\mathbf{x}')_i - \boldsymbol{\mu}')^{\odot p}.
\]

Above, $(\cdot)^{\odot p}$ denotes the element-wise $p$-th power.
Choosing the number of higher-order moments to align is another hyperparameter of the method.
For our benchmark, we choose $P=5$.

\paragraph{DANN.}
DANN is introduced to minimize an upper bound on the $\mathcal{H}$-divergence between source and target feature distributions.
Since it is intractable to compute this directly, the authors use a domain classifier in the form of a small MLP trained to distinguish whether a latent feature comes from the source or the target domain.
The error of this classifier is then used to compute the \ac{PAD}, which, up to a constant depending on the model's VC dimension, upper-bounds the $\mathcal{H}$-divergence \citep{ganin2015dann}.

Training is performed via a min–max optimization, i.e. the domain classifier is trained to maximize its classification accuracy, while the feature encoder $\phi$ is trained to \emph{minimize} this separability by using a gradient reversal layer.
This adversarial interaction encourages the latent representations of source and target samples to become indistinguishable, thereby promoting domain-invariant features.

\paragraph{DARE-GRAM.}
DARE-GRAM aims to align a selected low-rank subspace of the pseudo-inverse Gram matrices of source and target features.
Given feature matrices $\phi(\mathbf{x}),\phi(\mathbf{x}') \in \mathbb{R}^{n\times k}$ for a batch size $n$ and feature dimension $k$, we can compute their Gram matrices:

$$G = \phi(\mathbf{x})^\top \phi(\mathbf{x}), \qquad G' = \phi(\mathbf{x})'^\top \phi(\mathbf{x})'.$$

Each Gram matrix is then decomposed via eigendecomposition, and its truncated Moore–Penrose pseudo-inverse is formed by keeping the top $p^\ast$ eigenvalues that explain a fixed proportion of variance ($95\%$ for our implementation):

\[
G^{+} = U_{1:p^\ast}\,\Lambda_{1:p^\ast}^{-1}\,U_{1:p^\ast}^{\top},
\qquad
(G')^{+} = U'_{1:p^\ast}\,(\Lambda'_{1:p^\ast})^{-1}\,(U'_{1:p^\ast})^{\top}.
\]

We can then define the difference in angles as

\[
d_{\text{angle}}(G, G') =
\bigl\|\, \mathbf{1} - \cos(\theta_{1:p^\ast}) \,\bigr\|_{1},
\]

where $\cos(\theta_i)$ is the cosine similarity between the $i$-th column of $G^{+}$ and $(G')^{+}$.

Furthermore, we can define the difference in scale as

\[
d_{\text{scale}}(G, G')
= \left\| \lambda_{1:p^\ast} - \lambda'_{1:p^\ast} \right\|_2.
\]

The first term aligns the orientation of the dominant inverse-Gram subspaces, whereas the second term matches the principal eigenvalues of the Gram matrices to ensure that feature scale is consistent across source and target.

The total DARE-GRAM distance is defined as a weighted sum of the two:

\[
d_{\text{dare\_gram}} = \alpha_{\text{angle}}\, d_{\text{angle}} + \gamma_{\text{scale}}\, d_{\text{scale}},
\]

where the $\alpha$ and $\gamma$ are hyperparameters.
Following the original authors, we set $\alpha_{\text{angle}} = 0.02$ and $\gamma_{\text{scale}} = 0.001$.

\section{Model Architectures}
\label{app:model_architectures}
This section provides explanations of all model architectures used in our benchmark. All models are implemented in PyTorch and are adapted to our conditional regression task.
All models have in common, that they take node coordinates as inputs and embed them using a sinusoidal positional encoding.
Additionally, all models are conditioned on the input parameters of the respective simulation sample, which are encoded through a conditioning network described below.

\paragraph{Conditioning Network.}
The conditioning module used for all neural surrogate architectures embeds the simulation input parameters into a latent vector used for conditioning.
The network consists of a sinusoidal encoding followed by a simple MLP. The dimension of the latent encoding is 8 throughout all experiments.

\paragraph{PointNet.}
Our PointNet implementation is adapted from \citep{qi2017pointnet} for node-level regression. Input node coordinates are first encoded using sinusoidal embeddings and passed through an encoder MLP.
The resulting representations are aggregated globally using max pooling over nodes to obtain a global feature vector.
To propagate this global feature, it is concatenated back to each point's feature vector.
This fused representation is then fed into a final MLP, which produces the output fields.
The conditioning is performed by concatenating the conditioning vector to the global feature before propagating it to the nodes features. We use a PointNet base dimension of 16 for the small model and 32 for the larger model.

\paragraph{GraphSAGE.}
We adapt GraphSAGE \citep{graphsage} to the conditional mesh regression setting. Again, input node coordinates are embedded using a sinusoidal encoding and passed through an MLP encoder.
The main body of the model consists of multiple GraphSAGE message passing layers with mean aggregation. We support two conditioning modes, namely concatenating the latent conditioning vector to the node features, or applying FiLM style modulation \citep{perez2018film} to the node features before each message passing layer. We always use FiLM modulation in the presented results.
After message passing, the node representations are passed through a final MLP decoder to produce the output fields. The base dimension of the model is kept at 128 and we employ 4 GraphSAGE layers.

\paragraph{Transolver.}
The Transolver model follows the originally introduced architecture \citep{wu2024Transolver}.
Similar to the other models, node coordinates first are embedded using a sinusoidal encoding and passed through an MLP encoder to produce initial features.
Through learned assignement, each node then gets mapped to a slice, and inter- as well as intra-slice attention is performed.
Afterwards, fields are decoded using an MLP readout.
The architecture supports two conditioning modes: concatenation, where the conditioning vector is concatenated to the input node features before projection, or modulation through DiT layers across the network. For our experiments, DiT is used. We choose a latent dimension of 128, a slice base of 32 and we apply four attention blocks for the small model. For the larger model, we scale to 256, 128 and 8 layers respectively.

\paragraph{UPT.}
Our UPT implementation builds on the architecture proposed in~\citep{alkin2024upt}.
First, a fixed number of supernodes are uniformly sampled from the input nodes.
Node coordinates are embedded using a sinusoidal encoding followed by an MLP.
The supernodes aggregate features from nearby nodes using one-directional message passing and serve as tokens for subsequent transformer processing.
They are then processed by stack of DiT blocks, which condition the network on the simulation input parameters.
For prediction, we employ a DiT Perceiver \citep{jaegle2022perceiver} decoder that performs cross-attention between the latent representation and a set of query positions.
This allows the model to generate field predictions at arbitrary spatial locations, which is a desirable property for inference.
We sample 4096 supernodes and use a base dimension of 192. We use 8 DiT blocks for processing and 4 DiT Perceiver blocks for decoding.

\paragraph{GINO.}
GINO was proposed in \citep{li2023gino}.
Input coordinates are again embedded via sinusoidal encoding, after which the mesh is projected onto a regular latent grid.
This is achieved via message passing with connections generated via a radius graph.
On the latent grid, the conditioning is concatenated to the features at each grid point before Fourier Neural Operator (FNO) \cite{li2020fno} layers are employed.
Afterwards, features are mapped back onto the output grid by querying the latent grid, again via message passing.
Our implementation uses a latent grid of size $(16\times16\times16)$ with $16$ latent channels and a radius of 0.1 to construct the radius graph for message passing operations.
For our implementation, we use the library of the original authors.\footnote{\href{https://github.com/neuraloperator/neuraloperator}{https://github.com/neuraloperator/neuraloperator}}

\section{Experiments}
\label{app:experiments}
This section provides a detailed overview of the performed experiments for this benchmark. First, we explain the benchmarking setup used to generate the benchmarking results in detail in \cref{app:experimental_setup} and the evaluation procedure in \cref{app:evaluation_metrics}.
Furthermore, we provide information about training times for the presented methods in \cref{app:comp_resources}.

\subsection{Experimental Setup}
\label{app:experimental_setup}

\paragraph{Dataset Splits.}
We first split each dataset into source and target domains as outlined in \cref{sec:domain_splits}.
For the \emph{hot rolling}, \emph{sheet metal forming} and \emph{motor design} datasets, we use a 60\%/20\%/20\% split for training, validation, and testing in the source domains.
For the \emph{heatsink design} dataset, we use a 70\%/15\%/15\% split.
For target domains, where labels are unavailable during training in our \ac{UDA} setup, we use a 60\%/40\% split for training and test sets.
The large validation and test sets are motivated the industrial relevance of our benchmark, where reliable performance estimation on unseen data is a crucial factor.

\paragraph{Training Pipeline.}
For training, we use a dataset wide per field z-score normalization strategy, with statistics computed on the source domain training set.
We use a batch size of 16 and the AdamW optimizer~\citep{loshchilov2019adamw} with a weight decay of $10^{-5}$.
For the \emph{hot rolling}, \emph{sheet metal forming} and \emph{motor design} datasets, we use a learning rate of $10^{-3}$ while for the \emph{heatsink design} dataset we use $5\times10^{-4}$.
Gradients are clipped to a maximum norm of 1.
A cosine learning rate scheduler with a warmup of 100 epochs is employed.
For the large scale \emph{heatsink design} dataset, we enable Automatic Mixed Precision (AMP) and randomly subsample to 16,000 nodes per sample to reduce memory consumption and training time.
Additionally, we use Exponential Moving Average (EMA) updates with a decay factor of 0.95 to stabilize training.

Performance metrics are evaluated every 10 epochs, and we train all models for a maximum of 2000 epochs for the \emph{hot rolling}, \emph{sheet metal forming} and \emph{motor design} datasets, and 1500 epochs for the \emph{heatsink design} dataset.
Early stopping is applied if no improvement in the source domain validation loss is observed for 500 consecutive epochs.

\paragraph{Domain Adaptation Specifics.}
To enable \ac{UDA} algorithms, we jointly sample mini batches from the source and target domains at each training step and pass them thorugh the model.
Since target labels are not available, we compute supervised losses only on the source domain outputs.
In addition, we compute \ac{DA} losses on the latent representations of source and target domains in order to encourage domain invariance.

Since a crucial factor in the performance of \ac{UDA} algorithms is the choice of the domain adaptation loss weight $\lambda$ (see \cref{eq:principled_algorithm_problem_statement}), we perform extensive sweeps over this hyperparameter and select models using the unsupervised model selection strategies described in \cref{sec:model_selection}.
For all datasets, we sweep $\lambda$ logarithmically over $\lambda \in \{10^{-1}, 10^{-2}, \dots, 10^{-7}\}$.

\cref{tab:lambda_choices} provides an overview of the number of trained models for benchmarking performance of all models and all \ac{UDA} algorithms on the \emph{medium} difficulty domain shifts across all datasets.

\begin{table}[ht]
  \caption{Overview of the benchmarking setup and number of trained models across all datasets.}
  \label{tab:lambda_choices}
  \centering
  \resizebox{\textwidth}{!}{%
  \begin{tabular}{lccccc}
    \toprule
    \textbf{Dataset} & \textbf{Models} & \textbf{UDA algorithms} & \textbf{$\lambda$ values} & \textbf{\# seeds} & \textbf{\# models trained} \\
    \midrule
    \multirow{2}{*}{Rolling}   & \multirow{2}{*}{PointNet, GraphSAGE, Transolver} & Deep Coral, CMD, DANN, DARE-GRAM & $\{10^{-1};\,10^{-7}\}$ & 4 & 336 \\
              &                                 & w/o UDA               & --                      & 4 & 12 \\
    \midrule
    \multirow{2}{*}{Forming}   & \multirow{2}{*}{PointNet, GraphSAGE, Transolver} & Deep Coral, CMD, DANN, DARE-GRAM & $\{10^{-1};\,10^{-7}\}$ & 4 & 336 \\
              &                                 & w/o UDA               & --                      & 4 & 12 \\
    \midrule
    \multirow{2}{*}{Motor}     & \multirow{2}{*}{PointNet, GraphSAGE, Transolver} & Deep Coral, CMD, DANN, DARE-GRAM & $\{10^{-1};\,10^{-7}\}$ & 4 & 336 \\
              &                                 & w/o UDA               & --                      & 4 & 12 \\
    \midrule
    \multirow{2}{*}{Heatsink}  & \multirow{2}{*}{PointNet, Transover, UPT, GINO}                             & Deep Coral, CMD, DANN, DARE-GRAM & $\{10^{-1};\,10^{-7}\}$ & 4 & 448 \\
              &                                 & w/o UDA               & --                      & 4 & 16 \\
    \midrule
    \textbf{Sum} & & & & & \textbf{1,508} \\
    \bottomrule
  \end{tabular}
  }
\end{table}

\paragraph{Additional Details.}
For the three smaller datasets, we use smaller networks, while for the large scale \emph{heatsink design} dataset, we train larger model configurations to accommodate the increased data complexity.
An overview of model sizes along with average training times per dataset is provided in \cref{tab:training_times}.
We also refer to the accompanying code repository for a complete listing of all model hyperparameters, where we provide all baseline configuration files and detailed step by step instructions for reproducibility of our results.

Another important detail is that, during training on the \emph{heatsink design} dataset, we randomly subsample 16,000 nodes from the mesh in each training step to ensure computational tractability.
However, all reported performance metrics are computed on the full resolution of the data without any subsampling.

\subsection{Evaluation Metrics}
\label{app:evaluation_metrics}
\subsubsection{General Metrics}
\label{app:general_metrics}
We report the \ac{RMSE} for each predicted output field.
For field \( i \), the \ac{RMSE} is defined as:
\[
\text{RMSE}^{\text{field}}_i = \frac{1}{M} \sum_{m=1}^M \sqrt{ \frac{1}{N_m} \sum_{n=1}^{N_m} \left( y^{(i)}_{m,n} - \hat{y}^{(i)}_{m,n} \right)^2},
\]

where \( M \) is the number of test samples (graphs), \( N_m \) the number of nodes in graph \( m \), \( y^{(i)}_{m,n} \) the ground truth value of field \( i \) at node \( n \) of graph \( m \), and $\hat{y}^{(i)}_{m,n}$ the respective model prediction.

For aggregated evaluation, we define the total normalized \ac{RMSE} (nRMSE) as:
\[
\text{nRMSE} = \frac{1}{K} \sum_{i=1}^{K} \text{nRMSE}^{\text{field}}_i,
\]
where \( K \) is the number of predicted fields.
For this metric, all individual field errors are computed on normalized fields before aggregation.

In addition to the error on the fields, we report the mean Euclidean error of the predicted node displacement.
This is computed based on the predicted coordinates \( \hat{\mathbf{c}}_{m,n} \in \mathbb{R}^d \) and the ground truth coordinates \( \mathbf{c}_{m,n} \in \mathbb{R}^d \), where \( d \in \{2, 3\} \) is the spatial dimensionality, as follows:

\[
\text{RMSE}^{\text{deformation}} = \frac{1}{M} \sum_{m=1}^M \sqrt{\frac{1}{N_m} \sum_{n=1}^{N_m} \left\| \mathbf{c}_{m,n} - \hat{\mathbf{c}}_{m,n} \right\|_2}.
\]

\subsubsection{Physics Metrics}
\label{app:physics_metrics}

\paragraph{Von Mises stress consistency.}
For the \textit{hot rolling} and \textit{sheet metal forming} datasets, we predict both the relevant Cauchy stress tensor components and the von Mises equivalent stress.
This allows for an internal consistency check using the standard von Mises definition.
Under the assumed two-dimensional kinematics, where out of plane shear components are zero, the von Mises stress is given by
$$\sigma_{vM} = \sqrt{\frac{1}{2}\left[(\sigma_{11} - \sigma_{22})^2 + (\sigma_{22} - \sigma_{33})^2 + (\sigma_{33} - \sigma_{11})^2 + 6\tau_{12}^2\right]},$$
with $\sigma_{11}, \sigma_{22}, \sigma_{33}$ denoting the normal stresses and $\tau_{12}$ the in-plane shear stress.

We can recompute $\sigma_{vM}$ from the predicted tensor components and compare it to the predicted von Mises value using a normalized mean absolute error:
$$\mathcal{E}_{\text{vM\_consistency}} = \frac{\sum_{i=1}^{N} |\hat{\sigma}_{vM, i} - \hat{\sigma}_{vM_{recalc}, i}|}{\sum_{i=1}^{N} |\hat{\sigma}_{vM, i}|},$$
where the sum is taken over all nodes of a sample.

\paragraph{Constitutive law consistency I.}
For the \textit{sheet metal forming} dataset, the material is modeled as elastoplastic with von Mises plasticity and linear isotropic hardening.
This defines a yield surface, $\sigma_y$, which represents the material's current strength as a function of the equivalent plastic strain ($\varepsilon_p$):

$$\sigma_y(\varepsilon_p) = \sigma_{y0} + H\,\varepsilon_p,$$

where $\sigma_{y0}$ is the initial yield stress, H the hardening modulus, and $\varepsilon_p$ the equivalent plastic strain.

A physically-correct model must adhere to two conditions based on this law:
\begin{enumerate}
    \item Elastic nodes ($\varepsilon_p = 0$) must have a stress below this surface: $\sigma_{vM} \le \sigma_{y0}$.
    \item Plastic nodes ($\varepsilon_p > 0$) must have a stress on this surface: $\sigma_{vM} = \sigma_y(\varepsilon_p)$.
\end{enumerate}
Based on these two conditions, we introduce two metrics to evaluate the physical consistency of the predictions:
\begin{enumerate}
    \item Elastic violation rate (percentage of elastic nodes that incorrectly violate the initial yield stress):
    $$\mathcal{E}_{\text{viol\_plastic}}=
    \frac{1}{N_{\mathrm{el}}}
    \sum_{i \in \mathcal{E}}
    \mathbf{1}\!\left[\,\hat{\sigma}_{vM, i} > \sigma_{y0}\,\right],
    $$
    where $\mathcal{E}$ is the set points in the elastic regime and $N_{\text{el}} = |\mathcal{E}|$ is the number of elastic nodes.
    \item Plastic Law Residual (nMAE for all plastic nodes):
    $$
    \mathcal{E}_{\text{res\_plastic}} = \frac{1}{N_{\mathrm{pl}}} \sum_{i \in \mathcal{P}}
    \frac{\lvert \hat{\sigma}_{vM, i} -
    \bigl( \sigma_{y0} + H\,\hat{\varepsilon}_{p, i} \bigr)\rvert}{\sigma_{y0},
    }
    $$
    where $\mathcal{P}$ is the set of points in the plastic regime and $N_{\text{pl}} = |\mathcal{P}|$ is the number of plastic nodes.
\end{enumerate}

\paragraph{Constitutive law consistency II.}
Since all models directly predict both stress and strain fields for the \emph{electric motor design}, we asses the physical consistency of the predictions by evaluating a constitutive error based on linear elastic Hooke’s law used by the numerical simulator.
For each simulation sample, the Cauchy stress tensor $\boldsymbol{\sigma}_{\text{recalc}}$ is reconstructed from the predicted logarithmic strain tensor $\boldsymbol{\varepsilon}_{\text{pred}}$ using the corresponding linear elastic constitutive relation, and compared against the predicted Cauchy stress tensor $\hat{\boldsymbol{\sigma}}$.

Two material models are considered, consistent with the simulation setup.
For the rotor core, a plane stress formulation is employed, for which $\sigma_{zz}=0$ and the in-plane stresses are given by
\[
\sigma_{xx} = \frac{E}{1-\nu^2}(\varepsilon_{xx} + \nu \varepsilon_{yy}), \quad
\sigma_{yy} = \frac{E}{1-\nu^2}(\varepsilon_{yy} + \nu \varepsilon_{xx}), \quad
\sigma_{xy} = 2G \varepsilon_{xy},
\]
with Young’s modulus $E$, Poisson’s ratio $\nu$, and shear modulus $G = \tfrac{E}{2(1+\nu)}$.
For the permanent magnet regions, a plane strain formulation is used, where $\varepsilon_{zz}=0$ and the stresses are given by
\[
\sigma_{xx} = \frac{E}{(1+\nu)(1-2\nu)}\big((1-\nu)\varepsilon_{xx} + \nu \varepsilon_{yy}\big), \quad
\sigma_{yy} = \frac{E}{(1+\nu)(1-2\nu)}\big((1-\nu)\varepsilon_{yy} + \nu \varepsilon_{xx}\big),
\]
\[
\sigma_{xy} = 2G \varepsilon_{xy}, \quad
\sigma_{zz} = \nu(\sigma_{xx} + \sigma_{yy}).
\]

Based on these relations, a relative constitutive error is defined as the normalized $L^2$-norm of the stress discrepancy,
\[
\mathcal{E}_{\text{const}} =
\frac{\left\lVert \hat{\boldsymbol{\sigma}}_{\text{recalc}} - \hat{\boldsymbol{\sigma}} \right\rVert_2}
     {\left\lVert \hat{\boldsymbol{\sigma}} \right\rVert_2},
\]
where the norm is taken over all nodes and relevant stress components of a sample.
This metric validates the internal consistency of the predicted stress and strain fields under the assumption of the linear elastic constitutive law.

\paragraph{Boundary condition satisfaction.}
The \textit{heatsink design} simulations impose two important Dirichlet Boundary Conditions (BCs) on the fin surfaces: no slip velocity and the solid temperature of the fins.
Therefore we define the two following errors to measure the violation of these BCs for our surrogates:

$$
\mathcal{E}_{\text{BC\_viol\_T}}
= \frac{1}{N_{\text{fin}}}
\sum_{i \in \mathcal{F}}
\frac{\left| \hat{T}_{i} - T_{\text{solid}} \right|}{\left| T_{\text{solid}} - T_{\text{env}} \right|},
$$
and
$$
\mathcal{E}_{\text{BC\_viol\_u}} = \frac{1}{N_{\text{fin}}}
\sum_{i \in \mathcal{F}}
\left\| \hat{\mathbf{u}}_{i} \right\|_2,
$$
where $\mathcal{F}$ is the set of fin nodes, $N_{\text{fin}} = |\mathcal{F}|$ is the number of fin nodes, and $\hat{T}_{i}$ and $\hat{\mathbf{u}}_{i}$ are the respective predictions for temperature and velocity at node $i$.

\subsubsection{Custom (Engineering) Metrics}
\label{app:custom_metrics}
In practice, engineers look at simulation metrics in distinct regions in the domain.
As outlined in \crefrange{sec:rolling}{sec:heatsink}, the most important regions for the presented problems reduce to chords.
To obtain an interpretable error metric aligned with expert practice, we define a custom error computed along the selected chords as
\[
\mathcal{E}_{\text{custom}} =
\frac{1}{N_{\text{chord}}}
\sum_{i \in \mathcal{C}}
\frac{\left|\hat{y}_i-y_i\right|}{y_i + \epsilon}
,
\]
where \(\mathcal{C}\) denotes the set of nodes along the selected chords or slice,
\(N_{\mathcal{C}} = |\mathcal{C}|\) is its cardinality,
\(\hat{y}_i\) and \(y_i\) are the predicted and ground-truth (both denormalized) field values at node \(i\), respectively,
and \(\varepsilon > 0\) is a small constant added for numerical stability.

\subsection{Computational Resources and Timings}
\label{app:comp_resources}
While generating the results reported on the \emph{medium} difficulty level of our benchmark, we measured average training times per dataset and model architecture.
All models were trained and timed on a single NVIDIA A100 64GB GPU for a fair comparison.
While the total compute budget is difficult to estimate due early stopping, we provide a detailed analysis of the average training times for 2000 epochs (\emph{hot rolling}, \emph{sheet metal forming} and \emph{motor design}) 1500 epochs (\emph{heatsink design}) in \cref{tab:training_times}.

This table refers to models trained with Deep CORAL, however different UDA algorithms do not add significant computational cost.
What is more impactful concerning the full pipeline (including model selection) is the number of hyperparameter variations. The total cost of one UDA algorithm \& model selection pipeline can be estimated by multiplying the average training time by the number of trained models (e.g. $\times 7$ if one sweeps over 7 hyperparameters of $\lambda$), for sequential execution.
Furthermore, the model selection method's runtime training is negligible compared to the training times.

\begin{table}[ht]
  \caption{Average training times (averaged for 2000 epochs for \emph{hot rolling}, \emph{sheet metal forming} and \emph{electric motor design}, and 1500 epochs for \emph{heatsink design}) and parameter counts for each model on the \emph{medium} difficulty benchmark tasks. Times are measured on an A100 64GB GPU using a batch size of 16.}
  \label{tab:training_times}
  \centering
  \begin{tabular}{l c c l c c}
    \toprule
    \textbf{Dataset} & \textbf{\# Training samples} & \textbf{Avg. \# nodes} & \textbf{Model} & \textbf{\# Parameters} & \textbf{Avg. training time (h)} \\
    \midrule
    \multirow{3}{*}{Rolling} & \multirow{3}{*}{2,463} & \multirow{3}{*}{508} & PointNet & 0.3M & 1.22 \\
     &  &  & GraphSAGE & 0.2M & 3.49 \\
     &  &  & Transolver & 0.57M & 2.41 \\
    \midrule
    \multirow{3}{*}{Forming} & \multirow{3}{*}{1,919} & \multirow{3}{*}{9,080} & PointNet & 0.3M & 6.45 \\
     &  &  & GraphSAGE & 0.2M & 33.24 \\
     &  &  & Transolver & 0.57M & 9.90 \\
    \midrule
    \multirow{3}{*}{Motor} & \multirow{3}{*}{1,440} & \multirow{3}{*}{4,846} & PointNet & 0.3M & 2.63 \\
     &  &  & GraphSAGE & 0.2M & 13.57 \\
     &  &  & Transolver & 0.57M & 4.25 \\
    \midrule
    \multirow{4}{*}{Heatsink} & \multirow{4}{*}{277} & \multirow{4}{*}{16,000} & PointNet & 1.08M & 11.83 \\
     &  &  & Transolver & 4.07M & 13.47 \\
     &  &  & UPT & 5.77M & 12.41 \\
     &  &  & GINO & 2.5M & 14.99 \\
    \bottomrule
  \end{tabular}
\end{table}

\section{Dataset Details}
\label{app:dataset_generation}

\subsection{Hot Rolling}
\label{app:rolling_detailed}
The \emph{hot rolling} dataset represents a hot rolling process in which a metal slab undergoes plastic deformation to form a sheet metal product.
The model considers a plane-strain representation of a heated slab segment with a core temperature $T_{\text{core}}$ and a surface temperature $T_{\text{Surf}}$, initially at thickness $t$, passing through a simplified roll stand with a nominal roll gap $g$ (see \cref{fig:rolling_sketch}).
This roll gap effectively matches the exit thickness of the workpiece. Given the material properties, the initial temperature distribution over the slab thickness and the specified pass reduction, the model aims to capture the evolution of the thermo-mechanical state of the workpiece as it traverses the roll gap.

To reduce computational complexity, the analysis is confined to the vertical midplane along the rolling direction based on a plane-strain assumption.
This is well justified by the high width-to-thickness ratio characteristic of the workpiece. Additionally, vertical symmetry is also exploited.
Consequently, only the upper half of the workpiece and the upper work roll are modeled.

The workpiece is discretized using plane-strain, reduced-integration, quadrilateral elements (Abaqus element type CPE4RT).
Mesh generation is fully automated, with the element size calibrated according to findings from a mesh convergence study.
In terms of mechanical behavior, the workpiece is modeled as elasto-plastic with isotropic hardening, employing tabulated flow curves representative for a titanium alloy \citep{lesuer2000, lu2018}.
The elastic modulus and flow stress are temperature dependent, with the latter also influenced by the plastic strain rate.
In contrast, material density and Poisson’s ratio are assumed to remain constant.
The work roll with a diameter of 1000 mm is idealized as an analytically defined rigid body.

In addition to the mechanical behavior, the elements also feature a temperature degree of freedom that captures thermal phenomena, which are in turn fully coupled with the mechanical field.
Heat conduction within the workpiece is governed by temperature dependent thermal conductivity and specific heat capacity.
Heat transfer at the interface between the workpiece and the roll is modeled as proportional to the temperature difference between the contacting surfaces, using a heat transfer coefficient of 5 $\mathrm{mW/mm^2K}$.
The model also accounts for internal heat generation due to plastic deformation, based on the standard assumption that 90\% of plastic work is converted into heat.
Additionally, all frictional energy is assumed to be fully transformed into heat and evenly divided between the workpiece and the roll.
However, since the analysis focuses on the workpiece, only the portion of this heat entering the workpiece is considered.

The \ac{FE} simulation is performed with the \textit{Abaqus} explicit solver using a relatively high mass scaling factor of 100.
This mass scaling proved to be a suitable choice for maintaining both computational efficiency and solution accuracy. 
The pre-processing, evaluation and post-processing of the simulations was automated in Python.
To generate the dataset, we employ \ac{LHS} to sample the parameter ranges specified in \cref{tab:rolling_params}.
Simulation outputs from Abaqus (.odb files) were converted to a more suitable .h5 format in post-processing, enabling seamless integration into the \ourmethod~framework.
All simulations were run on a Gigabyte Aorus 15P KD consumer laptop equipped with an Intel Core i7-11800H CPU (8 cores, 16 threads, 2.30–4.60 GHz), 16 GB DDR4 RAM at 3200 MHz and a 1 TB NVMe SSD.
The single-core CPU time for one simulation was 25 seconds on average, depending on the mesh size and convergence speed.

\begin{table}[h!]
  \caption{Ranges of the varied input parameters for the \emph{hot rolling} simulations.
  Samples are generated using \ac{LHS}.}
  \label{tab:rolling_params}
  \centering
  \begin{tabular}{llcc}
    \toprule
    \textbf{Parameter}  & \textbf{Description}   & \textbf{Min} & \textbf{Max} \\
    \midrule
    $t$ $(mm)$                 & Initial slab thickness.          & 50.0   & 200 \\
    $\text{reduction}$ $(-)$  & Reduction of initial slab thickness. & 0.01  & 0.15 \\
    $T_{\text{core}}$ $(^\circ C)$    & Core slab temperature. & 900  & 1100 \\
    $T_{\text{surf}}$ $(^\circ C)$  & Surface slab temperature. & 900  & 1100 \\
    \bottomrule
  \end{tabular}
\end{table}

\subsection{Sheet Metal Forming}
\label{app:forming_detailed}
\begin{figure}[htbp]
\centering
\includegraphics[width=.9\textwidth, keepaspectratio]{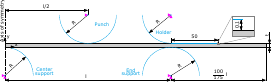}
\caption{Bending process abstraction, initial configuration.}
\label{fig:fe_model_initial}
\end{figure}

For the \emph{sheet metal forming} dataset, a w-shaped bending process was selected due to its complex contact interactions and the highly nonlinear progression of bending forces.
For this purpose, a parameterized 2D \ac{FE} model of the process was developed using the commercial \ac{FEM} software \textit{Abaqus}\ and its implicit solver, with the simulation pipeline implemented in Python.
The initial configuration of the finite element model is shown in \cref{fig:fe_model_initial} and described below.

Due to geometric and loading symmetry, only the right half of the sheet with a thickness~$t$ was modeled. The die and punch were idealized as rigid circular segments with a shared radius~$r$.
Additionally, a rigid blank holder comprising an arc and a straight segment was positioned $0.1\ \mathrm{mm}$ above the sheet to maintain contact and restrain vertical motion.
The required sheet length was determined by the support span~$l$, enabling material flow toward the center in response to the downward motion of the punch.

The sheet was discretized using bilinear, plane-strain quadrilateral elements with reduced integration and hourglass control (Abaqus element type CPE4R).
A prior mesh convergence study indicated that accurate simulation results require a minimum of 10 element rows across the sheet thickness.

The sheet material was modeled as elastoplastic with von Mises plasticity and linear isotropic hardening.
The following properties were assigned: Young’s modulus of 210 GPa, Poisson’s ratio of 0.3, yield stress of 410 MPa, and hardening modulus of 2268 MPa.

For all contact interfaces, a normal contact formulation with surface-to-surface discretization, penalty enforcement, and finite-sliding tracking was employed.
Tangential contact was modeled via a Coulomb friction law with a coefficient $\mu$.

\begin{figure}[htbp]
\centering
\includegraphics[width=.9\textwidth, keepaspectratio]{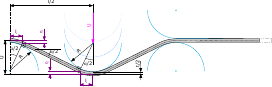}
\caption{Bending process abstraction, deformed configuration.}
\label{fig:fe_model_deformed}
\end{figure}

The supports and blank holder were fixed by constraining horizontal and vertical translations as well as in-plane rotations.
These constraints were applied at the centroid of each arc segment, representing the reference point for the respective rigid body.
The punch was similarly constrained against horizontal movement and rotation but retained vertical mobility.
The deformed configuration following a vertical displacement $U$ of the punch is illustrated in \cref{fig:fe_model_deformed}.

For dataset generation, \ac{LHS} was used to sample from the parameters outlined in \cref{tab:forming_params}.
As for the hot rolling simulations, outputs from Abaqus (.odb files) were converted to .h5 format in post-processing, to integrate them into the \ourmethod~framework.
All simulations were run on a Gigabyte Aorus 15P KD consumer laptop equipped with an Intel Core i7-11800H CPU (8 cores, 16 threads, 2.30–4.60 GHz), 16 GB DDR4 RAM at 3200 MHz and a 1 TB NVMe SSD.
The single-core CPU time for one simulation run was 300 seconds on average, depending on mesh size and convergence speed.

\begin{table}[h!]
  \caption{Ranges of the varied input parameters for the \emph{sheet metal forming} simulations.
  Samples are generated using \ac{LHS}.}
  \label{tab:forming_params}
  \centering
  \begin{tabular}{llcc}
    \toprule
    \textbf{Parameter}    & \textbf{Description}  & \textbf{Min}     & \textbf{Max}\\
    \midrule
    $r$ $(mm)$    & Roll radius.    & 10.0   & 50.0   \\
    $t$ $(mm)$    & Sheet thickness.       & 2.0    & 5.0    \\
    $l$ $(mm)$    & Sheet length.        & 175.0    & 350.0  \\
    $\mu$ $(-)$   & Friction coefficient between the sheet and the rolls.           & 0.1    & 0.5   \\
    \bottomrule
  \end{tabular}
\end{table}

\subsection{Electric Motor Design}
\label{app:motor_detailed}

\begin{figure}[h]
    \centering
    \includegraphics[width=0.8\textwidth]{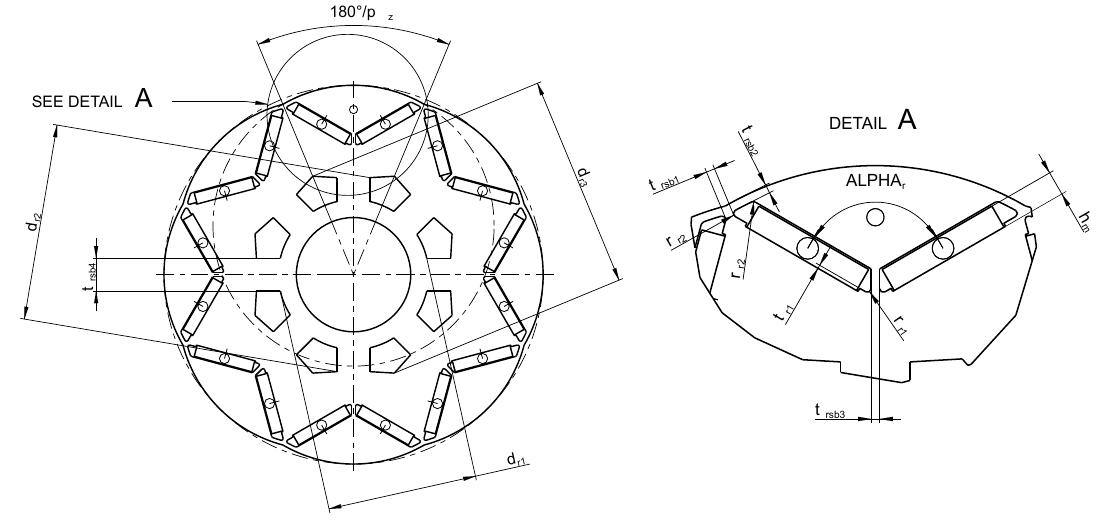}
    \caption{Technical drawing of the electrical motor. Sampling ranges for the shown parameters can be found in \cref{tab:motor_params}.}
    \label{fig:motor_technical_drawing}
\end{figure}

The \emph{electric motor design} dataset includes a structural \ac{FE} simulation of a rotor within electric machinery, subjected to mechanical loading at burst speed.
The rotor topology is modeled after the motor architecture of the 2010 Toyota Prius \citep{burress2011}, an industry-recognized benchmark frequently used for validation and comparison in academic and industrial research.
The Prius rotor topology is based on a V-shaped magnet configuration as shown in \cref{fig:motor_technical_drawing}.

Structural rotor simulations are essential in multi-physics design optimization, where motor performance is evaluated across multiple domains including electromagnetic, thermal, acoustic, and structural.
Using a design optimization framework, stator and rotor design are iteratively refined to identify Pareto-optimal solutions based on objectives such as efficiency, torque, weight, and speed.
In this process, the structural FE model predicts stress and deformation due to loading ensuring the rotor’s structural integrity.

The set up and execution of the structural simulations for this dataset are automated and implemented in the open source design optimization framework \textit{SyMSpace}\footnote{\href{https://symspace.lcm.at/}{https://symspace.lcm.at/}}.
The \ac{FE} simulation of the rotor is performed using a mixed 2D plane stress and plane strain formulation with triangular elements.
To enhance computational efficiency, geometric symmetry is exploited and only a 1/16 sector of the full rotor is modeled.
The mechanical simulation is static and evaluates the rotor under centrifugal loading, incorporating press-fit conditions between the rotor core and shaft, as well as contact interactions between the rotor core and embedded magnets.

An elastic material behavior is employed for all components, including the rotor core, shaft, and magnets. Material properties are summarized in \cref{tab:material_properties}.
Based on the parametrized CAD model of the rotor topology, the geometry is automatically meshed using \textit{Netgen}\footnote{\href{https://ngsolve.org/}{https://ngsolve.org/}}.
The design optimization tool also automatically identifies nodes for boundary conditions and contact surfaces and applies the corresponding constraints and interactions required for the simulation.
The implicit \ac{FE} solver \textit{HOTINT} is used to compute the quasi-static response of the system, providing local stress and strain fields across the rotor topology.

\begin{table}[h]
  \caption{Material parameters for the structural \emph{electric motor design} simulations.}
  \label{tab:material_properties}
  \centering
  \begin{tabular}{lcccc}
    \toprule
      & \textbf{Rotor Core}                      & \textbf{Rotor Shaft}      & \textbf{Permanent Magnet}    \\
    \midrule
    Material      & NO27-14 Y420HP       & 42CrMo4      & BMN-40SH  \\
    Density ($\mathrm{kg/dm^3}$)     & 7.6               & 7.72        &  7.55  \\
     Possions ratio (-) & 0.29               &  0.3       &  0.24  \\
    Young's Modulus ($\mathrm{kN/mm^2}$)         & 185.0               & 210.0        &  175.0  \\
    Tensile Strength ($\mathrm{kN/mm^2}$) & 550.0               & 850.0        &  250.0  \\
    \bottomrule
  \end{tabular}
\end{table}

To generate the electric motor dataset, a comprehensive motor optimization study was conducted using \textit{SyMSpace}, based on design specifications of the 2010 Toyota Prius.
The optimization aimed to minimize multiple performance metrics, including motor mass, material costs, rotor torque ripple, motor losses, coil temperature, stator terminal current, and elastic rotor deformation.
A genetic algorithm was employed to explore the design space and identify Pareto-optimal solutions.
In the process, 3,196 motor configurations were evaluated by varying, among other factors, the rotor’s topological parameters within the bounds specified in \cref{tab:motor_params}.
The outputs of the structural simulations were generated in .vtk format and then stored in .h5 files, allowing direct integration into the \ourmethod~framework.
Each structural simulation required approximately 4 to 5 minutes of single-core CPU time on a Intel Core i9-14900KS processor (24 Cores, 3200 MHz), depending on convergence speed of the contact algorithm.

\begin{table}[h]
  \caption{Input parameters for the \emph{electric motor design} simulations. The dataset was created by a design space exploration using a genetic algorithm.}
  \label{tab:motor_params}
  \centering
  \begin{tabular}{llccc}
    \toprule
    \textbf{Parameter}  & \textbf{Description}                       & \textbf{Min}      & \textbf{Max}    \\
    \midrule
    $d_{si}$ $(mm)$      & Stator inner diameter.       & 150.0      & 180.0 \\
    $h_m$ $(mm)$         & Magnet height.               & 6.0        & 9.0   \\
    $\alpha_r$ $(^\circ)$   & Angle between magnets.       & 120.0      & 160.0 \\
    $t_{r1}$ $(mm)$      & Magnet step.                 & 1.0        & 5.0   \\
    $r_{r1}$ $(mm)$      & Rotor slot fillet radius 1.  & 0.5        & 2.5   \\
    $r_{r2}$ $(mm)$      & Rotor slot fillet radius 2.  & 0.5        & 3.5   \\
    $r_{r3}$ $(mm)$      & Rotor slot fillet radius 3.  & 0.5        & 5.0   \\
    $r_{r4}$ $(mm)$      & Rotor slot fillet radius 4.  & 0.5        & 3.0   \\
    $t_{rsb1}$ $(mm)$    & Thickness saturation bar 1.  & 4.0        & 12.0  \\
    $t_{rsb2}$ $(mm)$    & Thickness saturation bar 2.  & 1.0        & 3.0   \\
    $t_{rsb3}$ $(mm)$    & Thickness saturation bar 3.  & 1.2        & 4.0   \\
    $t_{rsb4}$ $(mm)$    & Thickness saturation bar 4.  & 5.0        & 12.0  \\
    $d_{r1}$ $(mm)$      & Rotor slot diameter 1.       & 60.0       & 80.0  \\
    $d_{r2}$ $(mm)$      & Rotor slot diameter 2.       & 80.0       & 120.0 \\
    $d_{r3}$ $(mm)$      & Rotor slot diameter 3.      & 100.0      & 125.0 \\
    \bottomrule
  \end{tabular}
\end{table}

\newpage
\subsection{Heatsink Design}
\label{app:heatsink_detailed}

\begin{figure}[h]
    \centering
    \includegraphics[width=\textwidth]{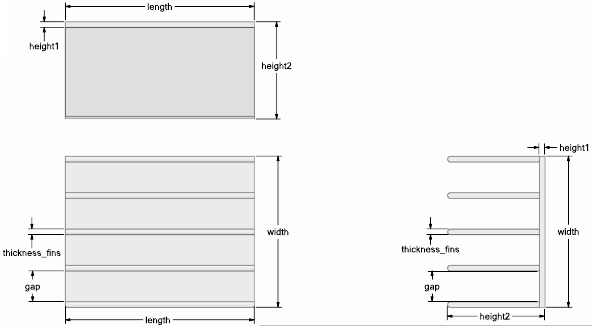}
    \caption{Technical drawing of the solid body in the \emph{heatsink design} dataset. Some of the shown parameters are varied for data generation (see \cref{tab:heatsink_params}).}
    \label{fig:heatsink_technical_drawing}
\end{figure}

The \emph{heatsink design} dataset consists of heatsink geometries similar to the example shown in \cref{fig:heatsink_technical_drawing}, placed centrally at the bottom of a surrounding box-shaped domain filled with air.
The dimensions of the surrounding enclosure are 0.14 m × 0.14 m × 0.5 m (length × width × height).

The geometric configuration of each heatsink is defined by several parameters, which were varied within specified bounds for the design study. These parameters and their corresponding value ranges are summarized in \cref{tab:heatsink_params}.
A total of 512 simulation cases were generated, with non-uniform sampling across the parameter space.

\begin{table}[h]
  \caption{Geometric and physical parameters of the \emph{heatsink design} simulations.
  In total, 512 simulations were performed.}
  \label{tab:heatsink_params}
  \centering
  \begin{tabular}{llccc}
    \toprule
    \textbf{Parameter}  & \textbf{Description}                     & \textbf{fixed Value}  & \textbf{Min}      & \textbf{Max}    \\
    \midrule
    length $(\mathrm{m})$      & Heatsink length                 & 0.1 & - & - \\
    width $(\mathrm{m})$       & Heatsink width                & 0.08 & - & -\\
    height1 $(\mathrm{m})$     & Baseplate height                 & 0.003 & - & -\\
    T(amb) $(\mathrm{K})$ & Ambient Temperature  & 300 & - & -\\
    fins $(-)$             & Number of fins  & -      & 5      & 14 \\
    gap $(\mathrm{m})$              & Gap between fins       & -        & 0.0023       & 0.01625   \\
    thickness\char`_fins $(\mathrm{m})$   &Thickness of fins   & -   & 0.003  & 0.004     \\
    height2 $(\mathrm{m})$          & Heatsink height    & -   & 0.053      & 0.083 \\
    T (solid) $(\mathrm{K})$        & Temperature of the solid fins    & -       & 340        & 400   \\
    \bottomrule
  \end{tabular}
\end{table}

The dataset was generated using \ac{CFD} simulations based on the \ac{RANS} equations coupled with the energy equation.
All simulations were conducted in the open-source \ac{CFD} suite \textit{OpenFOAM 9}.

The computational domain was discretized using a finite volume method with second-order spatial discretization schemes.
A structured hexahedral background mesh was generated with the blockMesh utility in OpenFOAM, followed by mesh refinement using snappyHexMesh to accurately resolve the heatsink structure defined in STL format.

To simulate buoyancy driven natural convection, the buoyantSimpleFoam solver was employed.
This solver is designed for steady state, compressible, buoyant flows, using the SIMPLE algorithm for pressure-momentum coupling, extended with under relaxation techniques to enhance numerical stability and robust convergence.

Boundary conditions were applied as follows:
\begin{itemize}
    \item Walls of the surrounding: no-slip velocity condition with fixed ambient temperature as defined in \cref{tab:heatsink_params}.
    \item Walls of the heatsink: no-slip velocity condition with solid temperature within the range specified for parameter T (solid) in \cref{tab:heatsink_params}.
\end{itemize}

Given the turbulent nature of the flow, the \ac{RANS} equations were closed using the SST k–$\omega$ turbulence model \citep{menter2003sst}.
Near-wall regions were modeled using a $y^{+}$-insensitive near-wall treatment, allowing accurate resolution of boundary layers without the need for excessively fine meshes.

A mesh convergence study was conducted to ensure numerical accuracy.
Depending on mesh resolution, each simulation required approximately 11 to 18 hours of single-core CPU time on an Intel Core i9-14900KS processor (24 cores, 3.2 GHz).

\FloatBarrier
\newpage
\section{Ablation Studies}
\label{app:ablation_studies}
In the following sections, we present ablations on the \ourmethod~framework.

\subsection{Geometric Encoding}
\label{app:geometric_pointnet}
The design concept of \ourmethod~is to allow plug-in integration of any \ac{UDA} algorithm and model architecture, as long as the model can be conditioned in some way (see \cref{fig:figure_1}).
However, explicitly conditioning models on scalar geometric parameters is not the only option: for instance, domain-specific information may be encoded implicitly in the mesh itself.
To investigate this, we provide an ablation in which the model encodes the mesh directly and is not explicitly conditioned on the scalar parameters.
Specifically, we replace the feed-forward conditioning network with a geometric PointNet based encoder to embed the input mesh into a global latent vector, on which \ac{UDA} is then performed.

We report results of this setup on the \textit{electric motor design} dataset.
The setup follows the benchmarking procedure described in \cref{sec:benchmarking_setup} and \cref{app:experimental_setup}: for each \ac{UDA} algorithm, we train across seven different regularizer strengths and four random seeds.

\begin{table}[h]
  \centering
  \caption{Performance metrics on the source and target domains (mean $\pm$ std over 4 seeds) for the \emph{electric motor design} task at \textit{medium} difficulty, when using a PointNet-based geometry encoder. RMSE is reported for all global metrics (All Fields Normalized Avg through Logarithmic Strain), while further columns report custom engineering and physics-based metrics. \textbf{Bold} indicates the overall best combination of \ac{UDA} algorithm and model selection. The unregularized baseline is shaded \tightbox{baselinerow}{beige}, and the best \ac{UDA} configuration is \underline{underlined} and shaded \tightbox{bestrow}{green}.}
  \label{tab:geometric_pointnet}
  \resizebox{\textwidth}{!}{%
  \definecolor{bestrow}{HTML}{DFF0D8}
  \definecolor{baselinerow}{HTML}{FFF4CC}
  \begin{tabular}{lllcccccccccccc}
    \toprule
    \multirow{2}{*}{\textbf{Model}} & \multirow{2}{*}{\makecell{\textbf{DA}\\ \textbf{Algorithm}}} & \multirow{2}{*}{\makecell{\textbf{Model}\\ \textbf{Selection}}} & \multicolumn{2}{c}{\textbf{All Fields Normalized Avg (-)}} & \multicolumn{2}{c}{\textbf{Deformation (m)}} & \multicolumn{2}{c}{\textbf{Cauchy Stress (MPa)}} & \multicolumn{2}{c}{\textbf{Logarithmic Strain ($\mathbf{\times 10^{-2}}$)}} & \multicolumn{2}{c}{\textbf{Rel Custom Error (-)}} & \multicolumn{2}{c}{\textbf{Constitutive Error ($\mathbf{\times 10^{-2}}$)}} \\
\cmidrule(lr){4-5} \cmidrule(lr){6-7} \cmidrule(lr){8-9} \cmidrule(lr){10-11} \cmidrule(lr){12-13} \cmidrule(lr){14-15}
      &   &  & \textbf{SRC} & \textbf{TGT} & \textbf{SRC} & \textbf{TGT} & \textbf{SRC} & \textbf{TGT} & \textbf{SRC} & \textbf{TGT} & \textbf{SRC} & \textbf{TGT} & \textbf{SRC} & \textbf{TGT} \\
    \midrule
    \multirow{17}{*}{\underline{\textbf{PointNetGeometry}}} & \cellcolor{baselinerow}- & \cellcolor{baselinerow}- & \cellcolor{baselinerow}$0.278(\pm0.010)$ & \cellcolor{baselinerow}$0.363(\pm0.016)$ & \cellcolor{baselinerow}$0.001(\pm0.000)$ & \cellcolor{baselinerow}$0.001(\pm0.000)$ & \cellcolor{baselinerow}$12.211(\pm0.510)$ & \cellcolor{baselinerow}$16.294(\pm0.810)$ & \cellcolor{baselinerow}$0.703(\pm0.032)$ & \cellcolor{baselinerow}$0.945(\pm0.052)$ & \cellcolor{baselinerow}$0.296(\pm0.014)$ & \cellcolor{baselinerow}$0.309(\pm0.006)$ & \cellcolor{baselinerow}$0.356(\pm0.024)$ & \cellcolor{baselinerow}$0.383(\pm0.052)$ \\
    \cmidrule(lr){2-15}
    & DANN & DEV & $0.215(\pm0.006)$ & $0.281(\pm0.007)$ & $0.001(\pm0.000)$ & $0.001(\pm0.000)$ & $11.728(\pm0.362)$ & $15.613(\pm0.403)$ & $0.673(\pm0.023)$ & $0.902(\pm0.026)$ & $0.282(\pm0.007)$ & $0.268(\pm0.008)$ & $0.379(\pm0.049)$ & $0.432(\pm0.085)$ \\
    & DANN & IWV & $0.215(\pm0.004)$ & $0.281(\pm0.008)$ & $0.002(\pm0.000)$ & $0.002(\pm0.001)$ & $11.717(\pm0.256)$ & $15.670(\pm0.456)$ & $0.672(\pm0.016)$ & $0.904(\pm0.028)$ & $0.283(\pm0.012)$ & $0.267(\pm0.016)$ & $0.346(\pm0.045)$ & $0.387(\pm0.060)$ \\
    & DANN & SB & $0.212(\pm0.002)$ & $0.274(\pm0.004)$ & $0.001(\pm0.000)$ & $0.001(\pm0.000)$ & $11.509(\pm0.111)$ & $15.213(\pm0.209)$ & $0.660(\pm0.006)$ & $0.879(\pm0.009)$ & $0.280(\pm0.010)$ & $0.261(\pm0.012)$ & $0.346(\pm0.067)$ & $0.387(\pm0.106)$ \\
    & DANN & TB & $0.214(\pm0.004)$ & $0.273(\pm0.002)$ & $0.001(\pm0.000)$ & $0.001(\pm0.000)$ & $11.630(\pm0.217)$ & $15.177(\pm0.168)$ & $0.667(\pm0.012)$ & $0.876(\pm0.007)$ & $0.282(\pm0.013)$ & $0.256(\pm0.011)$ & $0.335(\pm0.030)$ & $0.354(\pm0.032)$ \\
    \cmidrule(lr){2-15}
    & CMD & DEV & $0.356(\pm0.037)$ & $0.440(\pm0.033)$ & $0.001(\pm0.001)$ & $0.001(\pm0.001)$ & $20.155(\pm1.984)$ & $25.448(\pm1.734)$ & $1.160(\pm0.111)$ & $1.475(\pm0.097)$ & $0.550(\pm0.016)$ & $0.459(\pm0.004)$ & $0.277(\pm0.061)$ & $0.273(\pm0.060)$ \\
    & CMD & IWV & $0.216(\pm0.010)$ & $0.276(\pm0.010)$ & $0.001(\pm0.000)$ & $0.001(\pm0.000)$ & $11.799(\pm0.656)$ & $15.441(\pm0.662)$ & $0.678(\pm0.041)$ & $0.893(\pm0.040)$ & $0.270(\pm0.003)$ & $0.262(\pm0.018)$ & $0.335(\pm0.033)$ & $0.362(\pm0.057)$ \\
    & CMD & SB & $0.211(\pm0.002)$ & $0.269(\pm0.010)$ & $0.001(\pm0.000)$ & $0.001(\pm0.000)$ & $11.444(\pm0.114)$ & $14.870(\pm0.595)$ & $0.655(\pm0.008)$ & $0.857(\pm0.036)$ & $0.271(\pm0.003)$ & $0.254(\pm0.021)$ & $0.342(\pm0.032)$ & $0.377(\pm0.047)$ \\
    & CMD & TB & $0.211(\pm0.002)$ & $0.269(\pm0.010)$ & $0.001(\pm0.000)$ & $0.001(\pm0.000)$ & $11.444(\pm0.114)$ & $14.870(\pm0.595)$ & $0.655(\pm0.008)$ & $0.857(\pm0.036)$ & $0.271(\pm0.003)$ & $0.254(\pm0.021)$ & $0.342(\pm0.032)$ & $0.377(\pm0.047)$ \\
    \cmidrule(lr){2-15}
    & DARE-GRAM & DEV & $0.292(\pm0.095)$ & $0.364(\pm0.107)$ & $0.001(\pm0.000)$ & $0.001(\pm0.000)$ & $16.280(\pm5.626)$ & $20.645(\pm6.488)$ & $0.935(\pm0.324)$ & $1.193(\pm0.378)$ & $0.417(\pm0.163)$ & $0.367(\pm0.109)$ & $0.292(\pm0.038)$ & $0.310(\pm0.059)$ \\
    & DARE-GRAM & IWV & $0.217(\pm0.009)$ & $0.276(\pm0.011)$ & $0.001(\pm0.000)$ & $0.001(\pm0.000)$ & $11.857(\pm0.594)$ & $15.365(\pm0.714)$ & $0.680(\pm0.038)$ & $0.888(\pm0.045)$ & $0.283(\pm0.009)$ & $0.274(\pm0.007)$ & $0.324(\pm0.025)$ & $0.356(\pm0.020)$ \\
    & \cellcolor{bestrow}\underline{\textbf{DARE-GRAM}} & \cellcolor{bestrow}\underline{\textbf{SB}} & \cellcolor{bestrow}$0.210(\pm0.002)$ & \cellcolor{bestrow}$\underline{\mathbf{0.266(\pm0.001)}}$ & \cellcolor{bestrow}$0.001(\pm0.000)$ & \cellcolor{bestrow}$0.001(\pm0.000)$ & \cellcolor{bestrow}$11.341(\pm0.089)$ & \cellcolor{bestrow}$14.729(\pm0.128)$ & \cellcolor{bestrow}$0.648(\pm0.004)$ & \cellcolor{bestrow}$0.848(\pm0.008)$ & \cellcolor{bestrow}$0.271(\pm0.003)$ & \cellcolor{bestrow}$0.256(\pm0.017)$ & \cellcolor{bestrow}$0.323(\pm0.013)$ & \cellcolor{bestrow}$0.348(\pm0.020)$ \\
    & DARE-GRAM & TB & $0.211(\pm0.003)$ & $0.266(\pm0.001)$ & $0.001(\pm0.000)$ & $0.001(\pm0.001)$ & $11.402(\pm0.125)$ & $14.661(\pm0.062)$ & $0.651(\pm0.006)$ & $0.844(\pm0.004)$ & $0.273(\pm0.004)$ & $0.255(\pm0.017)$ & $0.340(\pm0.025)$ & $0.365(\pm0.017)$ \\
    \cmidrule(lr){2-15}
    & Deep Coral & DEV & $0.219(\pm0.015)$ & $0.288(\pm0.026)$ & $0.001(\pm0.000)$ & $0.002(\pm0.001)$ & $11.951(\pm0.909)$ & $16.207(\pm1.831)$ & $0.687(\pm0.056)$ & $0.940(\pm0.112)$ & $0.282(\pm0.009)$ & $0.258(\pm0.005)$ & $0.320(\pm0.014)$ & $0.341(\pm0.020)$ \\
    & Deep Coral & IWV & $0.220(\pm0.008)$ & $0.281(\pm0.010)$ & $0.001(\pm0.000)$ & $0.001(\pm0.000)$ & $12.027(\pm0.529)$ & $15.722(\pm0.653)$ & $0.691(\pm0.033)$ & $0.909(\pm0.041)$ & $0.284(\pm0.013)$ & $0.269(\pm0.008)$ & $0.335(\pm0.031)$ & $0.371(\pm0.041)$ \\
    & Deep Coral & SB & $0.209(\pm0.004)$ & $0.271(\pm0.004)$ & $0.001(\pm0.000)$ & $0.001(\pm0.001)$ & $11.334(\pm0.193)$ & $15.004(\pm0.231)$ & $0.647(\pm0.011)$ & $0.866(\pm0.014)$ & $0.279(\pm0.011)$ & $0.264(\pm0.012)$ & $0.336(\pm0.035)$ & $0.366(\pm0.032)$ \\
    & Deep Coral & TB & $0.209(\pm0.004)$ & $0.269(\pm0.006)$ & $0.001(\pm0.000)$ & $0.001(\pm0.000)$ & $11.328(\pm0.189)$ & $14.903(\pm0.335)$ & $0.648(\pm0.011)$ & $0.860(\pm0.021)$ & $0.278(\pm0.011)$ & $0.266(\pm0.010)$ & $0.325(\pm0.015)$ & $0.360(\pm0.019)$ \\
    \bottomrule
  \end{tabular}
  }
\end{table}

\cref{tab:geometric_pointnet} shows that \ac{UDA} algorithms can boost target performance compared to the unregularized baseline model also in this setting.
However compared to our chosen benchmarking design in \cref{tab:motor_results}, both the performance of the unregularized baseline as well as the one of the best performing \ac{UDA} method is worse, which supports our choice of explicitly conditioning on scalar parameters in the main benchmark.

\subsection{Two-dimensional Shifts}
\label{app:2d_shifts}
Defining shifts based on one parameter allows for controlled experiments, especially since that the parameters were picked based on preliminary experiments (see \cref{app:distribution_shifts}) and consultation with domain experts.
However, in real-world scenarios distribution shifts often affect multiple parameters simultaneously rather than being limited to a single one.
It is therefore important to investigate the performance of the benchmarked \ac{UDA} algorithms under multidimensional parameter shifts.
This is why \ourmethod~allows creating arbitrary dimensional n-dimensional shifts.
In this ablation, we investigate the behavior for a two-dimensional parameter shift on the \textit{hot rolling} dataset.

\begin{figure}[ht]
    \centering
    \includegraphics[width=0.4\linewidth]{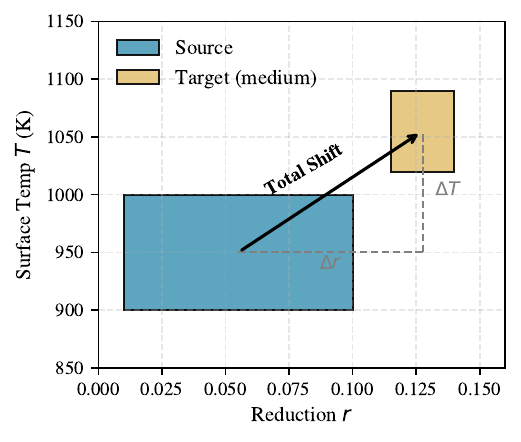}
    \caption{Visualization of the two-dimensional shift in parameter space for the \emph{hot rolling} dataset.}
    \label{fig:splits_visualization_2d}
\end{figure}

To be concise, we jointly shift reduction $r$ (this parameter also defines the one-dimensional shift in the main benchmark) and the surface temperature $T$ of the slab.
\cref{fig:splits_visualization_2d} visualizes the shift in parameter space between the source and the target domain.

We again train all models with each \ac{UDA} algorithm following the procedure in \cref{sec:experimental_setup} and \cref{app:experimental_setup} (seven regularization strengths and four seeds), and report the results in \cref{tab:results_2d_shifts}.

Comparing these results with the original one-dimensional shift (\cref{tab:rolling_results}), two observations stand out:  
\begin{enumerate*}[label=(\roman*)]
  \item Across all architectures architectures, the gap between source and target measured by the average field \ac{nRMSE} is higher than the one in the one-dimensional shift setting, confirming that the two-dimensional shift is a more challenging task.
  \item \ac{UDA} also improves target performance by remarkable margins in this challenging cale, although the relative improvements over the unregularized model are not as large as in the one-dimensional case.
\end{enumerate*}

These findings highlight the potential of \ac{UDA} to handle increasingly complex distribution shifts, underscoring its practical relevance for real-world applications.

\begin{table}[ht]
  \centering
  \caption{Performance metrics on the source and target domains (mean $\pm$ std over 4 seeds) for the \emph{hot rolling} with a two-dimensional shift. RMSE is reported for all global metrics (All Fields Normalized Avg through Equivalent Plastic Strain), while further columns report custom engineering and physics-based metrics. \textbf{Bold} indicates the overall best combination of architecture, \ac{UDA} algorithm, and model selection. Within each architecture group, the unregularized baseline is shaded \tightbox{baselinerow}{beige}, and the best \ac{UDA} configuration is \underline{underlined} and shaded \tightbox{bestrow}{green}.}  \label{tab:results_2d_shifts}
  \resizebox{\textwidth}{!}{%
  \definecolor{bestrow}{HTML}{DFF0D8}
  \definecolor{baselinerow}{HTML}{FFF4CC}
\begin{tabular}{lllcccccccccccccccccc}
    \toprule
    \multirow{2}{*}{\textbf{Model}} & \multirow{2}{*}{\makecell{\textbf{DA}\\ \textbf{Algorithm}}} & \multirow{2}{*}{\makecell{\textbf{Model}\\ \textbf{Selection}}} & \multicolumn{2}{c}{\textbf{All Fields Normalized Avg (-)}} & \multicolumn{2}{c}{\textbf{Deformation (mm)}} & \multicolumn{2}{c}{\textbf{Cauchy Stress (MPa)}} & \multicolumn{2}{c}{\textbf{Mises Stress (MPa)}} & \multicolumn{2}{c}{\textbf{Logarithmic Strain ($\mathbf{\times 10^{-2}}$)}} & \multicolumn{2}{c}{\textbf{Equivalent Plastic Strain ($\mathbf{\times 10^{-2}}$)}} & \multicolumn{2}{c}{\textbf{Temperature (K)}} & \multicolumn{2}{c}{\textbf{Relative Custom Error (-)}} & \multicolumn{2}{c}{\textbf{VM Consistency (-)}} \\
\cmidrule(lr){4-5} \cmidrule(lr){6-7} \cmidrule(lr){8-9} \cmidrule(lr){10-11} \cmidrule(lr){12-13} \cmidrule(lr){14-15} \cmidrule(lr){16-17} \cmidrule(lr){18-19} \cmidrule(lr){20-21}
      &   &  & \textbf{SRC} & \textbf{TGT} & \textbf{SRC} & \textbf{TGT} & \textbf{SRC} & \textbf{TGT} & \textbf{SRC} & \textbf{TGT} & \textbf{SRC} & \textbf{TGT} & \textbf{SRC} & \textbf{TGT} & \textbf{SRC} & \textbf{TGT} & \textbf{SRC} & \textbf{TGT} & \textbf{SRC} & \textbf{TGT} \\
\bottomrule
  \end{tabular}
  }
\end{table}


\end{document}